\documentclass[journal]{IEEEtran}
\usepackage{amsmath,amsfonts}
\usepackage{bm} 
\usepackage{algorithm}
\usepackage[noend]{algpseudocode}
\usepackage{array}
\usepackage{subcaption}
\usepackage{textcomp}
\usepackage{stfloats}
\usepackage{url}
\usepackage{verbatim}
\usepackage{graphicx}
\usepackage{cite}
\usepackage[numbers,sort&compress]{natbib}
\usepackage{color}
\usepackage[inline]{enumitem}
\usepackage{upgreek}
\usepackage{booktabs}
\usepackage{multirow}
\usepackage{tabularx}
\usepackage{ragged2e}
\usepackage{soul}
\usepackage[para,online,flushleft]{threeparttable}
\usepackage{mathtools}
\usepackage{stmaryrd}
\usepackage{tikz}
\usepackage{setspace}
\usepackage{amssymb} 

\DeclareMathOperator*{\argmax}{arg\,max}
\DeclareMathOperator*{\argmin}{arg\,min}

\newcommand{\cop}[1]{{#1}\allowbreak} 
\newcommand{\search}{ETS}
\newcommand{\dsearch}{D\text{-}ETS}
\newcommand{\classical}{C\text{-}ETS}
\newcommand{\dclassical}{CD\text{-}ETS}
\newcommand{\gspt}{GSPT}
\newcommand{\milaps}{Milaps}
\newcommand{\objective}{\mathrm{ET}}
\newcommand{\etal}{\textit{et~al.}}

\algrenewcommand\algorithmicrequire{\textbf{Input:}}
\algrenewcommand\algorithmicensure{\textbf{Output:}}
\algnewcommand\algorithmicparameters{\textbf{Parameters:}}

\allowdisplaybreaks 

\begin{document}

    \title{
        Anytime Metaheuristic Framework for Global Route Optimization in Expected-Time Mobile Search
    }

    \author{
    Jan Mikula$^{1,2}$, Miroslav Kulich$^{1}$%
    \thanks{
        This work was co-funded by the European Union under the project Robotics and advanced industrial production (reg. no. \texttt{\footnotesize CZ.02.01.01/\allowbreak{}00/\allowbreak{}22\_008/\allowbreak{}0004590}) and by the Vrant Agency of the Czech Technical University in Prague, grant no. \texttt{\footnotesize SVS23/\allowbreak{}175/\allowbreak{}OHK3/\allowbreak{}3T/\allowbreak{}13}.
    }%
    \thanks{
        $^{1}$Both authors are with the Czech Institute of Informatics, Robotics and Cybernetics, Czech Technical University in Prague, Jugoslavskych partyzanu 1580/3, Prague~6, 160\,00, Czech Republic.\\%
        \texttt{\footnotesize jan.mikula@cvut.cz},
        \texttt{\footnotesize miroslav.kulich@cvut.cz}
    }%
    \thanks{
        $^{2}$Jan Mikula is also with the Department of Cybernetics, Faculty of Electrical Engineering, Czech Technical University in Prague, Karlovo namesti 293/13, Prague~2, 121\,35, Czech Republic.
    }%
    }



    \maketitle

    \begin{abstract}
        Expected-time mobile search (\search{}) is a fundamental robotics task where a mobile sensor navigates an environment to minimize the expected time required to locate a hidden object.
        Global route optimization for \search{} in static 2D continuous environments remains largely underexplored due to the intractability of objective evaluation, stemming from the continuous nature of the environment and the interplay of motion and visibility constraints.
        Prior work has addressed this through partial discretization, leading to discrete-sensing formulations tackled via utility-greedy heuristics.
        Others have taken an indirect approach by heuristically approximating the objective using minimum latency problems on fixed graphs, enabling global route optimization via efficient metaheuristics.
        This paper builds on and significantly extends the latter by introducing \milaps{} (\setul{1.5pt}{}\nolinebreak{}\ul{Mi}nimum \ul{la}tency \ul{p}roblem\ul{s}\nolinebreak{}), a model-based solution framework for \search{}.
        \milaps{} integrates novel auxiliary objectives and adapts a recent anytime metaheuristic for the traveling deliveryman problem, chosen for its strong performance under tight runtime constraints.
        Evaluations on a novel large-scale dataset demonstrate superior trade-offs between solution quality and runtime compared to state-of-the-art baselines.
        The best-performing strategy rapidly generates a preliminary solution, assigns static weights to sensing configurations, and optimizes global costs metaheuristically.
        Additionally, a qualitative study highlights the framework’s flexibility across diverse scenarios.
    \end{abstract}


    \section{Introduction}
    \label{sec:introduction}

    Efficient target search is crucial in many real-world applications.
    These include search-and-rescue (S\&R) missions, where mobile robots navigate hazardous environments to locate survivors~\citep{Cubber2017}, industrial inspection, where autonomous robots enhance safety and monitoring efficiency~\citep{Huamanchahua2021}, and environmental monitoring and wildlife tracking, where mobile agents collect data in challenging terrains to support ecological research and conservation~\citep{Chen2024}.
    A key challenge in these scenarios is devising search strategies that minimize the time to locate hidden targets, ensuring timely and effective task completion.

    \begin{figure}
        \centering
        \begin{subfigure}[t]{0.325\columnwidth}
            \centering
            \includegraphics[width=\linewidth]{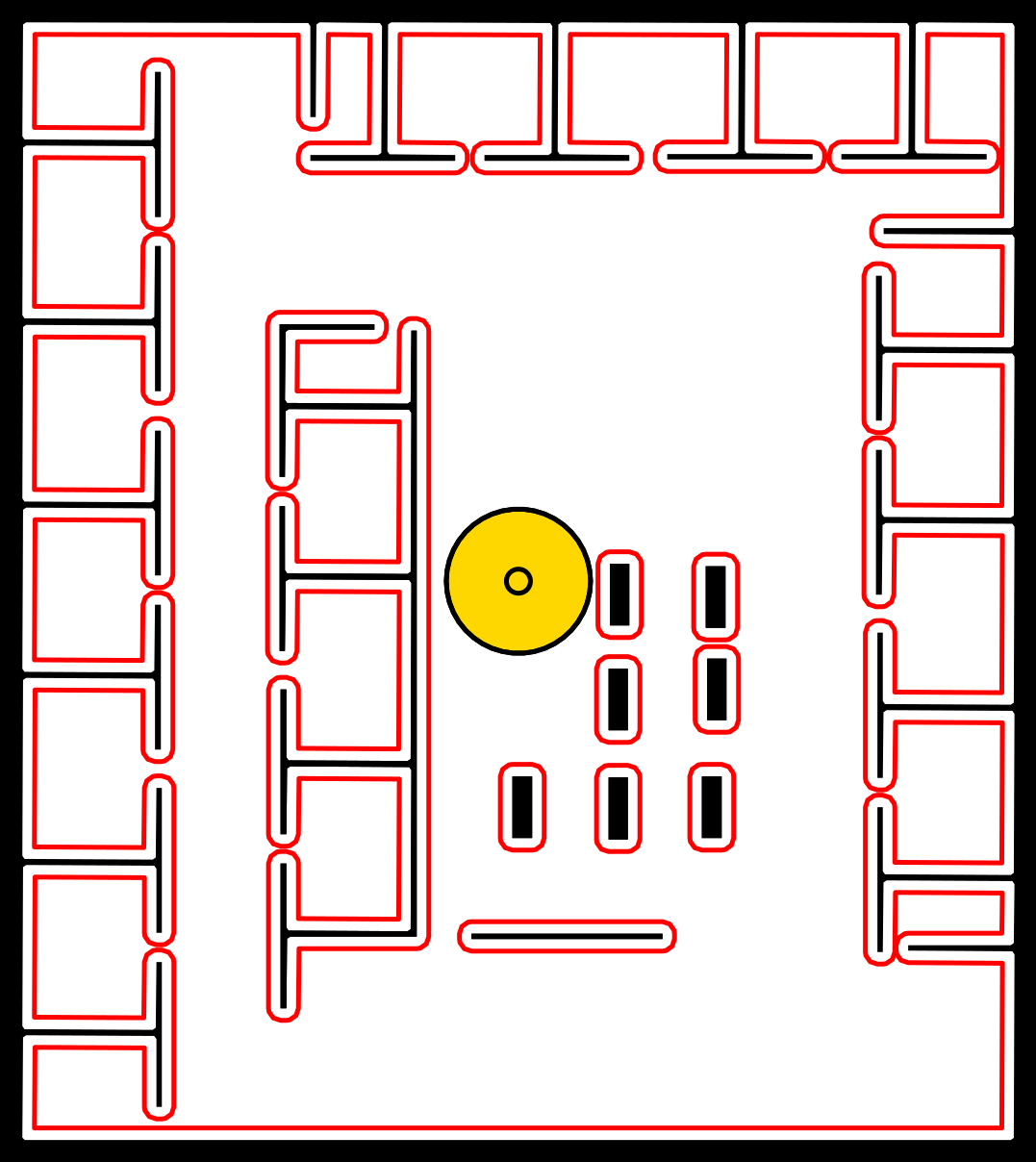}
            \caption{Map, sensor, conf. space.}
            \label{fig:motivation-example-map}
        \end{subfigure}
        \hfill
        \begin{subfigure}[t]{0.325\columnwidth}
            \centering
            \includegraphics[width=\linewidth]{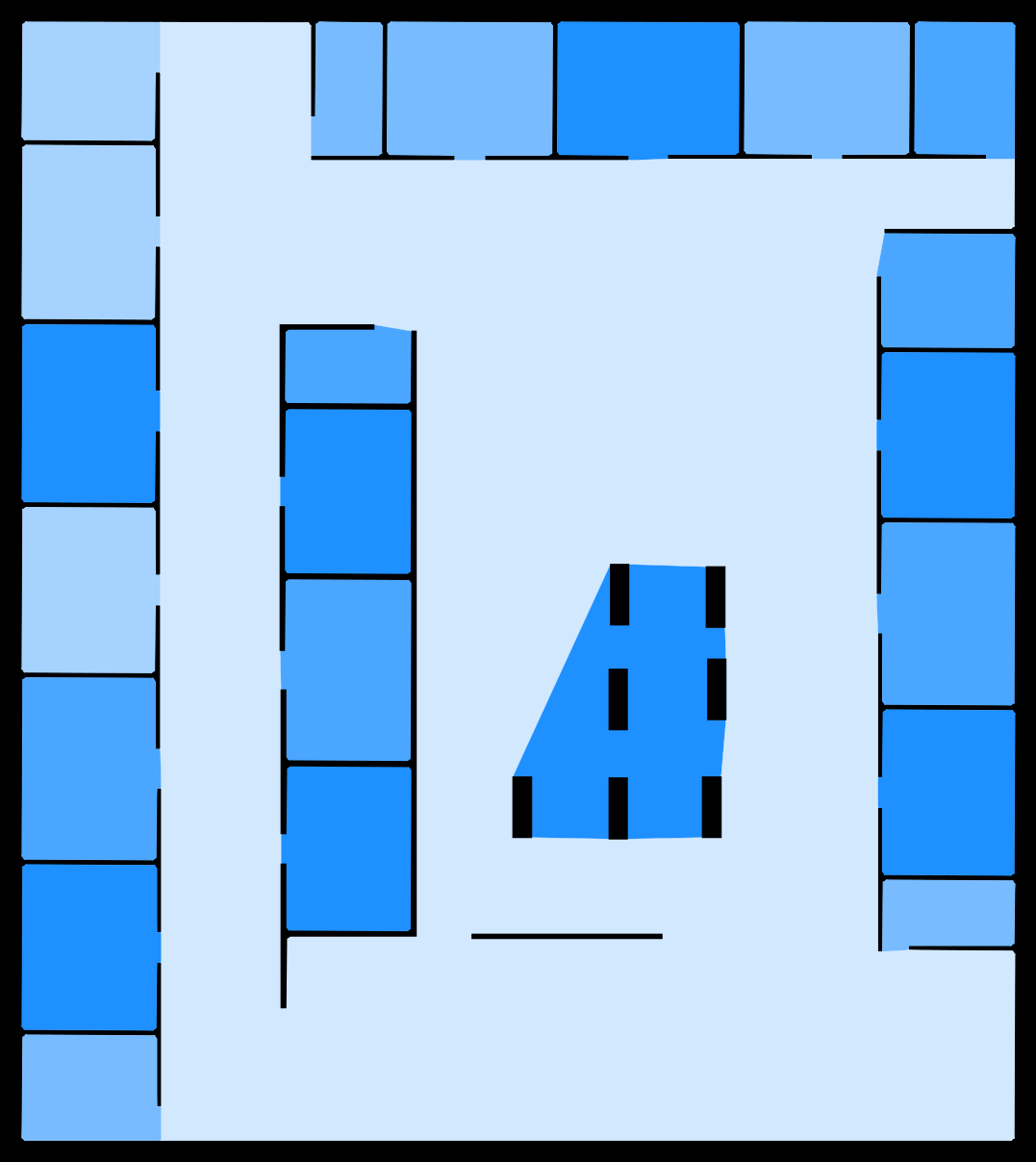}
            \caption{Object's prob. distrib.}
            \label{fig:motivation-example-target}
        \end{subfigure}
        \hfill
        \begin{subfigure}[t]{0.325\columnwidth}
            \centering
            \includegraphics[width=\linewidth]{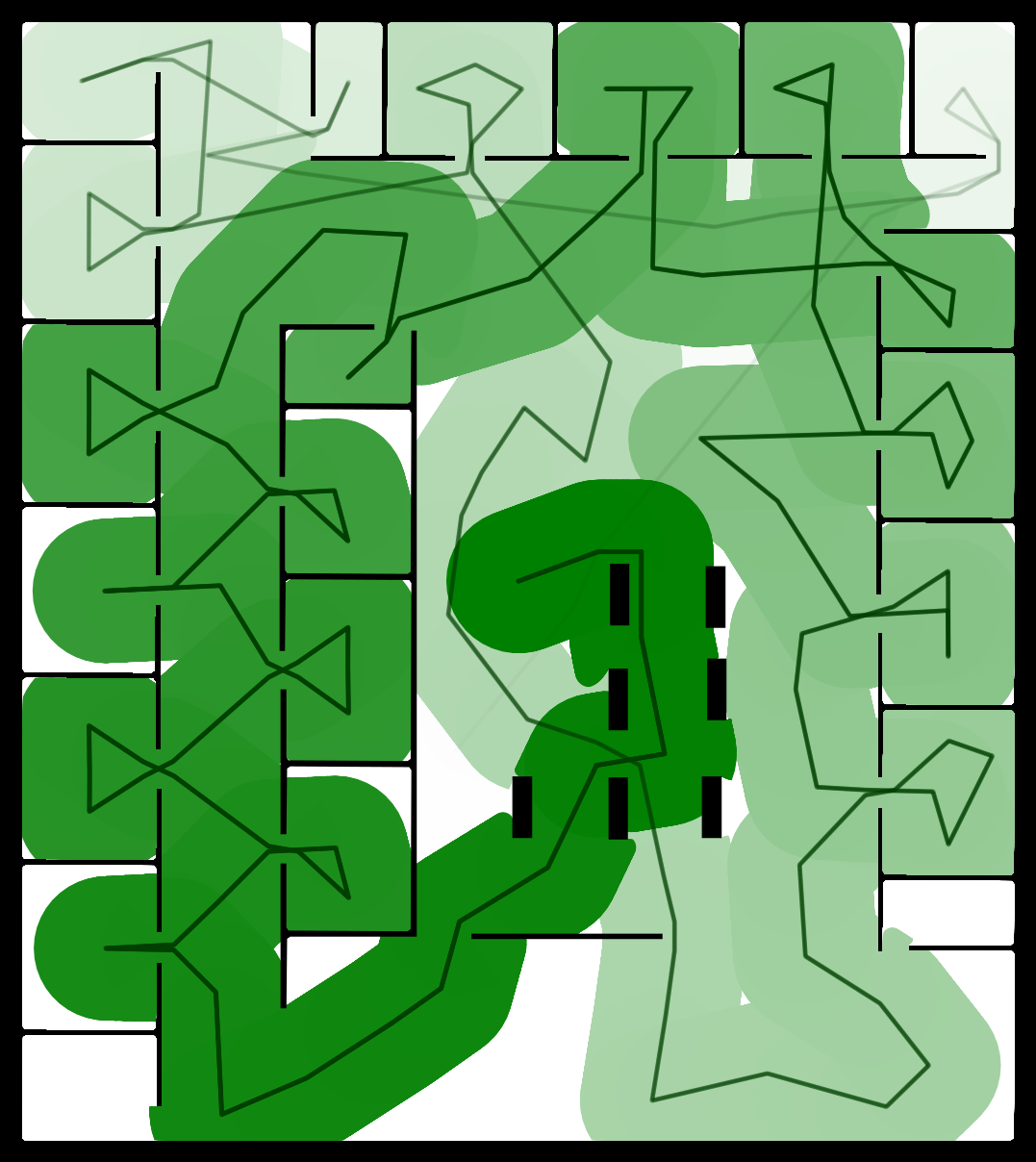}
            \caption{Optimized search route.}
            \label{fig:motivation-example-path}
        \end{subfigure}
        \caption{
            Motivating example of a search scenario.
        }
        \label{fig:motivation-example}
    \end{figure}

    This paper considers a search scenario in a static, 2D continuous environment with obstacles, where a mobile omnidirectional sensor must locate a static target (object) with a known probability distribution.
    Fig.~\ref{fig:motivation-example} illustrates a motivating example in which a sensor with a circular footprint searches for an object in a polygonal environment.
    In Fig.~\ref{fig:motivation-example-map}, black regions denote obstacles, white represents free space, and the sensor is shown in yellow in its initial configuration.
    The inner circle indicates the sensor's location and footprint, while the outer circle depicts its sensing range.
    Red lines mark the boundary of its free configuration space, where the sensor can move without colliding with obstacles.
    Fig.~\ref{fig:motivation-example-target} shows the probability distribution of the object's location, which spans the entire environment, with darker shades indicating higher probabilities.
    We assume this distribution is given and that the object is detected as soon as it enters the sensor's field of view.
    The challenge is to plan the sensor's route for the most efficient search, considering the environment, the probability distribution, and visibility and motion constraints.
    Fig.~\ref{fig:motivation-example-path} illustrates a partial search route starting from the initial configuration, where covered areas are shown in green, with the shade of green indicating the execution time of the search (darker shades represent areas covered earlier).

    The illustrated search task typically falls into two main categories~\citep{Sarmiento2003}: \emph{worst-case search} and \emph{expected-time search}.
    In \emph{worst-case search}, the objective is to provide a guarantee of finding the object while optimizing the time at which this guarantee is achieved, which leads to shortest-time routes that provide complete coverage of the environment, rendering the object's probability distribution uninformative.
    This scenario is relevant when the number of objects to be searched is unknown and the search must therefore be exhaustive, or in inspection tasks where all sites must be visited~\citep{Huamanchahua2021}.
    In contrast, \emph{expected-time search} minimizes the average time to find the object by prioritizing likely locations early, at the cost of a longer worst-case search time~\citep{Sarmiento2003}.
    This approach is crucial in scenarios like disaster victim searches, where survival chances decline over time, dynamic deadlines exist (e.g., building collapse), or even in routine tasks like locating household objects.

    In this work, we focus on the expected-time objective, which has received less attention in the literature compared to the well-studied worst-case objective, for which extensive inspection and coverage path planning methods exist \citep[e.g.,][]{Packer2008,Galceran2013,Huamanchahua2021}.
    We formulate the \emph{expected-time mobile search} (\search{}) as a constrained optimization problem in continuous space, aiming to minimize the expected time to locate the object while ensuring that the sensor follows a collision-free route from a given initial configuration and detects the object with a specified probability by the end of the route.
    Our formulation is closely related to \emph{robotics} due to its potential applications~\citep{Cubber2017}, to the class of NP-hard \emph{visibility-based optimization problems} in \emph{computational geometry} (CG)~\citep{ORourke2017}, and to NP-hard \emph{routing problems} in \emph{combinatorial optimization} (CO)~\citep{Macharet2018}, particularly when the space is discretized.

    This paper primarily focuses on optimizing the expected-time objective in continuous environments, which is significantly more challenging than typical objectives in related fields of CG and CO\@.
    For instance, the \emph{watchman route problem} (WRP)~\citep{Chin1986} in CG and the \emph{traveling salesman problem} (TSP)~\citep{Cook2015} in CO seek to minimize route length, an objective that is relatively easy to compute.
    In contrast, the \search{} objective requires integrating over all possible object locations visible to the sensor along the entire route.
    This integral is generally intractable, and even its discretized approximation (used in our solution) is computationally expensive, especially for long routes in large-scale, complex environments.

    Prior work has addressed this challenge through partial discretization, restricting sensing to a predefined discrete set of configurations within an otherwise continuous environment~\citep{Sarmiento2003}.
    The resulting \emph{discrete-sensing} \search{} (\dsearch{}) problem has been tackled using \emph{utility-based greedy constructive heuristics}~\citep{Sarmiento2003,Sarmiento2004c}, and locally optimal paths have been proposed to bridge the gap between discrete and continuous-sensing \search{}~\citep{Sarmiento2004,Sarmiento2009}.
    Other approaches approximate the \dsearch{} objective indirectly using \emph{minimum latency problems} (MLPs) defined on fixed graphs~\citep{Kulich2014,Kulich2017,Kulich2022}, such as the \emph{traveling deliveryman problem} (TDP)~\citep{Lucena1990} and the \emph{graph search problem} (GSP)~\citep{Koutsoupias1996}.
    This objective simplification enables global route optimization using efficient metaheuristic algorithms tailored to the TDP and GSP\@.

    In this paper, we build upon and significantly extend the MLP approach by introducing \emph{\milaps{}} (a name rather than an acronym, derived from \emph{\setul{1.75pt}{}\hspace{0.05em}\nolinebreak{}\ul{Mi}nimum \ul{la}tency \ul{p}roblem\ul{s}\hspace{0.05em}\nolinebreak{})}, a~model-based solution framework for \dsearch{}.
    This framework is employed within a standard \emph{heuristic decoupling scheme}~\cite{Packer2008} to also address the continuous-sensing \search{} problem.
    The decoupling scheme for \search{} constructs a discretized solution space, ensuring constraint satisfaction and transforming the problem into \dsearch{}, which is then solved by \milaps{}.
    The core components of \milaps{} include:
    \begin{enumerate*}[label=(\Roman*)]
        \item \emph{Novel auxiliary objectives} that approximate the \dsearch{} objective by estimating the probability of detecting the object at each sensing configuration using a static, route-independent weight.
        \item A novel MLP formulation, referred to as GSP \emph{with turning} (\gspt{}), which extends the previous GSP formulation to account for turning times at sensing locations.
        \item An adaptation of a recent anytime metaheuristic algorithm for the TDP~\citep{Mikula2022} to optimize the \gspt{} objective.
        \item A \emph{replanning scheme} for \gspt{} that improves \search{}/\dsearch{} solution quality when static weights prove significantly inaccurate.
    \end{enumerate*}

    It is worth noting that while we primarily address a static, offline version of the search scenario, our focus is on generating the best possible route plan within a given computational time budget, leveraging the anytime nature of the metaheuristic algorithm.
    As a result, \milaps{} may also be applicable to online, moderately dynamic settings through adaptations of the replanning scheme, although this aspect is not explicitly addressed in this paper.\footnote{
        A limiting factor in this case is the need for very high replanning frequencies, which may restrict the solution's applicability.
        In principle, this limitation could be mitigated through hierarchical planning, where \milaps{} operates at a higher level with a lower replanning frequency.
    }

    \milaps{} is extensively evaluated in a computational study, which includes generating a novel large-scale \dsearch{} dataset, performing a \emph{quantitative evaluation} against heuristic baselines in \emph{classic} \dsearch{}/\search{} scenarios, and conducting a \emph{qualitative study} to demonstrate the framework's flexibility.

    The remainder of this paper is structured as follows.
    Sec.~\ref{sec:problem-statement} introduces the problem formulation and key definitions.
    Sec.~\ref{sec:related-work} reviews related work on \search{} and similar problems, positioning this paper's contributions within the existing literature.
    Sec.~\ref{sec:proposed-solution} details the proposed solution framework.
    Sec.~\ref{sec:quantitative-evaluation} presents the methodology and results of its quantitative evaluation.
    Sec.~\ref{sec:qualitative-evaluation} provides a qualitative study showcasing its adaptability across various scenarios.
    Finally, Sec.~\ref{sec:conclusions} discusses the framework's broader applicability and concludes with directions for future research.

    \section{Problem Formulation and Related Definitions}
    \label{sec:problem-statement}

    \subsection{Notation and Nomenclature Remarks}
    \label{subsec:nomenclature}

    From this point onward, we adopt a consistent notation and nomenclature to enhance clarity while allowing flexibility for context-specific symbol definitions.
    Distinct fonts represent specific mathematical objects (e.g., $\mathbb{F}$, $\mathcal{F}$, $\mathfrak{F}$), while other fonts and symbols are used more flexibly (e.g., $\mathrm{f}$, $\mathrm{F}$, $f$, $F$, $\phi$, $\Phi$).

    Uppercase Roman letters in the font $\mathbb{A}\mathbb{B}\mathbb{C}$ are strictly reserved for {specific mathematical structures}, as follows:
    $\mathbb{C}$~represents the configuration space of a mobile sensor.
    $\mathbb{E}$~denotes the expected value of a random variable.
    $\mathbb{G}$~denotes a graph.
    $\mathbb{J}$~represents the parameter space of a local search operator in a metaheuristic algorithm.
    $\mathbb{O}$~represents the computational complexity of an algorithm.
    $\mathbb{R}$~denotes the set of real numbers.
    $\mathbb{Z}$~denotes the set of integers.
    The font $\mathcal{A}\mathcal{B}\mathcal{C}$ denotes spatial sets, specifically bounded, closed, but not necessarily connected subsets of $\mathbb{R}^2$.
    The boundary of a set $\mathcal{A}$ is denoted by $\partial\mathcal{A}$, and its area by $\mathrm{Area}(\mathcal{A})$.
    Finally, the font $\mathfrak{A}\mathfrak{B}\mathfrak{C}$ strictly represents random variables.
    All of the above symbols may include subscripts or superscripts, but their general meaning remains consistent.
    Other symbols, such as those in the default font (both lowercase and uppercase) or Greek letters, are context-dependent and may vary throughout the paper.

    The following notations are used for common functions and operations:
    $\llbracket \cop{.} \rrbracket$ denotes the Iverson bracket.
    $\| \cop{.} \|$ denotes the Euclidean norm.
    $\overline{qp}$ represents the line segment connecting two points $q$ and $p$.
    $|A|$ denotes the cardinality of a set $A$ or the absolute value of a scalar $A$, depending on the operand.
    $2^A$ represents the power set of $A$.
    $\mathrm{cl}(A)$ denotes the closure of a set $A$.
    The notation $\{i{:}j\}$, where $i, j \cop{\in} \mathbb{Z}$ and $i \cop{\leq} j$, represents the set of integers from $i$ to $j$, inclusive.
    In other cases, $\{. {:} .\}$ denotes set-builder notation.
    $\langle a_{i} \rangle_{i=k}^{n}$ represents an ordered sequence of $n \cop{-} k \cop{+} 1$ elements, indexed from $k$ to $n$ (inclusive).
    A similar notation, $\{a_{i}\}_{i=k}^{n}$, is used for finite sets, where the index $i$ labels the elements, but no specific order is implied.
    Finally, the symbol $\mid$ is reserved for representing conditioning, typically in probability distributions, though we also extend its use to other functions and objects when the context is clear.
    Any additional notation used in the paper follows standard conventions or is defined where it first appears.

    \subsection{Problem Definition: Expected-Time Mobile Search (\search{})}
    \label{subsec:problem-definition}

    The \search{} problem is set in a known, static \emph{environment} $\mathcal{W} \cop{\subset} \mathbb{R}^2$, which is a non-empty, path-connected, bounded, and closed region of the Euclidean plane.
    The complement of the environment, $\mathcal{W}_{\mathrm{obs}} \cop{\coloneqq} \mathbb{R}^2 \cop{\setminus} \mathcal{W}$, represents \emph{obstacles} that impose motion and visibility constraints on the observer.
    The observer is an omnidirectional \emph{mobile sensor} with configuration space $\mathbb{C} \cop{\coloneqq} \mathbb{R}^2$ and a circular footprint $\mathcal{A} \cop{:} \mathbb{C} \cop{\mapsto} 2^{\mathbb{R}^2}$, defined as $\mathcal{A}(q) \cop{\coloneqq} \{p \cop{\in} \mathbb{R}^2 \cop{:} \|p \cop{-} q\| \cop{\leq} r_{\mathrm{fp}}\}$, where $r_{\mathrm{fp}} \cop{\geq} 0$ is the footprint radius.
    The sensor's free configuration space is given by $\mathbb{C}_{\mathrm{free}} \cop{\coloneqq} \{ q \cop{\in} \mathbb{C} \cop{:} \mathcal{A}(q) \cop{\cap} \mathcal{W}_{\mathrm{obs}} \cop{=} \emptyset \}$.
    We define a sensor's path $\tau \cop{:} [0{,}1] \cop{\mapsto} \mathbb{C}$ as a connected curve in $\mathbb{C}$ parameterized by $\nu \cop{\in} [0{,}1]$.
    This definition encompasses both continuous and piecewise continuous paths.
    For brevity, we assume that the path is continuous in subsequent definitions.
    Nonetheless, these definitions can be easily adapted to accommodate piecewise continuous paths.

    Furthermore, the mobile sensor is described by two additional models: an omnidirectional \emph{visibility model} $\mathrm{Vis} \cop{:} \mathbb{C} \cop{\mapsto} 2^{\mathcal{W}}$ and a symmetric \emph{travel time model} $\mathrm{Time} \cop{|} \tau \cop{:} [0{,}1] \cop{\mapsto} \mathbb{R}_{\geq 0}$, conditioned on the path $\tau$.
    The former defines the set of points in $\mathcal{W}$ visible from a configuration $q \cop{\in} \mathbb{C}$, while the latter represents the time required to travel from $\tau(0)$ to $\tau(\nu)$ along $\tau$.
    The visibility model is parameterized by a limited visibility radius $r_{\mathrm{vis}} \cop{>} r_{\mathrm{fp}}$, such that
    \begin{flalign*}
        \mathrm{Vis}(q) \coloneqq \{ p \in \mathcal{W} : \overline{q p} \subset \mathcal{W} \land \| q - p \| \leq r_{\mathrm{vis}} \},
    \end{flalign*}
    denotes the set of all points in $\mathcal{W}$ that have a direct line of sight from $q$ and are within a distance of $r_{\mathrm{vis}}$.
    The travel time model is parameterized by the inverse linear and angular velocities $t_{\mathrm{lin}} \cop{\geq} 0$ and $t_{\mathrm{ang}} \cop{\geq} 0$, respectively.
    For any $\nu \cop{\in} [0{,}1]$ on a given path $\tau$, it is defined as
    \begin{flalign}
        \mathrm{Time}(\nu \mid \tau) &\coloneqq t_{\mathrm{lin}} \mathrm{Len}(\nu \mid \tau) + t_{\mathrm{ang}} \mathrm{Ang}(\nu \mid \tau), \label{eq:travel-time-model}\\
        \mathrm{Len}(\nu \mid \tau) &\coloneqq \int_{0}^{\nu} \| \dot{\tau}(s) \| \, \mathrm{d}s,\nonumber\\
        \| \dot{\tau}(s) \| &\coloneqq \sqrt{ \left( \frac{\mathrm{d}x}{\mathrm{d}s} \right)^2 + \left( \frac{\mathrm{d}y}{\mathrm{d}s} \right)^2 }, \nonumber\\
        \mathrm{Ang}(\nu \mid \tau) &\coloneqq \int_{0}^{\nu} \left| \frac{\mathrm{d}\phi(s)}{\mathrm{d}s} \right| \, \mathrm{d}s, \nonumber\\
        \phi(s) &\coloneqq \mathrm{arctan2} \left( \frac{\mathrm{d}y}{\mathrm{d}s}, \frac{\mathrm{d}x}{\mathrm{d}s} \right).\nonumber
    \end{flalign}
    Here, $x$ and $y$ denote the coordinates of $\tau(s)$.
    In other words, the travel time along $\tau$ is a linear combination of the path length $\mathrm{Len}(\nu \cop{\mid} \tau)$ and the total turning angle accumulated along the path, $\mathrm{Ang}(\nu \cop{\mid} \tau)$.
    The model is \emph{symmetric}, meaning that for any pair $\tau, \tau'$ where $\tau'(\nu) \cop{\coloneqq} \tau(1 \cop{-} \nu)$, the relation $\mathrm{Time}(\nu \cop{\mid} \tau) \cop{+} \mathrm{Time}(1 \cop{-} \nu \cop{\mid} \tau') \cop{=} \mathrm{Time}(1 \cop{\mid} \tau) \cop{=} \mathrm{Time}(1 \cop{\mid} \tau')$ holds for all $\nu \cop{\in} [0{,}1]$.

    The \search{} problem considers a static \emph{object of interest} in the environment, whose location is given by the random variable $\mathfrak{X} \cop{\in} \mathcal{W}$ with a known probability density function $f_{\mathfrak{X}}$.
    We introduce a random variable $\mathfrak{V} \cop{\in} [0{,}1]$, conditioned on the path~$\tau$, representing the parameter value~$\nu$ at which the object is detected along~$\tau$, assuming complete coverage at $\nu \cop{=} 1$.
    Defining $\mathcal{W}_{\mathrm{seen}}(\nu \cop{\mid} \tau) \cop{\coloneqq} \bigcup_{\nu' \cop{\in} [0,\nu]} \mathrm{Vis}(\tau(\nu'))$ as the region seen by the sensor up to $\nu$, and assuming $\tau$ completely covers the environment, i.e., $\mathcal{W}_{\mathrm{seen}}(1 \cop{\mid} \tau) \cop{=} \mathcal{W}$, the cumulative distribution function of $\mathfrak{V}$ is
    \begin{flalign*}
        F_{\mathfrak{V} \mid \tau}(\nu \mid \tau) \coloneqq \int_{\mathcal{W}_{\mathrm{seen}}(\nu \mid \tau)} f_{\mathfrak{X}}(s) \, \mathrm{d}s.
    \end{flalign*}
    This integral accumulates the probability density of $\mathfrak{X}$ over the region $\mathcal{W}_{\mathrm{seen}}$.
    Finally, we define $\mathfrak{T} \cop{\in} \mathbb{R}_{\geq 0}$ as the time when the object is first detected along the path.
    The random variables $\mathfrak{T}$ and $\mathfrak{V}$ are linked through the travel time model via $\mathfrak{T} {\mid} \tau \cop{=} \mathrm{Time}(\mathfrak{V} \cop{\mid} \tau)$, meaning the detection time is determined by evaluating the travel time model at the point of first sighting.

    Finally, we define the \search{} objective as
    \begin{flalign}
        \label{eq:objective-cont}
        \objective{}(\tau) \coloneqq \int_{0}^{1} \mathrm{Time}(\nu \mid \tau) f_{\mathfrak{V} \mid \tau}(\nu \mid \tau) \, \mathrm{d}\nu.
    \end{flalign}
    Under the complete coverage assumption, this integral represents the \emph{expected detection time}, i.e., $\objective{}(\tau) \cop{=} \mathbb{E}(\mathfrak{T} \cop{\mid} \tau)$ by the definition of expectation.
    To mitigate diminishing returns in achieving full coverage, we introduce $\epsilon \cop{\in} [0{,}1]$, which controls the required detection probability at the end of the path, where $\epsilon \cop{=} 0$ enforces full coverage.
    For $\epsilon \cop{>} 0$, the objective relaxes to the expected detection time with a guaranteed probability of $(1 \cop{-} \epsilon)$.
    The complete \search{} problem is formulated as $\argmin_{\tau} \objective{}(\tau)$, subject to constraints:
    \begin{flalign}
        \label{eq:ets-constraint-path-start}
        \tau(0) &= g_{0}, \\
        \label{eq:ets-constraint-path}
        \tau &\subset \mathbb{C}_{\mathrm{free}}, \\
        \label{eq:ets-constraint-path-end}
        f_{\mathfrak{V} \mid \tau}(1 \mid \tau) &\geq 1 - \epsilon.
    \end{flalign}
    Eq.~\eqref{eq:ets-constraint-path-start} ensures the sensor starts at the specified initial configuration $g_{0}\cop{\in}\mathbb{C}_{\mathrm{free}}$, while Eq.~\eqref{eq:ets-constraint-path} guarantees a collision-free path.
    Eq.~\eqref{eq:ets-constraint-path-end} enforces detection with at least $(1 \cop{-} \epsilon)$ probability by the end of the path.
    A feasible solution to the \search{} problem is referred to as an \emph{\search{} route}, or simply a \emph{route}, in contrast to a path, which does not imply feasibility.
    This formulation generalizes the original continuous problem addressed in~\citep{Sarmiento2004,Sarmiento2009}, referred to here as \emph{classical \search{}} (\classical{}).
    \classical{} assumes a uniform probability distribution for the object's location, such that $F_{\mathfrak{V} \cop{\mid} \tau}(\nu \cop{\mid} \tau) \cop{\propto} \mathrm{Area}(\mathcal{W}_{\mathrm{seen}}(\nu \cop{\mid} \tau))$, and considers a point sensor with simpler visibility and travel time models: $r_{\mathrm{fp}} \cop{=} 0, r_{\mathrm{vis}} \cop{=} \infty, t_{\mathrm{lin}} \cop{=} 1, t_{\mathrm{ang}} \cop{=} 0$, while requiring full coverage ($\epsilon \cop{=} 0$).

    The main challenge in solving \search{} is the intractability of the integral required to compute the objective in Eq.~\eqref{eq:objective-cont}.
    Prior work addressed this using numerical methods~\citep{Sarmiento2004,Sarmiento2009}, but only for local path segments and specific cases, such as peeking behind a reflex vertex in a simple polygon under the \classical{} scenario, with prohibitive computational costs for more general cases.
    Clearly, the continuous objective is unsuitable for global optimization.
    Other approaches~\citep{Sarmiento2003,Sarmiento2004c} tackled the \emph{discrete-sensing \search{}} (\dsearch{}) problem by restricting sensor readings to a predefined set of sensing configurations, $G \cop{=} \{g_{i} \cop{\in} \mathbb{C}_{\mathrm{free}}\}_{i=0}^{n}$, ensuring full coverage but overlooking sensing opportunities along the paths between these configurations.
    In the next section, we derive a new tractable formulation of the \search{} problem that addresses these limitations and generalizes \dsearch{} as a special case.

    \subsection{Reformulating \search{} with a Tractable Objective}
    \label{subsec:tractable-formulation}

    The \search{} reformulation relies on three key assumptions:
    \begin{enumerate*}[label=(\roman*)]
        \item Sensor readings occur exclusively at discrete configurations along the path, governed by a so-called discrete \emph{sensing policy} $\mathrm{sens}(\tau)$.
        \item The object's location probability distribution is represented by \emph{weighted target regions}, $P \cop{=} \{(p_{i} \cop{\in} \mathbb{R}_{> 0},\allowbreak \mathcal{P}_{i} \cop{\subset} \mathcal{W})\}_{i=1}^m$, where each region $\mathcal{P}_{i}$ is closed, bounded, and has a positive weight $p_{i}$.
        See Fig.~\ref{fig:motivation-example-target} for an illustration.
        \item Clipping operations on closed, bounded subsets of $\mathcal{W}$, say $\mathcal{X},\mathcal{Y} \cop{\subset} \mathcal{W}$, are well-defined, including $\mathcal{X} \cop{\cap} \mathcal{Y}$, $\mathcal{X} \cop{\cup} \mathcal{Y}$, and $\mathrm{cl}(\mathcal{X} \cop{\setminus} \mathcal{Y})$.
    \end{enumerate*}

\begin{figure}
        \centering
        \begin{subfigure}[t]{0.32\columnwidth}
            \centering
            \includegraphics[width=\linewidth]{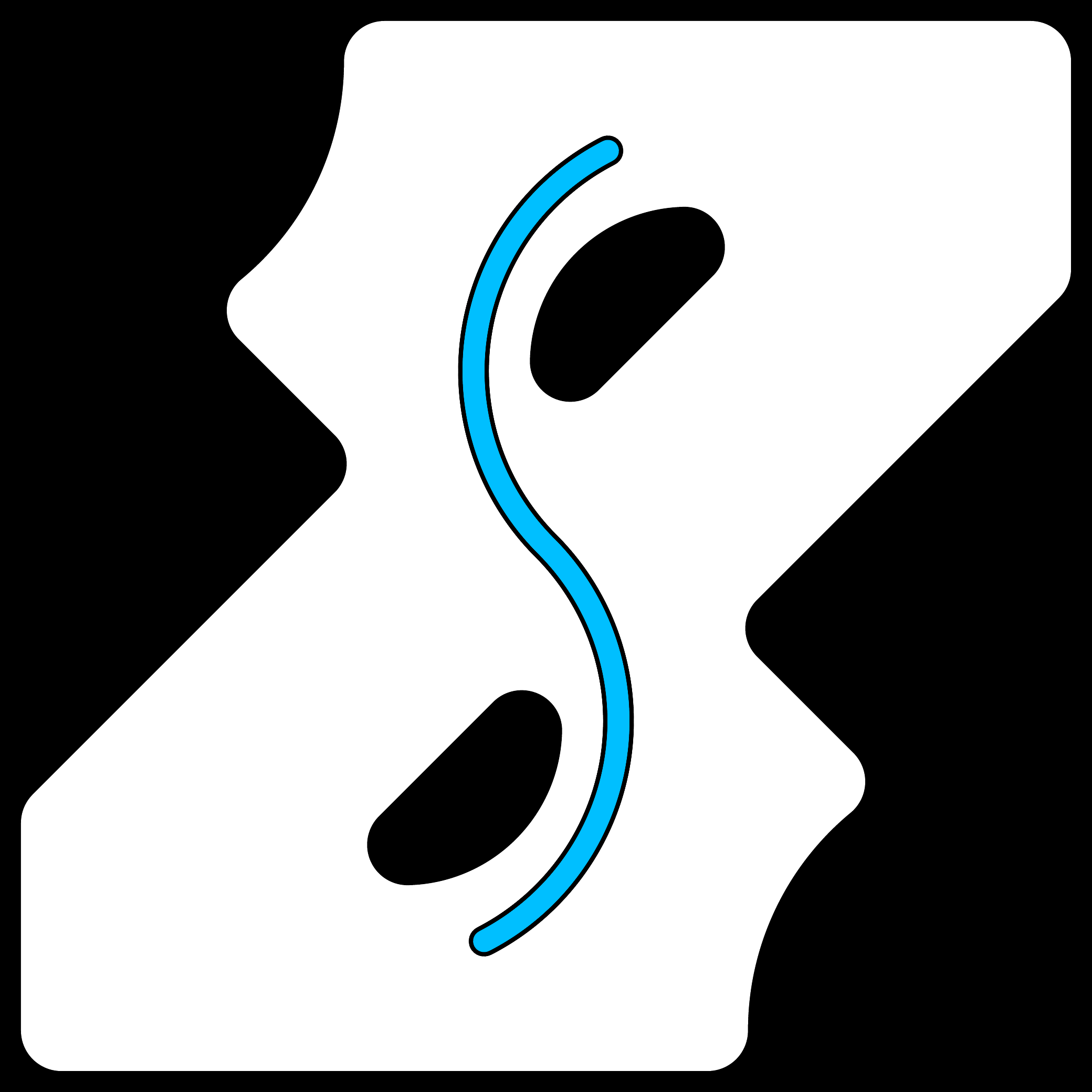}
            \caption{Environment $\mathcal{W}$, path $\tau$}
            \label{fig:ets-path}
        \end{subfigure}
        \hfill
        \begin{subfigure}[t]{0.32\columnwidth}
            \centering
            \includegraphics[width=\linewidth]{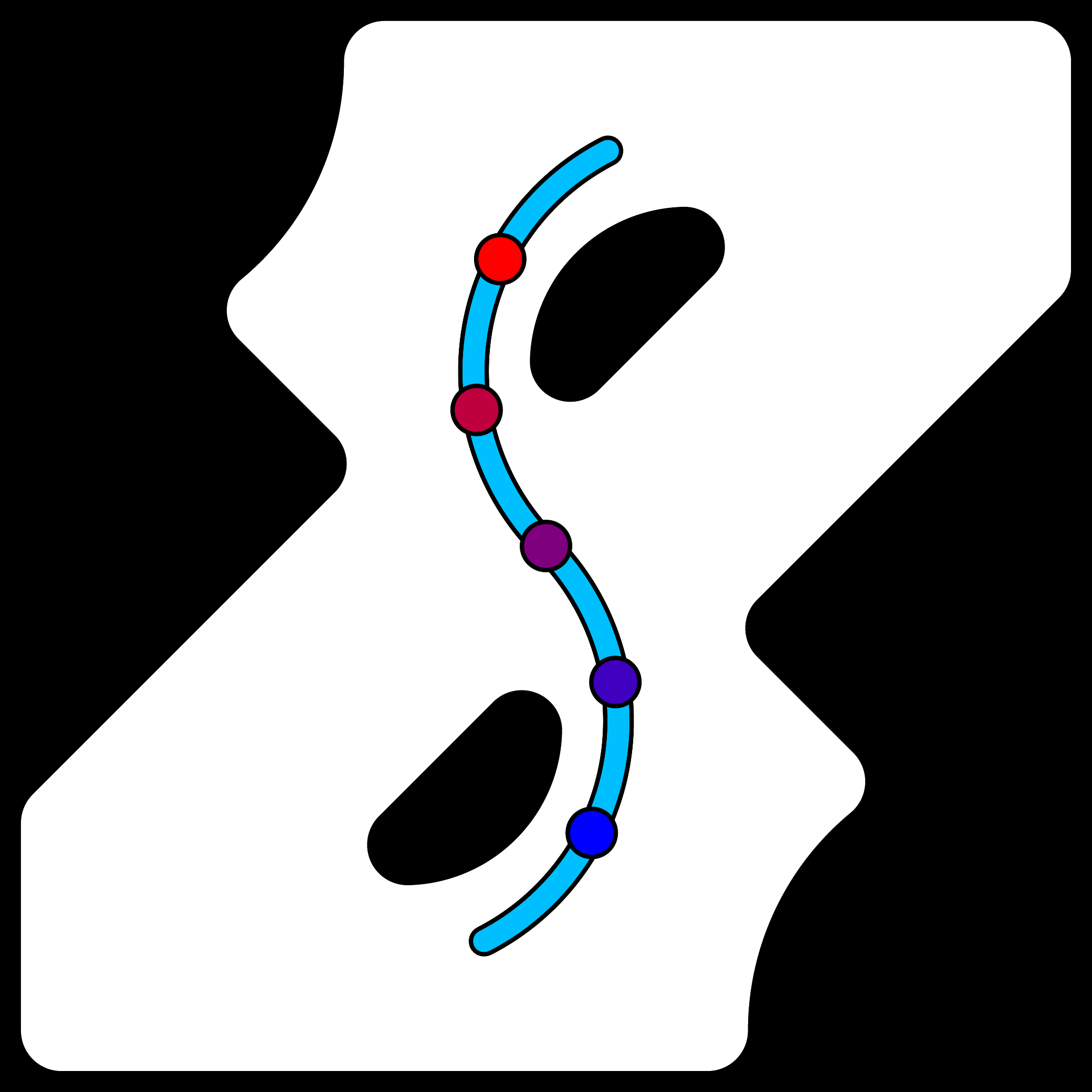}
            \caption{Sensing sequence $S_{n{=}4}$}
            \label{fig:ets-sensing-sequence}
        \end{subfigure}
        \hfill
        \begin{subfigure}[t]{0.32\columnwidth}
            \centering
            \includegraphics[width=\linewidth]{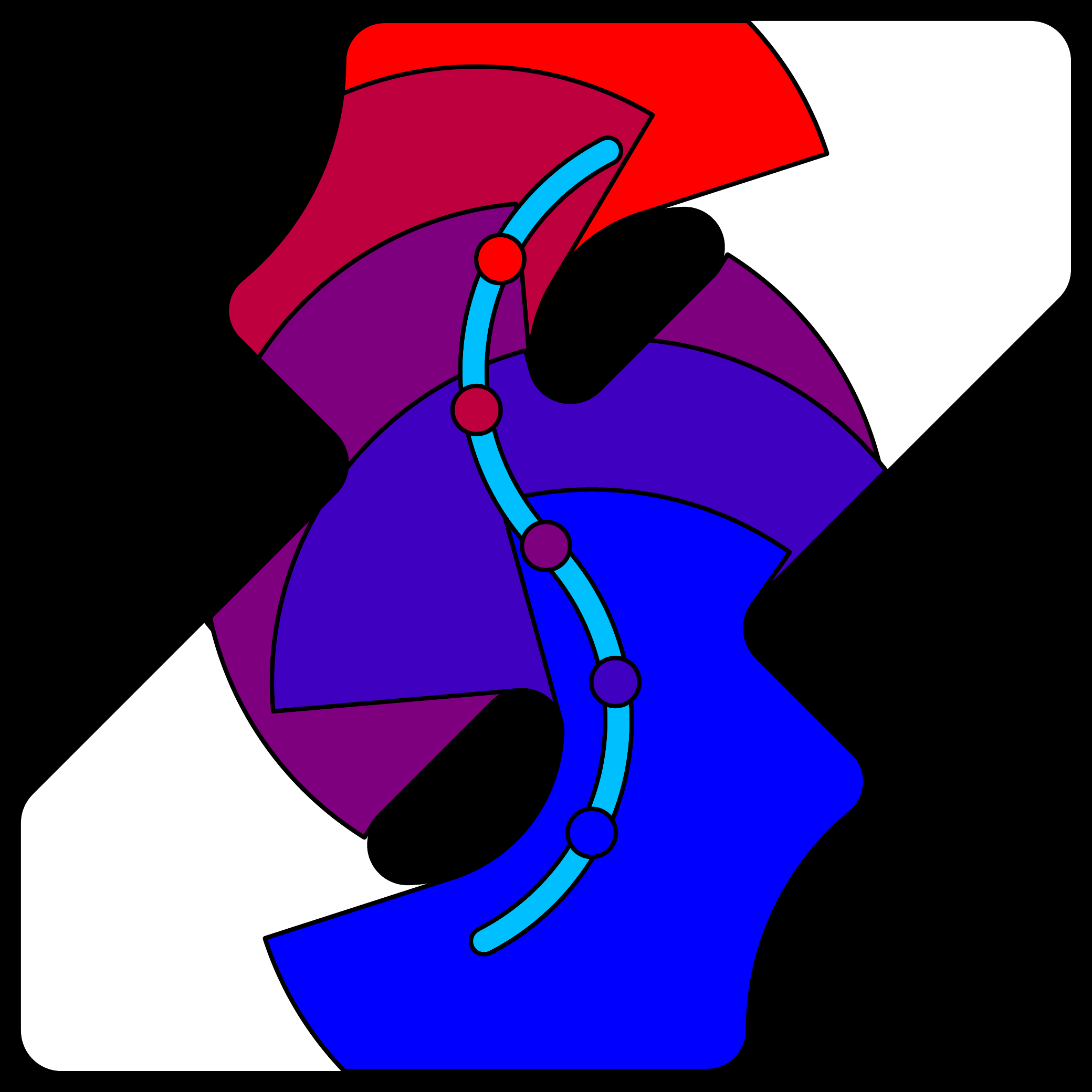}
            \caption{$\forall i \cop{:} \mathcal{W}_{\mathrm{new}}(s_{i} \cop{\mid} S_{i - 1})$}
            \label{fig:ets-new-visibility}
        \end{subfigure}
        \\\vspace{2pt}
        \begin{subfigure}[t]{0.32\columnwidth}
            \centering
            \includegraphics[width=\linewidth]{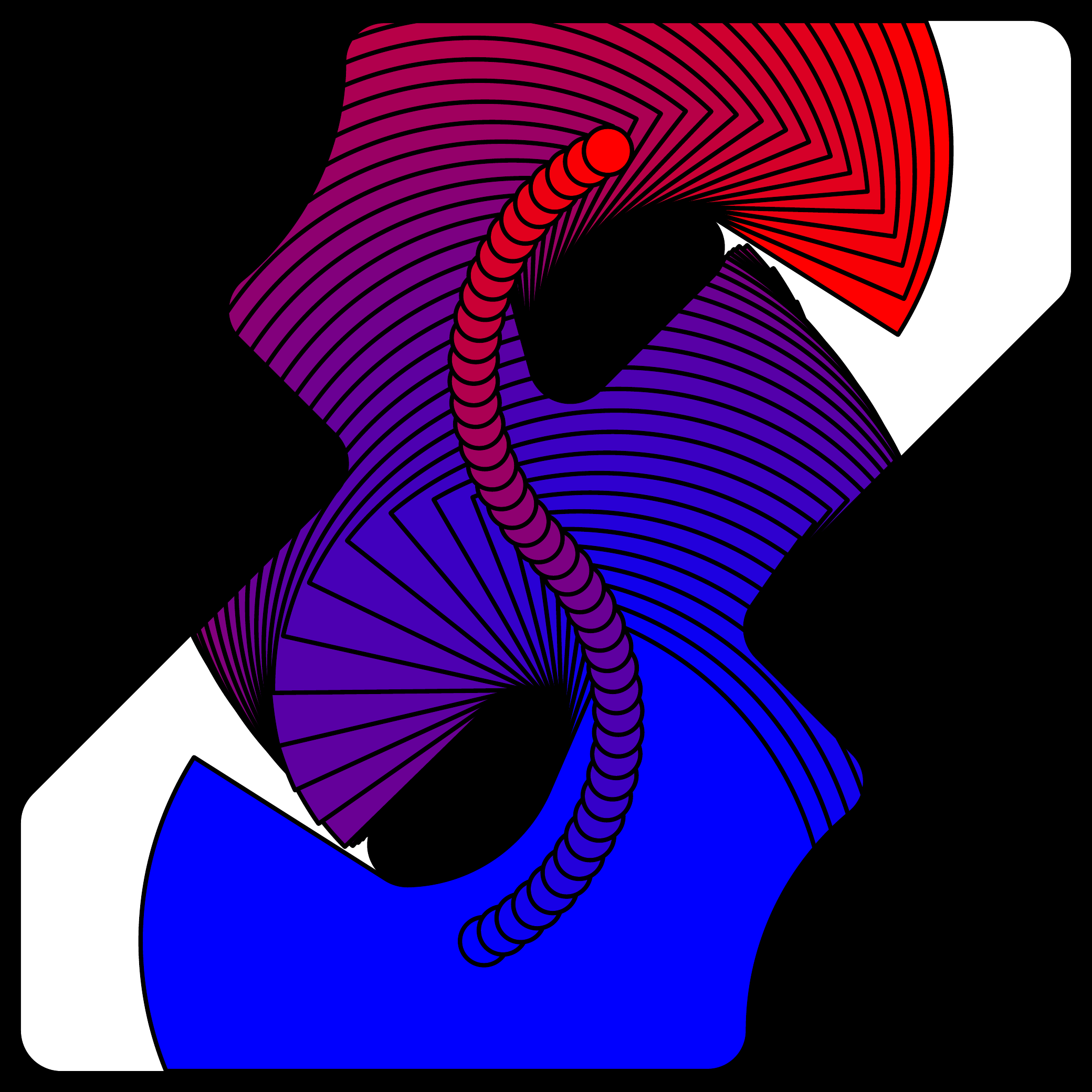}
            \caption{High-frequency sensing}
            \label{fig:ets-high-frequency}
        \end{subfigure}
        \hfill
        \begin{subfigure}[t]{0.32\columnwidth}
            \centering
            \includegraphics[width=\linewidth]{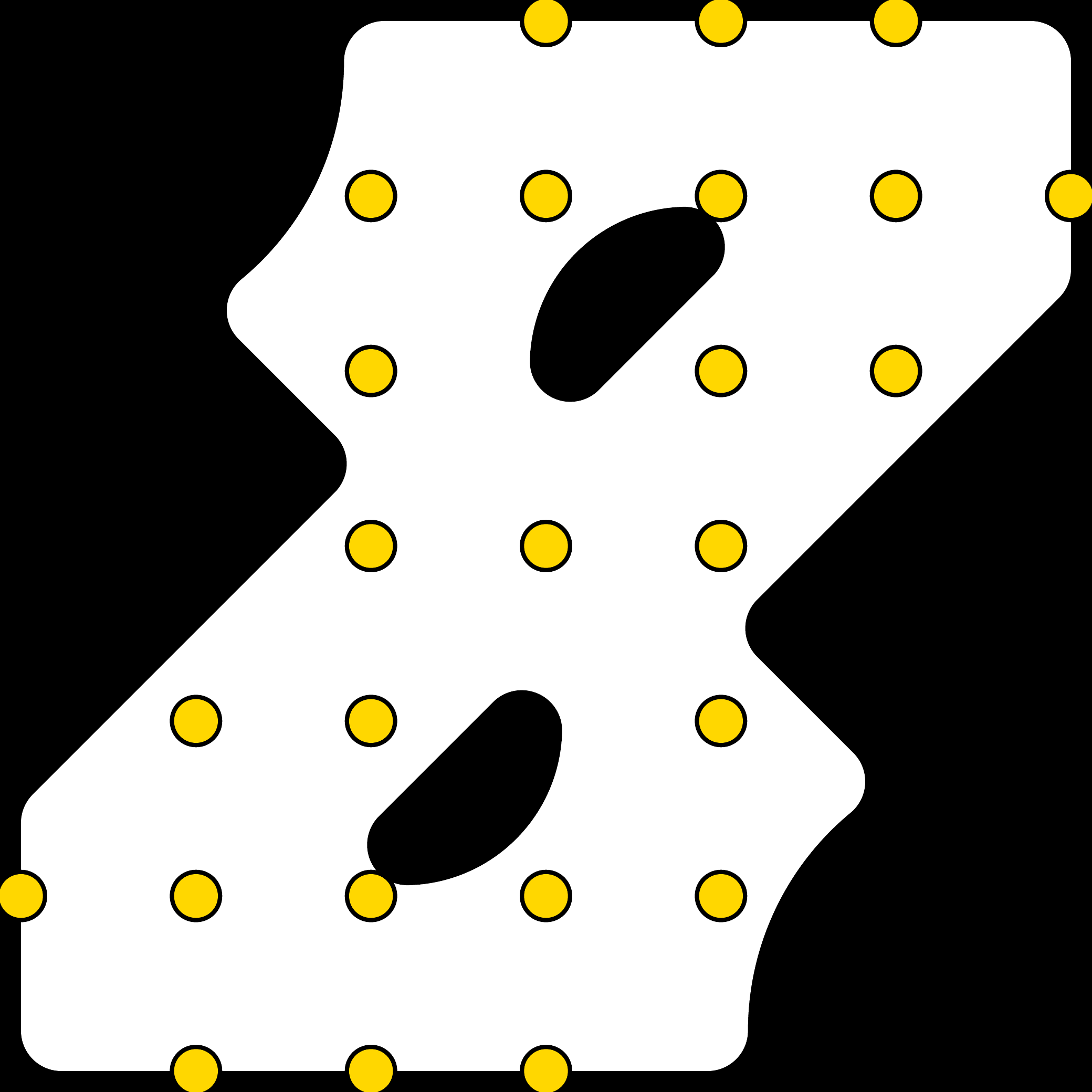}
            \caption{Guard set $G$}
            \label{fig:ets-guard-set}
        \end{subfigure}
        \hfill
        \begin{subfigure}[t]{0.32\columnwidth}
            \centering
            \includegraphics[width=\linewidth]{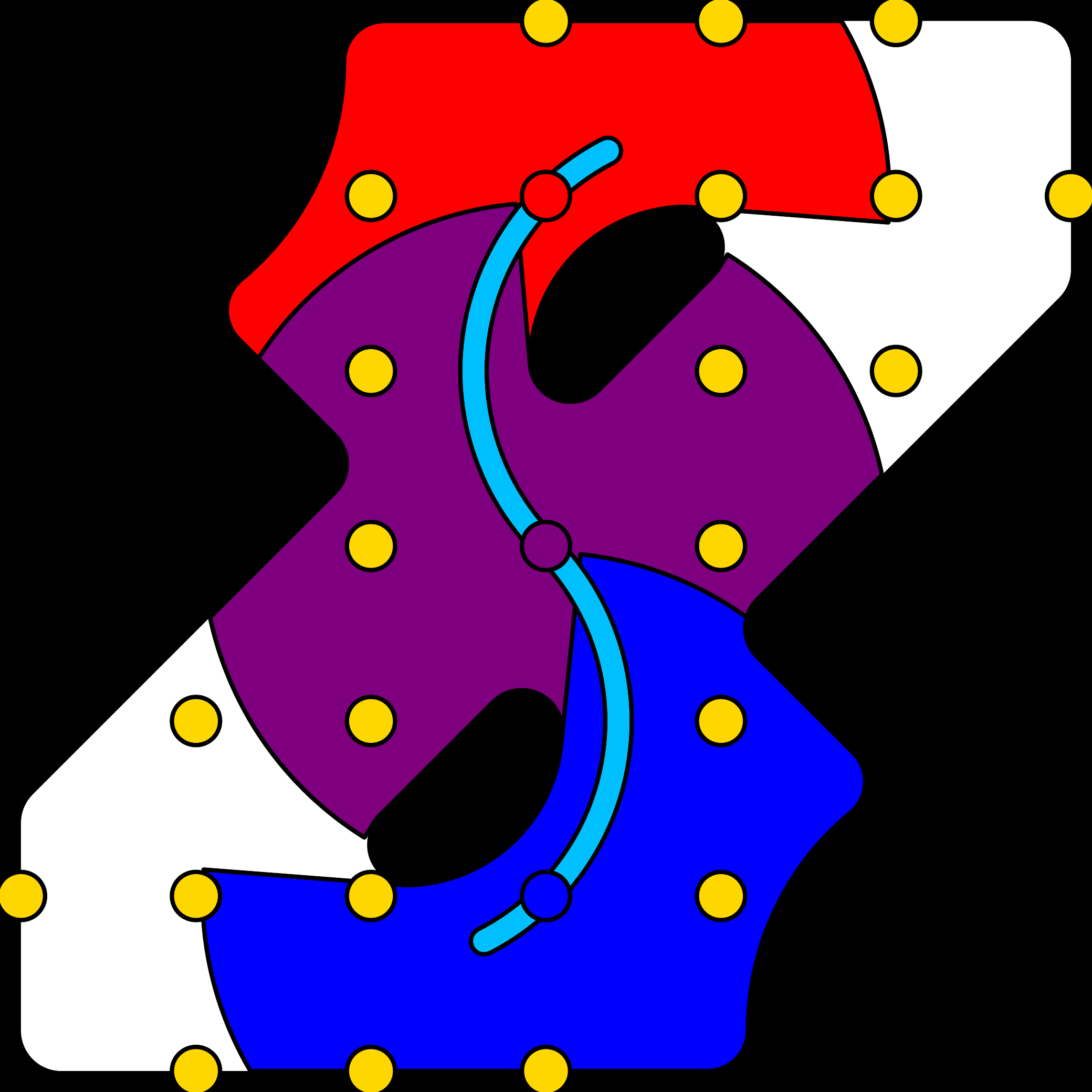}
            \caption{Sensing from $\tau$ s.t. $G$}
            \label{fig:ets-guarded-sensing}
        \end{subfigure}
        \caption{
            Illustrations of the tractable \search{} objective.
        }
        \label{fig:ets-illustrations}
    \end{figure}

    The sensing policy is a general rule set, possibly with additional inputs, such as a parameterized function or algorithm, that determines discrete sensing configurations along the path $\tau$.
    It outputs a pair $(\zeta, S_n) \cop{\gets} \mathrm{sens}(\tau)$, where: $\zeta \cop{:} \{0{:}n\} \cop{\mapsto} [0{,}1]$ is called the \emph{sensing mapping} and satisfies $\zeta(i) \cop{<} \zeta(j)$ for all $i, j \cop{\in} \{0{:}n\}$ such that $i \cop{<} j$; and $S_{i} \cop{\coloneqq} \langle s_{j} \cop{=} \tau(\zeta(j)) \rangle_{j=0}^{i}$ is the \emph{sensing sequence} up to the $i$-th configuration, where $i \cop{\in} \{0{:}n\}$, with $S_{n}$ representing the \emph{complete} sensing sequence.
    Fig.~\ref{fig:ets-path} illustrates an example path $\tau$ in the environment $\mathcal{W}$, while Fig.~\ref{fig:ets-sensing-sequence} shows an example sensing sequence $S_{n=4}$ for the path $\tau$, with a color gradient indicating the index ($\mathrm{blue}{=}0$, $\mathrm{red}{=}n$).

    Under an arbitrary sensing policy, the \search{} objective from Eq.~\eqref{eq:objective-cont} can be reformulated as
    \begin{flalign}
        \label{eq:objective}
        \objective{}(\tau, \mathrm{sens}) &= {\sum}_{i=0}^{n} \mathrm{Time}(\zeta(i) \mid \tau) p(s_{i} \mid S_{i - 1}),
    \end{flalign}
    where $p(s_{i} \cop{\mid} S_{i - 1})$ is the probability that the object is first detected at $s_{i}$, given the previously visited sensing configurations $S_{i - 1}$ (with $S_{-1} \cop{\coloneqq} \langle\rangle$).
    This probability is computed using the second assumption, which introduces a set of weighted target regions $ P \cop{=} \{(p_{i} \cop{\in} \mathbb{R}_{> 0},\allowbreak \mathcal{P}_{i} \cop{\subset} \mathcal{W})\}_{i=1}^m $.
    The probability that the object is located within any region $ \mathcal{X} \cop{\subset} \mathcal{W} $ is given by
    \begin{flalign*}
        \begin{split}
            p(\mathfrak{X} \in \mathcal{X} \mid P) &\coloneqq \frac{w(\mathcal{X} \mid P)}{\sum_{i=1}^m p_{i} \mathrm{Area}(\mathcal{P}_{i})},\\
            w(\mathcal{X} \mid P) &\coloneqq {\sum}_{i=1}^m p_{i} \mathrm{Area}(\mathcal{P}_{i} \cap \mathcal{X}).
        \end{split}
    \end{flalign*}
    Defining the newly sensed region at $s_{i}$ as
    \begin{flalign*}
        \begin{split}
            \mathcal{W}_{\mathrm{new}}(s_{i} \mid S_{i - 1}) &\coloneqq \mathrm{cl}(\mathrm{Vis}(s_{i}) \setminus \mathcal{W}_{\mathrm{seen}}(S_{i - 1})), \\
            \mathcal{W}_{\mathrm{seen}}(S) &\coloneqq {\bigcup}_{s \in S} \mathrm{Vis}(s),
        \end{split}
    \end{flalign*}
    we can express the detection probability at $s_{i}$ as
    \begin{flalign*}
        \begin{split}
            p(s_{i} \mid S_{i - 1}) &= p(\mathfrak{X} \in \mathcal{W}_{\mathrm{new}}(s_{i} \mid S_{i - 1}) \mid P),
        \end{split}
    \end{flalign*}
    and the constraint in Eq.~\eqref{eq:ets-constraint-path-end} can be rewritten as
    \begin{flalign}
        \label{eq:ets-constraint-path-end2}
        p(\mathfrak{X} \in \mathcal{W}_{\mathrm{seen}}(S_{n}) \mid P) \geq 1 - \epsilon.
    \end{flalign}
    Fig.~\ref{fig:ets-new-visibility} illustrates all regions $\mathcal{W}_{\mathrm{new}}(s_{i} \cop{\mid} S_{i - 1})$ for the example sensing sequence $S_{n=4}$, with colors corresponding to their respective sensing configurations.

    The objective in Eq.~\eqref{eq:objective} and the constraints in Eqs.~\eqref{eq:ets-constraint-path-start},~\eqref{eq:ets-constraint-path}, and~\eqref{eq:ets-constraint-path-end2} define the \emph{\search{} problem with a tractable objective}, which will be the \emph{default} \search{} formulation used throughout the rest of the paper.
    It is important to emphasize that the \search{} problem is not restricted to a specific sensing policy; various sensing strategies can be incorporated into the formulation, such as equidistant or equitemporal sensing.
    Fig.~\ref{fig:ets-high-frequency} illustrates high-frequency equidistant sensing along a path, which effectively approximates continuous sensing in practice.

    Finally, we define \dsearch{} as a special case of \search{}.
    \dsearch{} restricts sensing to a predefined set of configurations, $G \cop{=} \{g_{i} \cop{\in} \mathbb{C}_{\mathrm{free}}\}_{i=0}^{n}$ (including the start configuration $g_{0}$), satisfying $p(\mathfrak{X} \cop{\in} \mathcal{W}_{\mathrm{seen}}(G) \cop{\mid} P) \cop{\geq} 1 \cop{-} \epsilon$.
    The set $G$, called the \emph{guard set}, consists of so-called \emph{guards}, which the \dsearch{} solution typically visits (assuming they are path-connected in the collision-free space) to satisfy all \search{} constraints.
    The \dsearch{} objective follows a specific sensing policy $\mathrm{sens}(\tau) \cop{\coloneqq} \mathrm{\dsearch{}}(\tau \cop{,} G)$, defined by $G$ as:
    \emph{``Sensing along the path $\tau$ occurs exclusively at the first visit to each guard in $G$.''}
    Fig.~\ref{fig:ets-guard-set} illustrates a guard set $G$, while Fig.~\ref{fig:ets-guarded-sensing} depicts sensing along the example path $\tau$ constrained by $G$.
    Note that sensing occurs only where $\tau$ exactly intersects the guard set.

    Notably, the \emph{classical} version of \dsearch{} (\dclassical{}) has been proven NP-hard~\citep{Sarmiento2003}.
    Since \dsearch{} and \search{} generalize it, they are at least as hard to solve optimally.



    \section{Related Work}
    \label{sec:related-work}

    \subsection{Broad Introduction to Search Strategies in S\&R and Beyond}

    While effective search strategies may appear to be a crucial component of S\&R operations, the optimization of expected-time search routes has received {limited} attention in the related literature.
    Instead, research efforts have predominantly focused on the development of robotic systems for hazardous environments~\citep{Zhao2017}, probabilistic modeling of moving targets~\citep{Heintzman2021}, analyzing camera images for target detection~\citep{Niedzielski2021}, and addressing other practical challenges.
    In practice, search strategies in S\&R rarely incorporate global route optimization.
    Rather, they often rely on {predefined} area coverage patterns~\citep{Queralta2020}, focus on {local} objectives such as minimizing collision risks and complementing human search efforts~\citep{Heintzman2021}, or employ simple predefined or reactive behaviors that give rise to complex {emergent} behaviors in multi-robot systems~\citep{Arnold2018}.
    Furthermore, S\&R scenarios involving autonomous agents often rely on {unmanned aerial vehicles} (UAVs), which typically operate outdoors at {high} altitudes~\citep{Cubber2017,Queralta2020,Heintzman2021,Niedzielski2021,Arnold2018}.
    As a result, visibility constraints and the time required to navigate around obstacles---critical factors in our formulation---are less relevant in these scenarios.

    S\&R operations, however, are {not} the only domain where search strategies are relevant.
    Consider how often people search for everyday items such as keys, wallets, or phones.
    Some professions are even dedicated to searching for specific objects, such as uncovering evidence or identifying threats.
    In many of these scenarios, both the environment and the object of the search are sufficiently well-known to allow us, in principle, to model the problem and {optimize} the expected time to exclaim, \emph{``Eureka!''}.
    With advances in robotics, we can now delegate these search tasks to autonomous mobile robots, which are capable of navigating known environments, particularly {large} ones, more efficiently than humans.

    \subsection{Foundational Contributions to \search{}}
    \label{subsec:foundational-contributions}

    Sarmiento~\etal{}~\citep{Sarmiento2003,Sarmiento2004,Sarmiento2004c,Sarmiento2009} made a seminal contribution to the study of \search{} by being the first to examine the expected-time objective in comparison to the worst-case objective in the context of search route optimization, thereby introducing a transformative {search paradigm}.
    Their work investigated a holonomic mobile sensor with unlimited range and omnidirectional visibility, modeled as a point navigating a polygonal environment with obstacles (\classical{} scenario).
    The object's location was assumed to follow a uniform probability distribution throughout the environment, consistent with the \emph{principle of indifference}~\citep{Williamson2018}, which applies when no prior information about the object's likely location is available.

    At first glance, one might assume that all sensing configurations are equivalent under a uniform distribution, leading to the belief that the worst-case and expected-time objectives would yield similar outcomes.
    However, this assumption is {fundamentally flawed}.
    Sensing configurations vary significantly due to {visibility constraints}: open areas, where larger regions can be sensed, provide a higher likelihood of detecting the object, whereas narrow corridors restrict visibility, reducing the chances of success.
    Furthermore, sensed regions from different configurations often overlap, introducing {interdependencies} between probabilities.
    These factors collectively contribute to the complexity of the expected-time objective, which cannot be directly reduced to the worst-case objective~\citep{Sarmiento2003}.
    The situation is, in fact, even more nuanced than suggested by~\citep{Sarmiento2003}.
    In a complete graph where edges are weighted by symmetric travel times and nodes have {uniform}, {independent} probabilities, the expected-time objective {still} does {not} reduce to the worst-case objective.
    While the worst-case objective involves finding the shortest Hamiltonian path (i.e., the TSP objective~\citep{Cook2015}), the \search{} problem reduces to the \emph{traveling deliveryman problem} (TDP)~\citep{Lucena1990}, which minimizes the sum of delivery latencies rather than the total travel time~\citep{Mikula2022}.

    In their original work~\citep{Sarmiento2003}, Sarmiento~\etal{} addressed the \search{} variant, where sensing is limited to a predefined set of sensing configurations within an otherwise continuous polygonal environment (\dclassical{}).
    They demonstrated that the problem is NP-hard and, due to its complexity, proposed a heuristic utility-based greedy algorithm.
    The utility function, maximized at each step, combines the time required to reach a sensing configuration with the probability of detecting the object there, expressed as $w(v) \cop{/} d(u, v)$, where $w(v)$ represents the area {newly} visible from $v$ and $d(u, v)$ is the {shortest-path distance} between $u$ and $v$.
    The greedy algorithm is further enhanced by incorporating a {lookahead} strategy that examines several steps ahead within a {reduced} search\footnote{
        It is worth noting that, in the context of optimization, the term \emph{search} has a second meaning, referring to the process of exploring the solution or partial solution space to identify feasible or high-quality solutions.
        This should not be confused with the concept of searching for a target object, as in \search{}.
    } tree, with the lookahead depth {adaptively} determined based on the tree's average branching factor.
    In~\citep{Sarmiento2004c}, this approach was extended to the multi-robot case, and in~\citep{Sarmiento2004,Sarmiento2009}, they addressed continuous sensing by proposing a method for constructing locally optimal search paths within a simple polygon.

    Although foundational, the work of Sarmiento~\etal{} has notable shortcomings.
    First, it includes only a limited computational evaluation, providing insufficient evidence for the effectiveness of the proposed heuristic.
    Notably, the experiments were conducted on a few small instances, and the evaluated lookahead strategy was not even compared to the single-step version.
    Second, the authors did not address how to optimize the search globally, beyond the scope of a greedy approach.
    In contrast, the operations research (OR) community has studied other NP-hard optimization problems for decades, leading to the development of metaheuristic algorithms that typically outperform greedy methods in practice~\citep{Gendreau2010}, while simultaneously offering a well-balanced trade-off between solution quality and runtime.

    \subsection{Metaheuristic Approaches for \search{}}
    \label{subsec:metaheuristic-approaches}

    The first attempt to address global route optimization in \search{} metaheuristically was made by Kulich~\etal{}~\citep{Kulich2014,Kulich2017,Kulich2022}, who proposed modeling the \search{} objective using a simplified version based on the TDP~\citep{Lucena1990} and later its generalization, the \emph{graph search problem} (GSP)~\citep{Koutsoupias1996}.
    This approach assumes that the probabilities of detecting the object at different sensing configurations are either {independent and uniform} (TDP)~\citep{Kulich2014} or independent and proportional to the area visible from each configuration (GSP)~\citep{Kulich2017,Kulich2022}.
    Although these assumptions do not generally hold, the simplified objectives enable efficient exploration of the solution space and can still yield high-quality solutions due to the close relationship between the TDP, GSP, and \search{} objectives.
    Since the GSP, like \search{}, has received limited attention in the literature, the authors proposed the first tailored metaheuristic algorithm for the GSP in~\citep{Kulich2017}, based on the \emph{greedy randomized adaptive search procedure} (GRASP)~\citep{Resende2016}, a general framework that combines local search heuristics with randomized construction heuristics and {multiple} restarts.
    The effectiveness of the proposed approach has been demonstrated in two scenarios: the exploration of an unknown environment while simultaneously searching for a target object~\citep{Kulich2014,Kulich2017}, and search in a known environment involving multiple agents, including a real-world robotic experiment~\citep{Kulich2022}.

    Another related work~\citep{Mikula2022} focused on the TDP\@.
    A key advantage of the TDP over the GSP is that the TDP has been extensively studied by the OR community~\citep{Lucena1990,Silva2012,Mladenovic2013}, resulting in well-established metaheuristic algorithms available as benchmarks.
    While these metaheuristics are designed to produce high-quality solutions within a \emph{``reasonable''} amount of time, they typically require tens of minutes or even hours of runtime for instances ranging from 500 to 1,000 nodes~\citep{Silva2012}.
    Although these runtime values may seem high, they represent a significant improvement over exact algorithms, which scale exponentially with problem size.
    By contrast, the authors of~\citep{Mikula2022} focused on developing a metaheuristic for the TDP capable of delivering high-quality solutions within a computational time budget of {just 1 to 100 seconds} on instances with up to {1,000 nodes}.
    This approach aligns with the context of \search{}, where runtime constraints are often critical in practical applications.
    The best version of the metaheuristic, called \emph{multi-start generalized variable neighborhood search} (Ms-GVNS), demonstrated significant improvements over the state-of-the-art TDP metaheuristic at the time~\citep{Silva2012} under identical runtime constraints.
    Among its favorable properties, Ms-GVNS is an anytime algorithm, meaning it can be stopped at any time and return the best solution found so far.

    Ms-GVNS was recently extended to address the GSP in an application-driven scenario involving a real robotic system designed to monitor the ecosystem of a honeybee hive~\citep{Ulrich2024}.
    The algorithm was tailored to optimize the GSP objective and was employed for the specific subtask of locating the {queen bee}, a critical step in studying the hive's overall behavior~\citep{Blaha2024}.

    \subsection{Broader Literature on \search{}}

    A substantial body of research examines variants of the \search{} problem, emphasizing {different} aspects of the search scenario \citep[e.g.,][]{Lau2007,Chung2012,PerezCarabaza2021,Huynh2024}.
    Notably, visibility constraints, collision-free pathfinding, and travel time modeling---key elements of our formulation---are often {simplified} or {omitted} entirely, with some works prioritizing the worst-case objective over the expected-time objective.
    Furthermore, many existing works are limited to discrete, grid-based environments, whereas our approach is applicable to both continuous and discrete environments.
    On the other hand, many of these remotely related studies focus heavily on {uncertainty} and {dynamic} aspects of the search, which are only {marginally} addressed in our formulation.
    Our approach incorporates a general {probability} distribution for the searched object, focuses on the {expected value} of the search until the object is found, and balances solution quality with runtime, enabling {replanning} at low to moderate frequencies.
    However, {highly} dynamic or uncertain environments are {not} the primary focus of our work.

    The following works are representative but not exhaustive of the broader literature on \search{}:
    \citep{Lau2007}~addressed \search{} in highly structured environments, where the highest-resolution version of the problem involved ordering predefined regions (rooms) in an indoor environment to determine the optimal visiting sequence.
    However, unlike our formulation, the authors did not explicitly consider obstacles or a visibility model.
    \citep{Chung2012}~studied a probabilistic, grid-based search scenario where the object is either present or absent in the environment.
    The work optimizes the expected time to reach a conclusion about the object's presence using an imperfect sensor capable of inspecting at most one cell at a time.
    Such a scenario is clearly not compatible with our formulation.
    \citep{PerezCarabaza2021}~examined a probabilistic search problem with the \search{} objective, where a group of UAVs searches for multiple targets, and all agents are modeled probabilistically on a grid.
    Since the scenario assumes search operations at high altitudes over large areas, the environment is treated as an open space, unlike our formulation.
    Lastly,~\citep{Huynh2024}~represents a recent theoretical contribution to visibility-based search in polygonal environments.
    Instead of optimizing the expected time to detect the target, the authors focus on other objectives, such as minimizing route length to cover a specific percentage of the environment or maximizing coverage under a route-length constraint.

    \subsection{Positioning This Work Within Related Research}

    In this work, we build upon previous research~\citep{Kulich2014,Kulich2017,Kulich2022,Mikula2022,Blaha2024} and introduce several key contributions to the metaheuristic optimization of the \search{} objective:
    \begin{enumerate*}[label=(\arabic*)]
        \item We \emph{refine} the approximation from~\citep{Kulich2017}, which links detection probability to static weights based on visible area, and propose three novel alternatives for a more accurate \search{} objective approximation.
        \item We \emph{extend} the GSP objective by introducing a new triplet-based cost for sensor turning, leading to the novel \gspt{} objective.
        \item We \emph{adapt} Ms-GVNS~\cite{Mikula2022} to optimize \gspt{}, leveraging efficient local search operators and building on prior works~\citep{Kulich2022,Mikula2022,Blaha2024}.
        \item We \emph{introduce} a replanning scheme for \gspt{} to enhance solution quality for \dsearch{} and \search{} when static weights are significantly inaccurate.
        \item We \emph{replace} the uniform probability distribution used in prior works with a general distribution model, broadening the framework’s applicability.
        \item We \emph{create} the first large-scale \dsearch{} dataset, comprising 240 instances with up to 2,000 guards, divided into 16 subsets based on guard count and visibility overlap.
        \item We \emph{conduct} a comprehensive computational evaluation on \dclassical{} and \classical{} scenarios, comparing our framework to single-step and adaptive-depth utility greedy heuristics~\citep{Sarmiento2003} and prior metaheuristic approaches~\citep{Kulich2014,Kulich2017}.
        \item Finally, we \emph{demonstrate} the framework’s flexibility through a qualitative study on diverse \search{} scenarios.
    \end{enumerate*}

    \section{Proposed Solution Framework}
    \label{sec:proposed-solution}

    \subsection{Decoupling Scheme for \search{}}
    \label{subsec:decoupling-framework}

    The proposed solution framework for the \search{} problem constructs a discretized solution space where all paths satisfy the problem constraints.
    Although optimality cannot be guaranteed within this discretized space, this is not critical since the focus lies on empirical performance and scalability over strict optimality.
    The framework employs a standard decoupling approach (Alg.~\ref{alg:decoupling}), splitting \search{} into two subproblems: \emph{the sensor placement problem} (SPP) and \dsearch{}.

    \begin{algorithm}
        \caption{Decoupling Scheme for \search{}}
        \label{alg:decoupling}
        \begin{algorithmic}[1]%
            \Require Environment $\mathcal{W}$, weighted target regions $P$, initial configuration $g_{0}$, visibility model $\mathrm{Vis}$, \search{} sensing policy $\mathrm{sens}$, $\epsilon \cop{\in} [0{,}1]$.
            \Ensure \search{} solution path $\tau_{\,\text{\search{}}}$.
            \State \textbf{Generate} a guard set $G \cop{=} \{g_{i} \cop{\in} \mathbb{C}_{\mathrm{free}}\}_{i=0}^{n}$ such that:
            \begin{itemize}
                \item $p(\mathfrak{X} \cop{\in} \mathcal{W}_{\mathrm{seen}}(G) \cop{\mid} P) \cop{\geq} 1 \cop{-} \epsilon$.
                \item For all pairs $i, j \cop{\in} \{0{:}n\}$, a collision-free path $\tau_{i,j}$ exists such that $\tau_{i,j}(0) \cop{=} g_{i}$, $\tau_{i,j}(1) \cop{=} g_{j}$.
            \end{itemize}\label{alg:decoupling:spp}
            \State \textbf{Solve} \dsearch{} s.t. $G$ using the \milaps{} method from Alg.~\ref{alg:milaps}, recording all intermediate solutions $\Upsilon \cop{=} \langle \tau_{k} \rangle_{k=1}^{z}$.
            Alternatively, use \milaps{} with replanning from Alg.~\ref{alg:milaps-replanning}.\label{alg:decoupling:detms}
            \State $\tau_{\,\text{\search{}}} \gets \tau_{z}$\label{alg:decoupling:tauz}
            \State \textit{(optional)} $\tau_{\,\text{\search{}}} \gets \argmin_{\tau_{k} \in \Upsilon} \objective{}(\tau_{k}, \mathrm{sens})$.\label{alg:decoupling:final}
            \State \Return $\tau_{\,\text{\search{}}}$ \label{alg:decoupling:return}
        \end{algorithmic}
    \end{algorithm}

    Defined in line~\ref{alg:decoupling:spp} of Alg.~\ref{alg:decoupling}, the SPP constructs the discretized solution space for \search{} by generating a guard set $G$ that ensures a detection probability of at least $(1 \cop{-} \epsilon)$ and collision-free paths between guards.
    However, this may not always be feasible in arbitrary environments, especially with large sensor footprints or small visibility ranges, as some regions may be inaccessible to the sensor's view or disconnected in $\mathbb{C}_{\mathrm{free}}$.
    This poses a challenge in scenarios where solving \search{} is critical, and any feasible search strategy is preferable to none.
    In such cases, we propose relaxing the probability constraint by increasing $\epsilon$ or restricting the search to a subset of the environment with looser constraints.

    A recent study~\citep{Mikula2024b} provides a detailed analysis of omnidirectional SPP in continuous (polygonal) environments, comparing several heuristic methods.
    The evaluated approaches include \emph{sampling-based methods}~\citep[e.g.,][]{Gonzalez-Banos1998}, \emph{convex-partitioning techniques}~\citep[e.g.,][]{Kazazakis2002}, and novel \emph{hybrid accelerated-refinement} (HAR) heuristics, which combine and refine outputs from multiple methods.
    The study shows that HAR achieves the lowest guard count among all tested approaches while guaranteeing feasibility for path-connected environments and point sensors.
    We adopt the HAR-KA,RV variant, as it offers the best balance between guard count and runtime.
    Here, KA,RV denotes the combination of the Kazazakis and Argyros (KA) method from~\citep{Kazazakis2002} and a reflex vertex (RV) heuristic, which includes all reflex vertices of the environment in the guard set.
    Since the original method assumes point sensors, we adapt it for a nonzero footprint radius, though without feasibility guarantees, by replacing the RV method with the inclusion of all vertices on the boundary of $\mathbb{C}_{\mathrm{free}}$ and applying the KA method twice—once on $\mathcal{W}$ with $r_{\mathrm{vis}}$ and once on $\mathbb{C}_{\mathrm{free}}$ with a reduced visibility radius of $r_{\mathrm{vis}} \cop{-} r_{\mathrm{fp}}$.
    The refinement procedure then filters out guards that are unreachable from $g_0$ or redundant.

    The second decoupled problem, \dsearch{}, is addressed in detail in the following sections.
    After solving it at line~\ref{alg:decoupling:detms}, the decoupling scheme accounts for the differences between the \dsearch{} and \search{} sensing policies.
    It assumes a sequence of solutions $\Upsilon \cop{=} \langle \tau_{k} \rangle_{k=1}^{z}$, representing the best and intermediate solutions obtained during \dsearch{} optimization.
    These are evaluated against the \search{} objective, and the best one is selected (line~\ref{alg:decoupling:final}).
    Alternatively, the last solution $\tau_z$ can be returned directly to reduce computational cost (bypassing line~\ref{alg:decoupling:final}).

    The feasibility of the decoupling scheme depends on the SPP method, as discussed earlier, and the alignment between the \search{} and \dsearch{} sensing policies.
    Since the decoupling scheme ensures that only sensing from the guards $G$ satisfies the probabilistic coverage constraint in Eq.~\eqref{eq:ets-constraint-path-end2}, the \search{} policy $\mathrm{sens}$ must also guarantee sensing at $G$ to inherit this property.
    Consider a counterexample where a specific guard is the only one capable of sensing a certain region.
    If the \search{} policy does not sense at this guard---despite possibly sensing nearby---the probabilistic coverage constraint may be violated.
    This issue can be mitigated in practice by employing high-frequency sensing to minimize these misalignment effects.

    \subsection{\milaps{} for \dsearch{}}
    \label{subsec:milaps}

    This section introduces the \milaps{} framework for solving \dsearch{} by heuristically approximating it as a generalized graph search problem, \gspt{}, defined on a fixed graph with static costs and weights.
    \milaps{} transforms a \dsearch{} instance into \gspt{}, solves it using an adapted Ms-GVNS metaheuristic from~\citep{Mikula2022}, and maps the solution back to \dsearch{} to obtain a feasible route.
    As the transformation between \dsearch{} and \gspt{} is heuristic and not uniquely defined---particularly in the choice of \gspt{}'s static weights---\milaps{} provides multiple alternatives for this step.
    We first define \gspt{} and then present \milaps{} as the bridge between \dsearch{} and \gspt{}.

    \subsubsection{\gspt{} Definition}

    The \emph{graph search problem with turning} (GSPT) is defined on a complete graph $\mathbb{G} \cop{=} (V,\allowbreak{}v_{\mathrm{s}},\allowbreak{}w,\allowbreak{}\vartheta,\allowbreak{}d,\allowbreak{}\theta)$, where $V$ is the set of $n \cop{+} 1$ vertices, $v_{\mathrm{s}} \cop{\in} V$ is the starting vertex, $w \cop{:} V \cop{\mapsto} \mathbb{R}_{>0}$ is a 1D weight function, $\vartheta \cop{:} V \cop{\mapsto} \mathbb{R}_{\geq 0}$ is a 1D cost function, $d \cop{:} V \cop{\times} V \cop{\mapsto} \mathbb{R}_{\geq 0}$ is a 2D cost function, and $\theta \cop{:} V \cop{\times} V \cop{\times} V \cop{\mapsto} \mathbb{R}_{\geq 0}$ is a 3D cost function.
    The goal is to find a permutation $\sigma \cop{:} \{0{:}n\} \cop{\mapsto} V$ that starts at $v_{\mathrm{s}}$, i.e., $\sigma(0) \cop{=} v_{\mathrm{s}}$, and minimizes the objective:
    \begin{flalign}
        \label{eq:gspt-objective}
        \argmin_{\sigma} \quad {\sum}_{i=1}^{n} \delta(i \mid \sigma) w(\sigma(i)),
    \end{flalign}
    where $\delta(i \cop{\mid} \sigma)$ represents the \emph{latency} of the $i$-th vertex in the permutation ($\delta(. \cop{\mid} \sigma) \cop{:} \{1{:}n\} \cop{\mapsto} \mathbb{R}_{\geq 0}$), and is defined as:
    \begin{flalign*}
        \delta(i \mid \sigma) &\coloneqq {\sum}_{j=1}^{i} \theta(j \mid \sigma) + d(\sigma(j - 1), \sigma(j)), \\
        \theta(i \mid \sigma) &\coloneqq
        \begin{cases}
            \vartheta(\sigma(1)), \text{ if } i = 1, \\
            \theta(\sigma(i-2), \sigma(i-1), \sigma(i)), \text{ otherwise}.
        \end{cases}
    \end{flalign*}
    A \gspt{} instance is called \emph{symmetric} if the 2D and 3D costs are symmetric with respect to the vertex order, i.e., $d(u, v) \cop{=} d(v, u)$ and $\theta(h, u, v) \cop{=} \theta(v, u, h)$ for all $h, u, v \cop{\in} V$.
    When all $\vartheta$ values and $\theta$ values are {zero}, the problem reduces to the GSP~\citep{Koutsoupias1996}, and when additionally all $w$ values are {constant}, the problem reduces to the TDP~\citep{Lucena1990}.




    \subsubsection{Bridging \gspt{} and \dsearch{}}

\begin{figure}
        \centering
        \begin{subfigure}[t]{0.32\columnwidth}
            \centering
            \includegraphics[width=\linewidth]{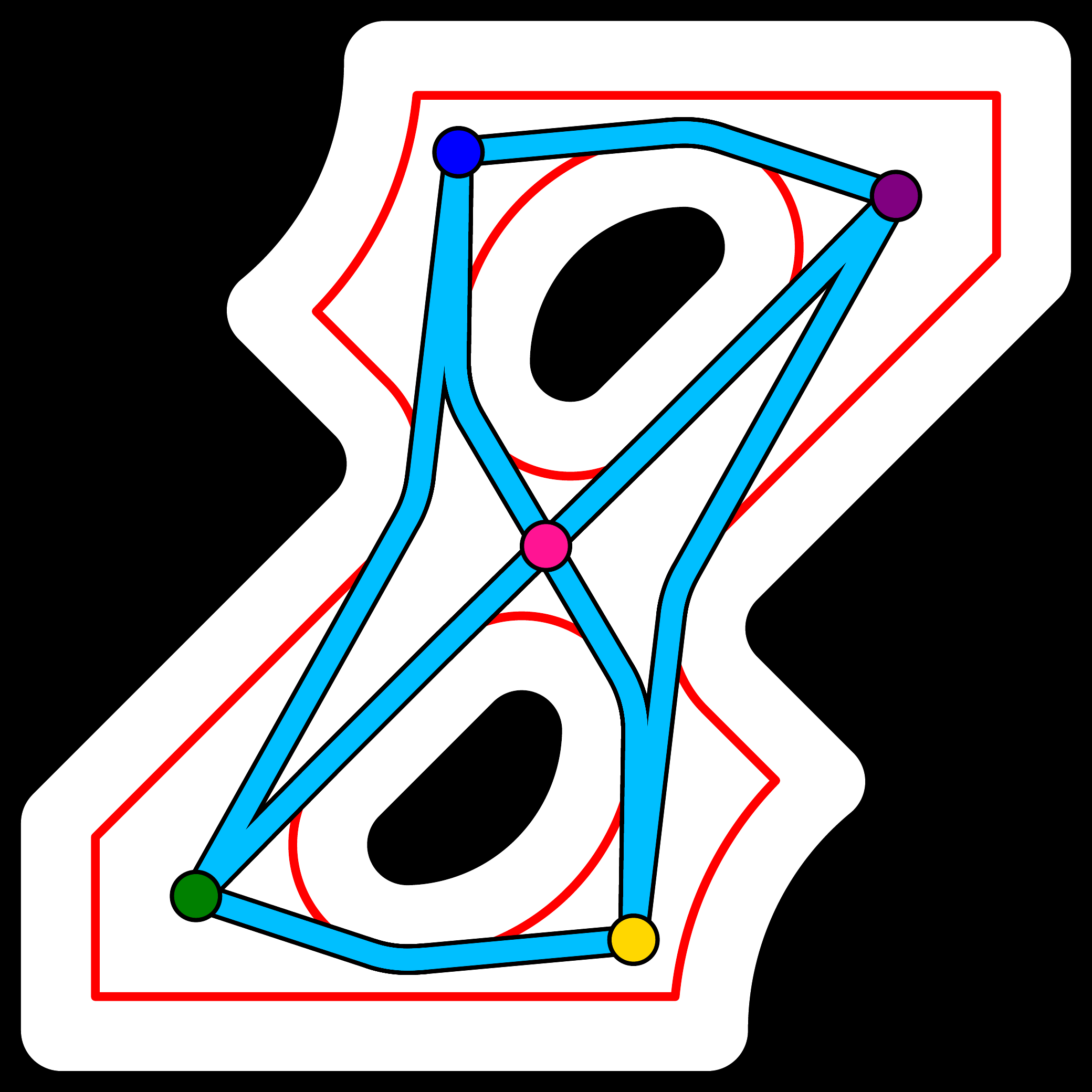}
            \caption{Graph $\mathbb{G}$}
            \label{fig:graph}
        \end{subfigure}
        \hfill
        \begin{subfigure}[t]{0.32\columnwidth}
            \centering
            \includegraphics[width=\linewidth]{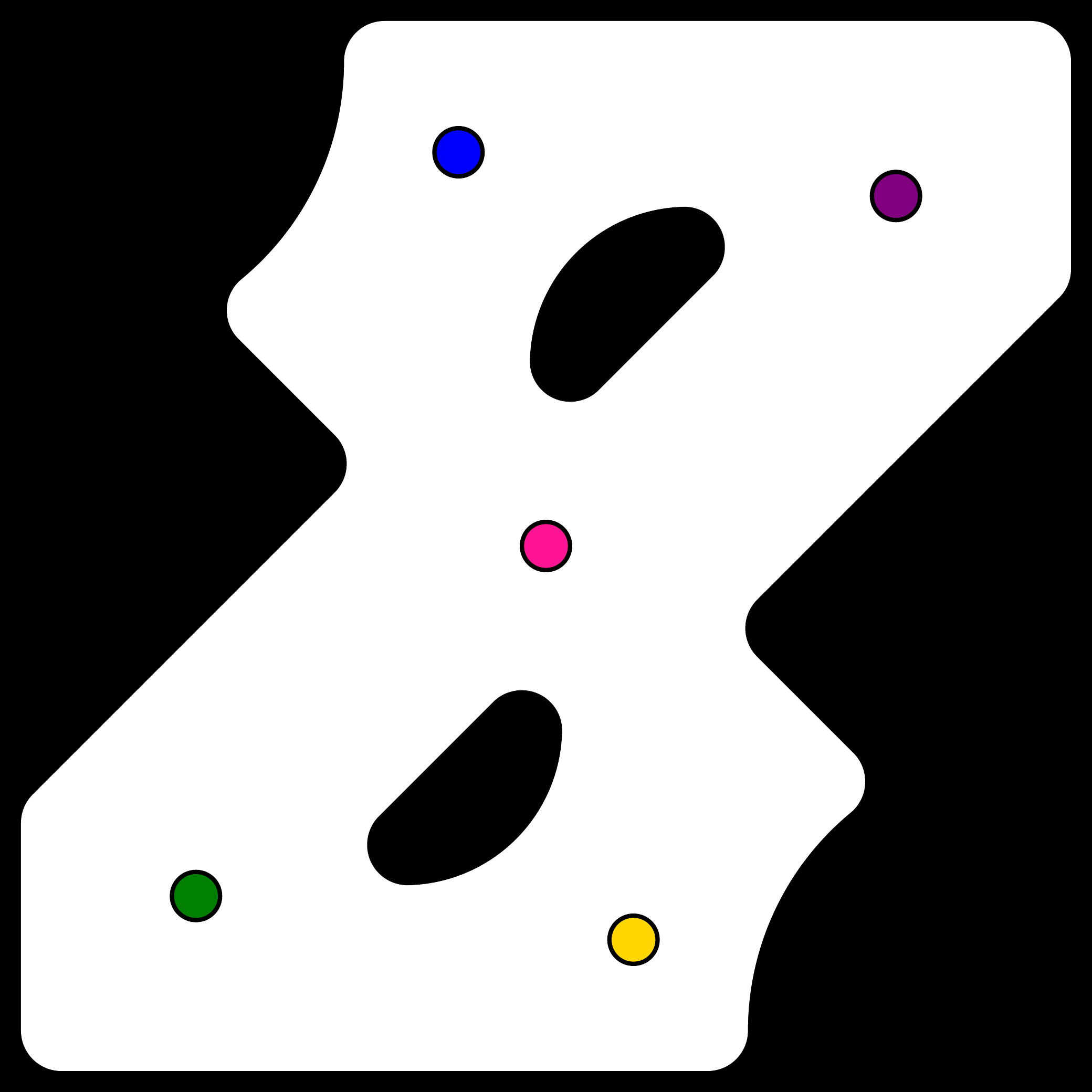}
            \caption{$\mathrm{wtype} \cop{=} \text{Const}$}
            \label{fig:w1}
        \end{subfigure}
        \hfill
        \begin{subfigure}[t]{0.32\columnwidth}
            \centering
            \includegraphics[width=\linewidth]{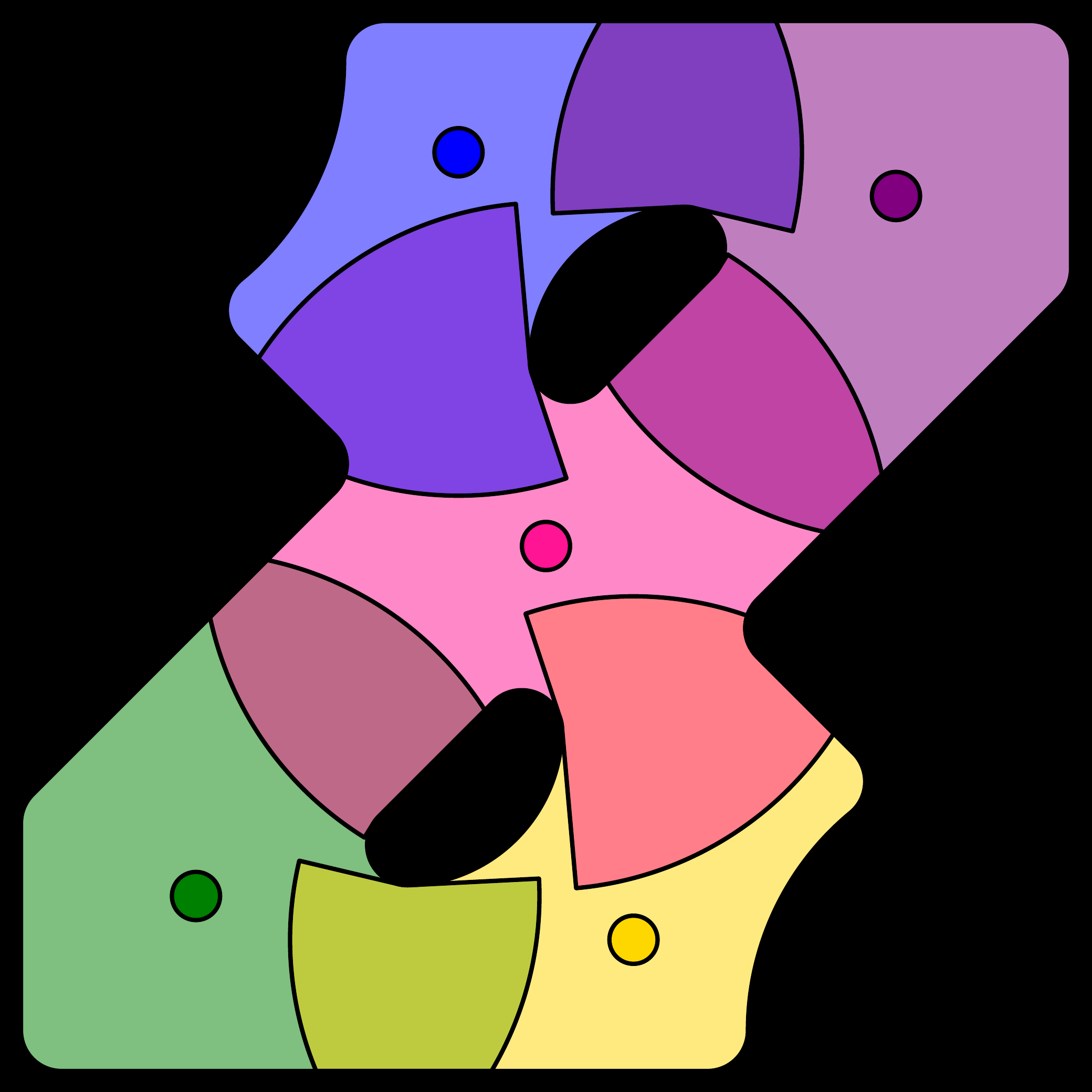}
            \caption{$\mathrm{wtype} \cop{=} \text{Vis}$}
            \label{fig:w2}
        \end{subfigure}
        \\\vspace{2pt}
        \begin{subfigure}[t]{0.32\columnwidth}
            \centering
            \includegraphics[width=\linewidth]{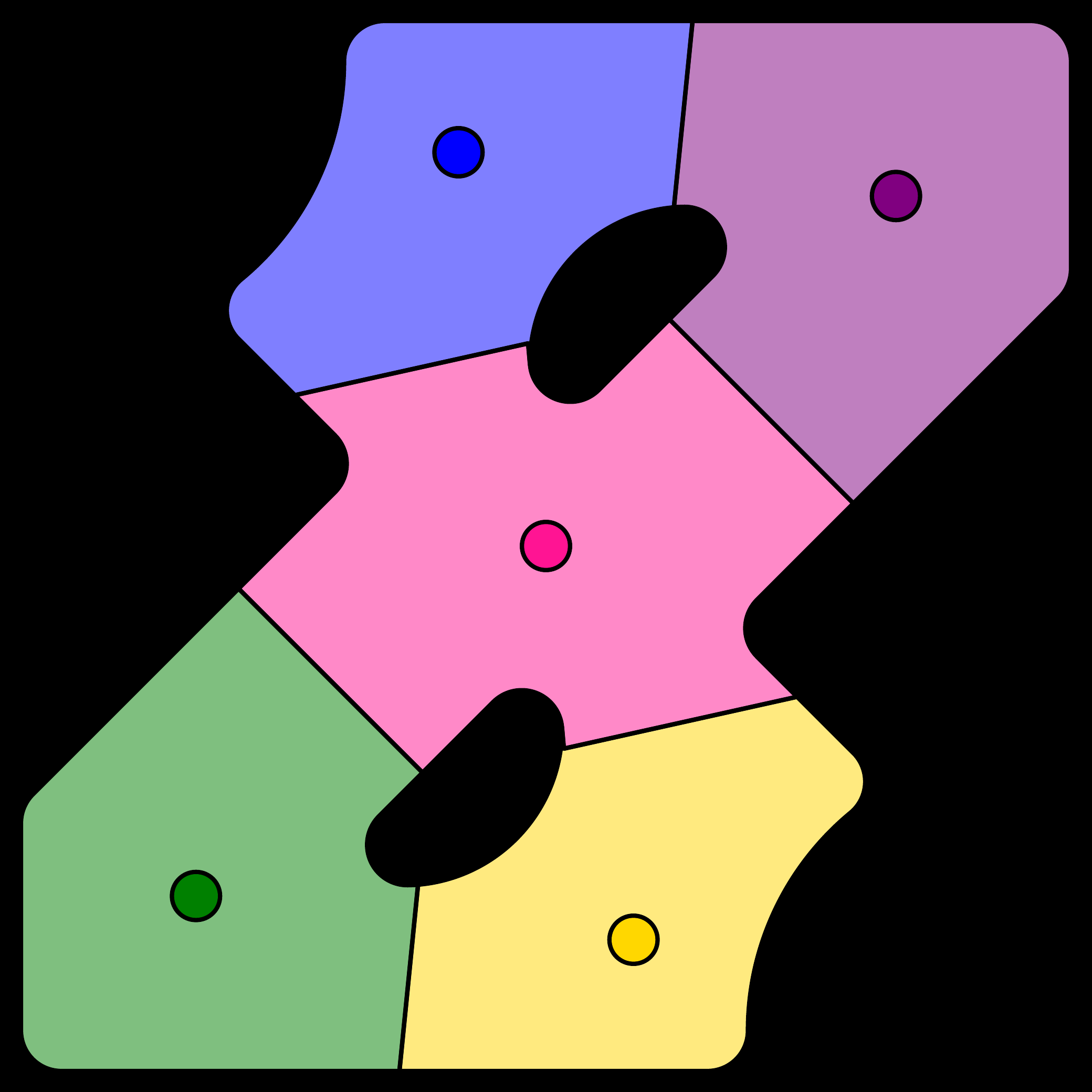}
            \caption{$\mathrm{wtype} \cop{=} \text{DisSplit}$}
            \label{fig:w3}
        \end{subfigure}
        \hfill
        \begin{subfigure}[t]{0.32\columnwidth}
            \centering
            \includegraphics[width=\linewidth]{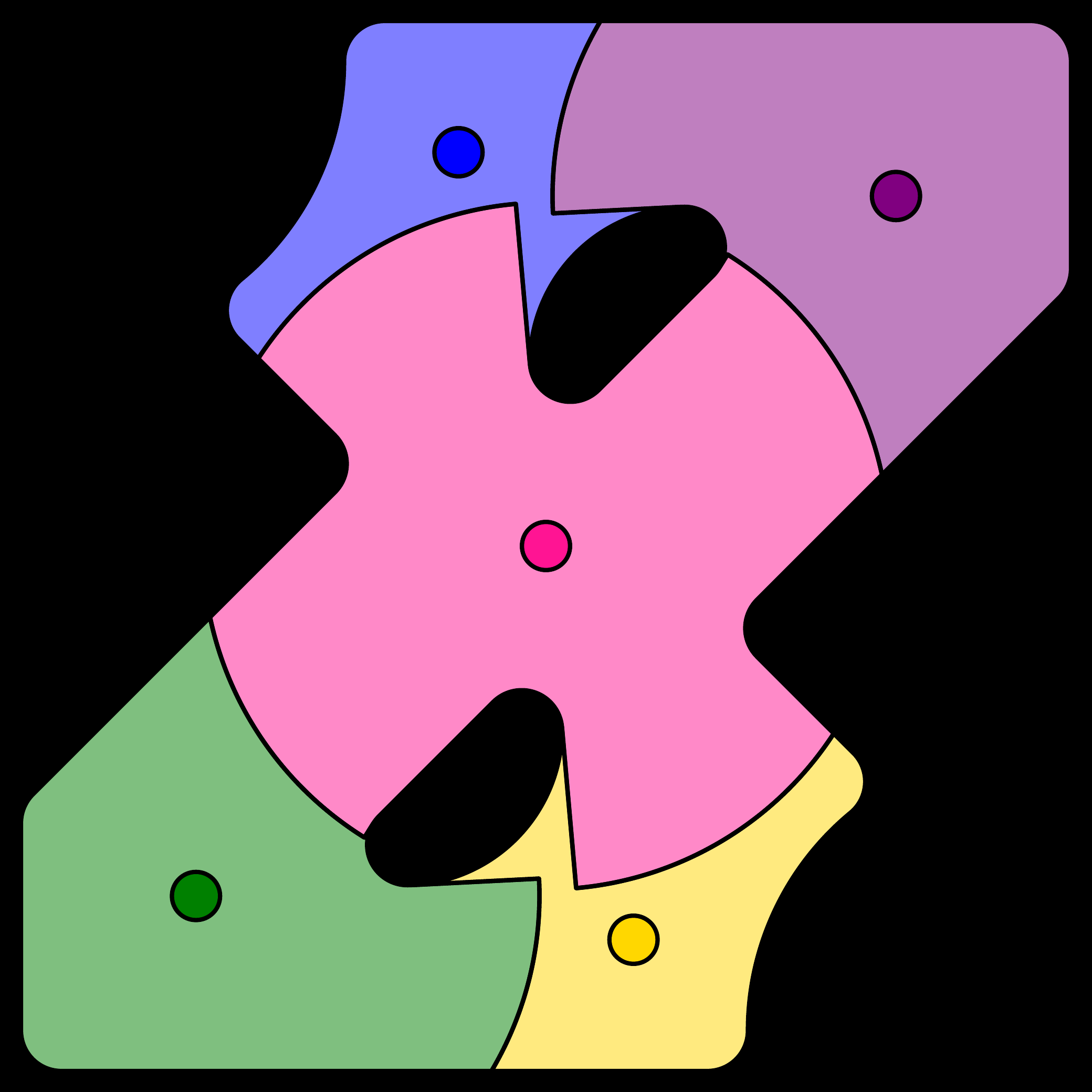}
            \caption{$\mathrm{wtype} \cop{=} \text{DisMaxW}$}
            \label{fig:w4}
        \end{subfigure}
        \hfill
        \begin{subfigure}[t]{0.32\columnwidth}
            \centering
            \includegraphics[width=\linewidth]{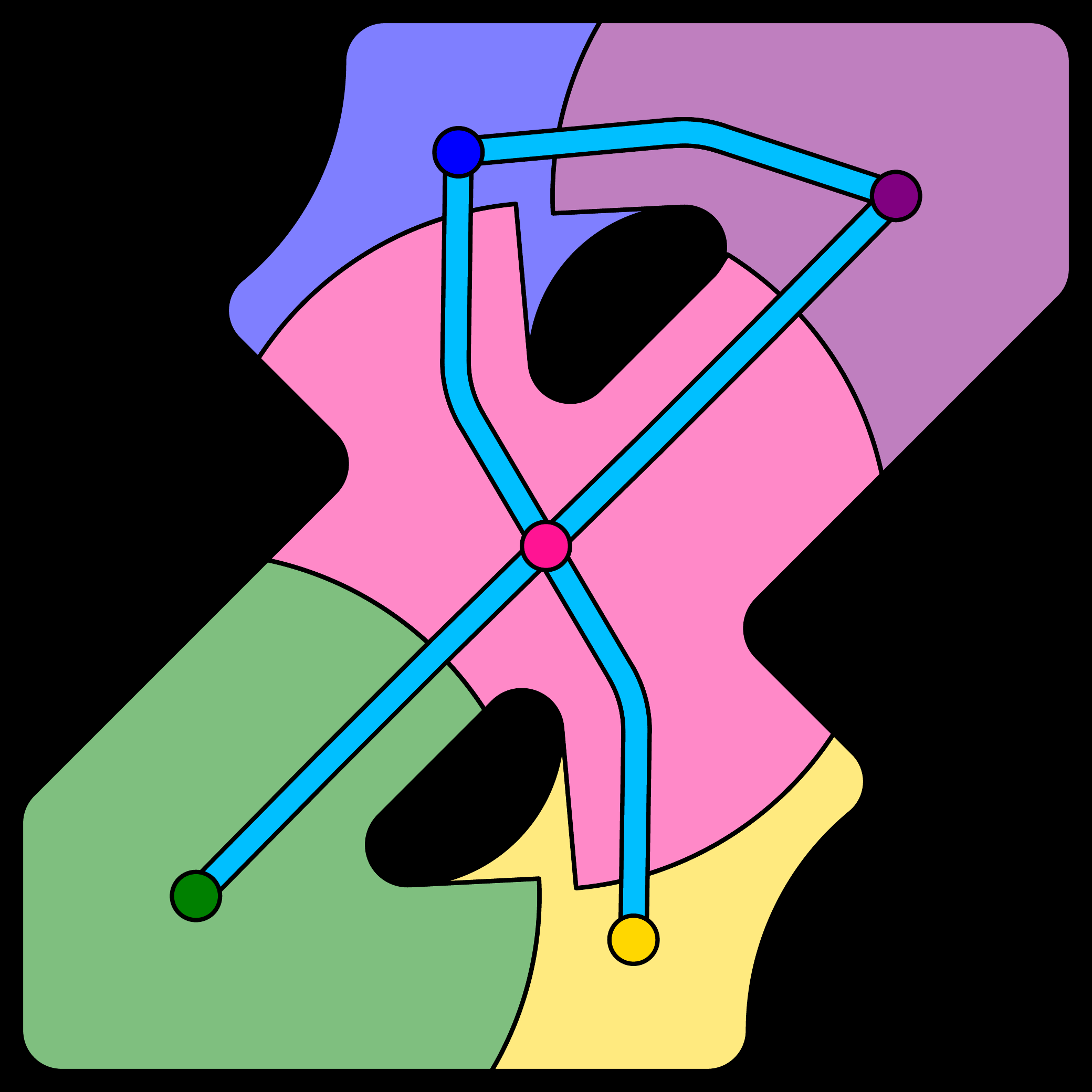}
            \caption{$\mathrm{wtype} \cop{=} \text{DisGreedy}$}
            \label{fig:w5}
        \end{subfigure}
        \caption{
            Illustration of the complete \gspt{} graph $\mathbb{G}$ (\subref{fig:graph}) and the five types of node weights (\subref{fig:w1}--\subref{fig:w5}).
            All weights, except Const (\subref{fig:w1}), are computed using the visibility model and have associated weight-defining regions, depicted in (\subref{fig:w2}--\subref{fig:w5}).
            The weights DisGreedy are determined based on the utility-greedy solution to \dsearch{}, as additionally illustrated in (\subref{fig:w5}).
        }
    \end{figure}

    The objectives of \gspt{} (Eq.~\eqref{eq:gspt-objective}) and \dsearch{} (Eq.~\eqref{eq:objective}) are closely related.
    To align them, we associate the permutation $\sigma$ with a route $\tau$, the latency $\delta(i \cop{\mid} \sigma)$ with the travel time $\mathrm{Time}(\zeta(i) \cop{\mid} \tau)$, and the weight $w(\sigma(i))$ with the probability $p(s_{i} \cop{\mid} S_{i - 1})$.
    While the first two associations hold under a specific condition, the third requires heuristic approximation due to the static nature of $w(\sigma(i))$ versus the conditional nature of $p(s_{i} \cop{\mid} S_{i - 1})$.
    The first two associations are valid if, for every pair $u, v \cop{\in} V$, there exists a path $\tau_{u,v} \cop{:} [0{,}1] \cop{\mapsto} \mathbb{C}_{\mathrm{free}}$ such that $\tau_{u,v}(0) \cop{=} u$ and $\tau_{u,v}(1) \cop{=} v$.
    This condition enables the construction of the \gspt{} graph $\mathbb{G}$, where vertices correspond to guards $V \cop{\coloneqq} G$, with $v_{\mathrm{s}} \cop{\coloneqq} g_{0}$, and an edge $(u, v) \cop{\in} V \cop{\times} V$ represents the quickest path $\iota_{u,v} \cop{\coloneqq} \argmin_{\tau_{u,v}} \mathrm{Time}(1 \cop{\mid} \tau_{u,v})$ (see Fig.~\ref{fig:graph}).
    In the decoupling scheme (Alg.~\ref{alg:decoupling}), this condition is ensured by the SPP at line~\ref{alg:decoupling:spp}.

    \subsubsection{Associating $\sigma$ with $\tau$}

    Given the graph $\mathbb{G}$, associating a vertex permutation $\sigma$ with a route $\tau$ is straightforward.
    The permutation $\sigma$ defines the route $\tau \cop{=} \iota_{g_{0}, \sigma(1)} \cop{\circ} \iota_{\sigma(1), \sigma(2)} \cop{\circ} \ldots \cop{\circ} \iota_{\sigma(n-1), \sigma(n)}$, where $\circ$ denotes associative concatenation.
    Starting at $g_{0}$, the route follows the order in $\sigma$, with each segment corresponding to the quickest path between consecutive guards, as defined by the edges in $\mathbb{G}$.
    Concatenation $\tau \cop{=} \tau_1 \cop{\circ} \tau_2$ merges two paths sharing an endpoint $c \cop{=} \tau_1(1) \cop{=} \tau_2(0)$, forming a new path $\tau$ from $\tau(0) \cop{=} \tau_1(0)$ to $\tau(1) \cop{=} \tau_2(1)$.\footnote{
        Note that if $c$ introduces a discontinuity, $\tau$ becomes piecewise continuous, even if $\tau_1$ and $\tau_2$ are individually continuous.
    }

    \subsubsection{Associating $\delta(i \cop{\mid} \sigma)$ with $\mathrm{Time}(\zeta(i) \cop{\mid} \tau)$}

    This association is established by defining the \gspt{} costs based on the travel time model, where $\vartheta(v)$ represents the time spent at $v_{\mathrm{s}}$ before moving toward the first guard $v$; $d(u, v)$ is the travel time between two guards $u$ and $v$; and $\theta(h, u, v)$ accounts for additional time spent at $u$ when traveling from $h$ to $v$ via $u$.
    The costs for all $h, u, v \cop{\in} V$ are defined as:
    \begin{flalign*}
        \begin{split}
            d(u, v) &\coloneqq \mathrm{Time}(1 \mid \iota_{u,v}), \\
            \theta(h, u, v) &\coloneqq \mathrm{Time}(1 \mid \iota_{h,u} \circ \iota_{u,v}) - d(h, u) - d(u, v), \\
            \vartheta(v) &\coloneqq \theta(v_{\mathrm{aux}}, v_{\mathrm{s}}, v),
        \end{split}
    \end{flalign*}
    where $v_{\mathrm{aux}} \cop{\in} \mathbb{C}$ is an auxiliary configuration defining the sensor's initial orientation (if applicable).
    Combining these cost definitions with the travel time model from Eq.~\eqref{eq:travel-time-model}, we obtain:
    \begin{flalign*}
        \begin{split}
            d(u, v) &= t_{\mathrm{lin}} \mathrm{Len}(\iota_{u,v}) + t_{\mathrm{ang}} \mathrm{Ang}(\iota_{u,v}), \\
            \theta(h, u, v) &= t_{\mathrm{ang}} \mathrm{Ang}(\iota_{h, u}, \iota_{u, v}), \\
            v_{\mathrm{aux}} &\coloneqq (x_0 - \cos(\alpha_0), y_0 - \sin(\alpha_0)),
        \end{split}
    \end{flalign*}
    where $\mathrm{Len}(\tau) \cop{\coloneqq} \mathrm{Len}(1 \cop{\mid} \tau)$ and $\mathrm{Ang}(\tau) \cop{\coloneqq} \mathrm{Ang}(1 \cop{\mid} \tau)$ denote~the total path length and accumulated turning angle, respectively,
    $(x_0\cop{,} y_0) \cop{\coloneqq} g_0$, $\alpha_0$ represents the sensor's initial orientation (if applicable), and
    $\mathrm{Ang}(\tau_1\cop{,} \tau_2) \cop{\coloneqq} \mathrm{Ang}(\tau_1 \cop{\circ} \tau_2) \cop{-} \mathrm{Ang}(\tau_1) \cop{-} \mathrm{Ang}(\tau_2)$ is the turning angle at the connection point of two paths.
    Furthermore, since $\mathrm{Len}(\iota_{u,v})\cop{=}\mathrm{Len}(\iota_{v,u})$,~$\mathrm{Ang}(\iota_{u,v})\cop{=}\mathrm{Ang}(\iota_{v,u})$,~$\mathrm{Ang}(\iota_{h, u}, \iota_{u, v}) \cop{=} \mathrm{Ang}(\iota_{v, u}, \iota_{u, h})$ for all $h, u, v \cop{\in} V$, the travel time model defines a \emph{symmetric} \gspt{} instance.

    \subsubsection{Associating $w(\sigma(i))$ with $p(s_{i} \cop{\mid} S_{i - 1})$}

    This association is heuristic, as the probability $p(s_{i} \cop{\mid} S_{i - 1})$ depends on the sequence $S_{i - 1}$, whereas the static weight $w(\sigma(i))$ depends only on $\sigma(i)$ and not its position $i$.
    Despite this discrepancy, approximating dynamic probabilities with static weights offers a significant computational advantage, as discussed later in Sec.~\ref{subsec:msgvns}.
    We consider three approaches for this association: two baselines and one novel method:
    \begin{itemize}
        \item \emph{Constant Uniform Weights:} $w(v) \cop{\coloneqq} 1 \, \forall v \cop{\in} V$~\citep{Kulich2014}
        \begin{small}
            \begin{itemize}
                \setlength{\leftskip}{0pt}
                \item This method assumes equal importance for all guards, which is generally unrealistic since a guard's relevance depends on its coverage, sequence position, and the object's probability distribution.
            \end{itemize}
        \end{small}
        \item \emph{Visibility Weights:} $w(v) \cop{\coloneqq} w(\mathrm{Vis}(v) \cop{\mid} P) \, \forall v \cop{\in} V$~\citep{Kulich2017}
        \begin{small}
            \begin{itemize}
                \setlength{\leftskip}{0pt}
                \item Here, a guard $v$ is weighted by the probability of the object being in its visible region $\mathrm{Vis}(v)$.
                However, this approach ignores the sequence position and overlaps between visible regions from different configurations.
            \end{itemize}
        \end{small}
        \item \emph{Disjoint Weights:} $w(v) \cop{\coloneqq} w(\mathcal{X}_{v} \cop{\mid} P) \, \forall v \cop{\in} V$ (novel)
        \begin{small}
            \begin{itemize}
                \setlength{\leftskip}{0pt}
                \item This approach mitigates region overlap by assigning disjoint regions $\mathcal{X}_{v}$ to each guard $v$ and weighting it by the probability of the object being in $\mathcal{X}_{v}$.
                In Sec.~\ref{subsec:disjoint-weights}, we introduce three methods for constructing these disjoint regions.
            \end{itemize}
        \end{small}
    \end{itemize}
    Illustrations of all three approaches (comprising five methods in total) are shown in Figs.~\ref{fig:w1}--\subref{fig:w5}.

    \subsubsection{\milaps{}---The Algorithm}

    The complete \milaps{} solution method for \dsearch{}, whose principles have been described thus far, is summarized in Alg.~\ref{alg:milaps}.
    Lines~\ref{alg:milaps:replicate-start}--\ref{alg:milaps:replicate-end} detail the construction of the \gspt{} instance, which is then solved using the Ms-GVNS metaheuristic~\citep{Mikula2022} on line~\ref{alg:milaps:msgvns}.
    This anytime process generates a sequence of graph node permutations, each corresponding to a \gspt{} solution that improves the \gspt{} objective compared to the previous one.
    Lines~\ref{alg:milaps:z}--\ref{alg:milaps:return} conclude the algorithm by optionally evaluating these solutions against the \dsearch{} objective, returning either the best or last index, and transforming all permutations into solution routes.

    \begin{algorithm}
        \caption{\milaps{} for \dsearch{}}
        \label{alg:milaps}
        \begin{algorithmic}[1]%
            \Require Environment $\mathcal{W}$, weighted target regions $P$, initial position and orientation $g_{0},\alpha_0$, visibility model $\mathrm{Vis}$, travel time model $\mathrm{Time}(. \cop{\mid} \tau)$, guard set $G$, $\mathrm{wtype} \cop{\in} \{\text{Const},\allowbreak\text{Vis},\allowbreak\text{DisSplit},\allowbreak\text{DisMaxW},\allowbreak\text{DisGreedy}\}$, and max runtime in seconds $t_{\mathrm{max}}$.
            \Ensure Sequence of solutions $\Upsilon \cop{=} \langle \tau_{k} \cop{:} [0{,}1] \cop{\mapsto} \mathbb{C}_{\mathrm{free}} \rangle_{k=1}^{z}$ and the index of the best \dsearch{} solution $k_{\text{\dsearch{}}} \cop{\in} \{1{:}z\}$.
            \State $V \gets G$; $v_{\mathrm{s}} \gets g_{0}$ \label{alg:milaps:replicate-start}
            \State $\iota_{u,v} \gets \argmin_{\tau_{u,v} : [0{,}1] \mapsto \mathbb{C}_{\mathrm{free}}} \mathrm{Time}(1 \mid \tau_{u,v})$
            \Statex \quad $\text{s.t. }\tau_{u,v}(0) {=} u, \tau_{u,v}(1) \cop{=} v \, \forall u,v \cop{\in} V$\label{alg:milaps:paths}
            \State Compute the costs $\vartheta$ $d$, and $\theta$.\label{alg:milaps:costs}
            \If{$\mathrm{wtype} = \text{Const}$}
                \label{alg:milaps:wtype-start}
                \State $w(v) \gets 1 \, \forall v \cop{\in} V$
            \EndIf{}
            \If{$\mathrm{wtype} = \text{Vis}$}
                \State $\mathcal{X}_{v} \gets \mathrm{cl}(\mathrm{Vis}(v) \setminus \mathrm{Vis}(g_{0})) \, \forall v \cop{\in} V$
                \State $w(v) \gets w(\mathcal{X}_{v} \mid P) \, \forall v \cop{\in} V$
            \EndIf{}
            \If{$\mathrm{wtype} \in \{\text{DisMaxW},\text{DisGreedy},\text{DisSplit}\}$}
                \State Compute disjoint regions $\mathcal{X}_{v} \, \forall v \cop{\in} V$.
                \State $w(v) \gets w(\mathcal{X}_{v} \mid P) \, \forall v \cop{\in} V$
            \EndIf{}\label{alg:milaps:wtype-end}
            \State $\mathrm{gsp} \gets \llbracket \vartheta(v) \cop{=} 0\, \forall v \cop{\in} V \rrbracket \land \llbracket \theta(h,u,v) \cop{=} 0 \, \forall h,u,v \cop{\in} V\rrbracket$\label{alg:milaps:properties-start}
            \State $\mathrm{tdp} \gets \mathrm{gsp} \land \mathrm{wtype} \cop{=} \text{Const}$\label{alg:milaps:properties-end}
            \If{$\mathrm{tdp}$}
                \label{alg:milaps:graph-start}
                $\mathbb{G} \gets (V, v_{\mathrm{s}}, d)$
            \ElsIf
                    {$\mathrm{gsp}$}
                $\mathbb{G} \gets (V, v_{\mathrm{s}}, w, d)$
            \Else
                    {}
                $\mathbb{G} \gets (V, v_{\mathrm{s}}, w, \vartheta, d, \theta)$
            \EndIf{}\label{alg:milaps:replicate-end}
            \State $\langle \sigma_{k} \rangle_{k=1}^{z} \gets \mathrm{MsGVNS}(\mathbb{G}, t_{\mathrm{max}})$\label{alg:milaps:msgvns}
            \State $k_{\text{\dsearch{}}} \gets z$\label{alg:milaps:z}
            \State \emph{(optional)} $S_{n;k} \gets \langle s_{i;k} \cop{=} \sigma_{k}(i) \rangle_{i=0}^{n} \, \forall k \cop{\in} \{1{:}z\}$;
            \Statex \quad $k_{\text{\dsearch{}}} \gets \argmin_{k=1}^{z} \sum_{i=1}^{n} \delta(i \mid \sigma_{k}) p(s_{i;k} \mid S_{i-1;k})$\label{alg:milaps:bestdetms}
            \State $\Upsilon \gets \langle \tau_{k} \cop{=} \iota_{g_{0},\sigma_{k}(1)} \cop{\circ} \iota_{\sigma_{k}(1),\sigma_{k}(2)} \cop{\circ} \ldots \cop{\circ} \iota_{\sigma_{k}(n-1),\sigma_{k}(n)} \rangle_{k=1}^{z}$\label{alg:milaps:contpaths}
            \State \Return $\Upsilon$, $k_{\text{\dsearch{}}}$\label{alg:milaps:return}
        \end{algorithmic}
    \end{algorithm}

    Ms-GVNS operates on the \gspt{} graph, taking the runtime limit $t_{\mathrm{max}}$ as a parameter, terminating execution if exceeded.
    It generates a sequence of node permutations $\langle \sigma_{k} \rangle_{k=1}^{z}$, where $k \cop{<} z$ represents intermediate best solutions, and $k \cop{=} z$ is the final \gspt{} solution.
    Ms-GVNS was chosen for its ability to deliver high-quality solutions within restricted time frames, as demonstrated in~\citep{Mikula2022}.
    It is a multi-start variant of the \emph{generalized variable neighborhood search} (GVNS) metaheuristic, which builds on the stochastic \emph{variable neighborhood search}~(VNS)~\citep{Mladenovic1997,Hansen2019} and integrates the deterministic \emph{variable neighborhood descent} (VND)~\citep{Mladenovic1997,Duarte2018} for local search.
    Originally developed for the TDP~\citep{Mikula2022}, Ms-GVNS was later adapted for the GSP in~\citep{Blaha2024} and further tailored for \gspt{} in this work, with details provided in Sec.~\ref{subsec:msgvns}.


    \subsection{\milaps{} with Replanning for \dsearch{}}
    \label{subsec:milaps-replanning}

    As an alternative to Alg.~\ref{alg:milaps}, we propose a \milaps{} variant with a replanning mechanism to improve solution quality.
    It leverages the anytime nature of Ms-GVNS to iteratively refine the solution within a fixed runtime limit, with each refinement step incorporating updated weights in the \gspt{} objective.
    In each iteration, the tail of the current solution is discarded, the \gspt{} weights are updated based on collected observations, and the Ms-GVNS metaheuristic is restarted to generate a refined replacement for the discarded tail.
    This replanning procedure brings the solution closer to the \dsearch{} objective and is designed to mitigate the limitations of static weights in the \gspt{} objective.
    The \emph{\milaps{} with replanning} algorithm, presented in Alg.~\ref{alg:milaps-replanning}, extends the original \milaps{} algorithm (Alg.~\ref{alg:milaps}) by replacing line~\ref{alg:milaps:msgvns} with lines~\ref{alg:milaps-replanning:runtime}--\ref{alg:milaps-replanning:return}.
    It introduces two new parameters: the replanning schedule $\langle \mathrm{replan}_{i} \rangle_{i=1}^{n}$, a binary sequence indicating when replanning occurs ($\mathrm{replan}_{i} \cop{=} 1$ triggers replanning after the $i$-th guard), and the sensing policy, which defaults to \dsearch{} for pure \dsearch{} instances or is set to either \dsearch{} or \search{} in the decoupling scheme (Sec.~\ref{subsec:decoupling-framework}).
    Additionally, Alg.~\ref{alg:milaps-replanning} assumes $\mathrm{wtype} \cop{\neq} \text{Const}$, as the replanning mechanism relies on weight updates, which are inapplicable to constant uniform weights.

    \begin{algorithm}
        \caption{\milaps{} with Replanning for \dsearch{}}
        \label{alg:milaps-replanning}
        \begin{algorithmic}[1]%
            \Require Same as in Alg.~\ref{alg:milaps}, except $\mathrm{wtype} \cop{\neq} \text{Const}$, and the following additional inputs: replanning schedule $\langle \mathrm{replan}_{i} \cop{\in} \{0, 1\} \rangle_{i=1}^{n}$, sensing policy $\mathrm{sens} \cop{\in} \{\text{\dsearch{}}, \text{\search{}}\}$.
            \Ensure Same as in Alg.~\ref{alg:milaps}.
            \State Follow steps from Alg.~\ref{alg:milaps} (lines~\ref{alg:milaps:replicate-start}--\ref{alg:milaps:replicate-end}) to obtain:
            \Statex \quad $V$, $v_{\mathrm{s}}$, $\iota_{u,v} \, \forall u, v \cop{\in} V$, $\mathcal{X}_{v} \, \forall v \cop{\in} V$, and $\mathbb{G}$.
            \State $t'_{\mathrm{max}} \gets t_{\mathrm{max}} \,/\, (1 + \sum_{i=1}^{n} \mathrm{replan}_{i})$\label{alg:milaps-replanning:runtime}
            \State $\langle \sigma_{k} \rangle_{k=1}^{z} \gets \mathrm{MsGVNS}(\mathbb{G}, t'_{\mathrm{max}})$\label{alg:milaps-replanning:msgvns}
            \State $\sigma_{\mathrm{curr}} \gets \sigma_{z}$\label{alg:milaps-replanning:initial}
            \State $b \gets v_{\mathrm{s}}$\label{alg:milaps-replanning:loop}
            \For{$r \gets 1$ \textbf{to} $n$}
                \State $a \gets \sigma_{\mathrm{curr}}(r)$
                \State $V \gets V \setminus \{b\}$\label{alg:milaps-replanning:remove}
                \If{$\mathrm{sens} = \text{\dsearch{}}$}
                    \State $\mathcal{X}_{v} \gets \mathrm{cl}(\mathcal{X}_{v} \setminus \mathrm{Vis}(a)) \, \forall v \cop{\in} V$
                \EndIf{}
                \If{$\mathrm{sens} = \text{\search{}}$}
                    \State $(\hat{\zeta},\hat{S}_{\hat{n}}) \gets \mathrm{sens}(\iota_{a,b})$
                    \State $\mathcal{X}_{v} \gets \mathrm{cl}(\mathcal{X}_{v} \setminus \mathcal{W}_{\mathrm{seen}}(\hat{S}_{\hat{n}})) \, \forall v \cop{\in} V$
                \EndIf{}\label{alg:milaps-replanning:policy}
                \If{$\mathrm{replan}_{r} = 1$}
                    \label{alg:milaps-replanning:replan}
                    \State $\mathbb{G} \gets \mathbb{G}[V]$\label{alg:milaps-replanning:graph}
                    \State $\mathbb{G}.v_{\mathrm{s}} \gets a$
                    \State $\mathbb{G}.w(v) \gets w(\mathcal{X}_{v} \mid P) \, \forall v \cop{\in} V$\label{alg:milaps-replanning:w}
                    \State $\langle \hat{\sigma}_{k} \rangle_{k=1}^{\hat{z}} \gets \mathrm{MsGVNS}(\mathbb{G}, t'_{\mathrm{max}})$\label{alg:milaps-replanning:msgvns-replan}
                    \State $\sigma_{\mathrm{curr}}(i) \gets \hat{\sigma}_{\hat{z}}(i - r) \, \forall i \cop{\in} \{r{:}n\}$\label{alg:milaps-replanning:replace}
                    \State $z \gets z + 1$; $\sigma_{z} \gets \sigma_{\mathrm{curr}}$\label{alg:milaps-replanning:store}
                \EndIf{}
                \State $b \gets a$\label{alg:milaps-replanning:next}
            \EndFor{}\label{alg:milaps-replanning:return}
            \State Follow steps from Alg.~\ref{alg:milaps} (lines~\ref{alg:milaps:z}--\ref{alg:milaps:return}).
        \end{algorithmic}
    \end{algorithm}

    Alg.~\ref{alg:milaps-replanning} begins by constructing the \gspt{} instance and solving it with Ms-GVNS, as in the original Alg.~\ref{alg:milaps}, but divides the runtime limit $t_{\mathrm{max}}$ between the initial solution and replanning steps (line~\ref{alg:milaps-replanning:runtime}).
    After obtaining the initial sequence of solutions (line~\ref{alg:milaps-replanning:msgvns}), the last solution is selected for refinement through replanning (line~\ref{alg:milaps-replanning:initial}).
    The replanning loop (lines~\ref{alg:milaps-replanning:loop}--\ref{alg:milaps-replanning:next}) iterates over the guards, storing the current guard in $a$ and the previous one in $b$.
    At each step, $b$ is removed from $V$, marking it as visited, and the sensing policy updates the weight-defining regions $\mathcal{X}_{v}$ for all $v \cop{\in} V$ (lines~\ref{alg:milaps-replanning:remove}--\ref{alg:milaps-replanning:policy}).
    If the replanning schedule triggers replanning after the current guard (line~\ref{alg:milaps-replanning:replan}), the graph is updated to its subgraph induced by $V$, with the starting vertex set to $a$, and weights recomputed based on the current weight-defining regions (lines~\ref{alg:milaps-replanning:graph}--\ref{alg:milaps-replanning:w}).
    Ms-GVNS is applied to the updated graph, and the last solution (excluding intermediate ones) replaces the tail of the current solution (lines~\ref{alg:milaps-replanning:msgvns-replan}--\ref{alg:milaps-replanning:replace}).
    The updated solution is added to the output sequence, increasing its size $z$ by 1 (line~\ref{alg:milaps-replanning:store}).
    The loop continues until all guards are visited.
    Finally, the algorithm concludes with lines~\ref{alg:milaps:z}--\ref{alg:milaps:return} of the original Alg.~\ref{alg:milaps}.


    \subsection{Disjoint Regions for Weight Approximation}
    \label{subsec:disjoint-weights}

    The disjoint weights method assumes that a guard's importance is proportional to the probability of the object being within its \emph{weight-defining region}, which is \emph{disjoint} from those of other guards.
    Although this assumption does not strictly align with the \dsearch{} objective, it provides a heuristic approximation for estimating guard importance, extending the visibility weights concept~\citep{Kulich2017}, which includes overlapping regions.
    This section introduces three novel methods for constructing disjoint weight-defining regions $\mathcal{X}_{v} \, \forall v \cop{\in} V$, where each region is a subset of its corresponding guard's visibility region.

    \subsubsection{Fair-Split Disjoint Regions (DisSplit)}
    The first method iteratively divides the visible regions of neighboring guards into disjoint regions.
    Overlapping areas are fairly split, with each region retaining the portion closer to its associated guard, determined by Euclidean distance (see Fig.~\ref{fig:w3}).
    Given a guard ordering $\mathrm{o} \cop{:} \{0{:}n\} \cop{\mapsto} V$, the fair-split disjoint regions method is detailed in Alg.~\ref{alg:disjoint-split}.
    Here, $\mathcal{H}(u, v)$ defines the half-plane closer to $u$ than $v$: $\mathcal{H}(u, v) \cop{\coloneqq} \{ x \cop{\in} \mathbb{R}^2 \cop{:} \|x \cop{-} u\| \cop{\leq} \|x \cop{-} v\| \}$.
    Preliminary tests indicate that the ordering $\mathrm{o}$ has minimal impact on the method, allowing it to be chosen arbitrarily.
    Notably, without considering obstacles and visibility constraints, Alg.~\ref{alg:disjoint-split} would generate Voronoi regions around the guards.
    However, in our case, the resulting disjoint regions do not necessarily resemble Voronoi regions, particularly near obstacles, and may even include multiple disconnected components.
    \begin{algorithm}
        \caption{Fair-Split Disjoint Regions}
        \label{alg:disjoint-split}
        \begin{algorithmic}[1]
            \State $\mathcal{X}_{v} \gets \mathrm{Vis}(v) \, \forall v \cop{\in} V$
            \For{$i \gets 0$ \textbf{to} $n$}
                \State $u \gets \mathrm{o}(i)$
                \For{$j \gets i + 1$ \textbf{to} $n$}
                    \State $v \gets \mathrm{o}(j)$
                    \State $\mathcal{X}_{u} \gets \mathrm{cl}(\mathcal{X}_{u} \setminus (\mathcal{H}(v, u) \cap \mathcal{X}_{v}))$
                    \State $\mathcal{X}_{v} \gets \mathrm{cl}(\mathcal{X}_{v} \setminus (\mathcal{H}(u, v) \cap \mathcal{X}_{u}))$
                \EndFor{}
            \EndFor{}
            \State \Return $\mathcal{X}_{v} \, \forall v \cop{\in} V$
        \end{algorithmic}
    \end{algorithm}

    \subsubsection{Maximum-Weight Disjoint Regions (DisMaxW)}

    The next two methods follow a similar approach.
    Given a guard ordering $\mathrm{o} \cop{:} \{0{:}n\} \cop{\mapsto} V$, the weight-defining region for the $i$-th guard is defined as its visible region minus those visible from preceding guards: $\mathcal{X}_{\mathrm{o}(i)} \cop{\coloneqq} \mathrm{cl}(\mathrm{Vis}(\mathrm{o}(i)) \cop{\setminus} {\bigcup}_{j=0}^{i-1} \mathrm{Vis}(\mathrm{o}(j))) \,\forall i \cop{\in} \{0{:}n\}$.
    Instead of using a predefined ordering, we propose a greedy algorithm (Alg.~\ref{alg:disjoint-utility}) that dynamically determines the guard ordering based on a utility function while simultaneously computing the weight-defining regions.
    At each step, the algorithm selects the guard with the highest utility, and its visible region is subtracted from those of all remaining guards.
    Alg.~\ref{alg:disjoint-utility} applies to both DisMaxW and DisGreedy, which differ only in their utility function definitions.

    \begin{algorithm}
        \caption{Disjoint Regions Based on Utility Maximization}
        \label{alg:disjoint-utility}
        \begin{algorithmic}[1]
            \State $\mathcal{X}_{v} \gets \mathrm{Vis}(v) \, \forall v \cop{\in} V$
            \State $\mathrm{o}(0) \gets v_{\mathrm{s}}$
            \State $V' \gets V \setminus \{v_{\mathrm{s}}\}$
            \For{$i \gets 1$ \textbf{to} $n$}
                \State $\mathrm{o}(i) \gets \argmax_{v \in V'} \mathrm{Util}(v, i \mid \mathrm{o})$
                \State $V' \gets V' \setminus \{\mathrm{o}(i)\}$
                \State $\mathcal{X}_{v} \gets \mathrm{cl}(\mathcal{X}_{v} \setminus \mathcal{X}_{\mathrm{o}(i)}) \, \forall v \cop{\in} V'$
            \EndFor{}
            \State \Return $\mathcal{X}_{v} \, \forall v \cop{\in} V$
        \end{algorithmic}
    \end{algorithm}

    The DisMaxW method, illustrated in Fig.~\ref{fig:w4}, maximizes the weight of the next guard, ignoring travel time costs:
    \begin{flalign*}
        \mathrm{Util}(v, i \mid \mathrm{o}) \coloneqq w(\mathcal{X}_{v} \mid P).
    \end{flalign*}

    \subsubsection{Greedy-Solution Disjoint Regions (DisGreedy)}

    This method maximizes the ratio of the next guard's weight to the travel time cost of reaching it:
    \begin{flalign}
        \mathrm{Util}(v, i \mid \mathrm{o}) &\coloneqq \frac{w(\mathcal{X}_{v} \mid P)}{\theta(v, i \mid \mathrm{o}) + d(\mathrm{o}(i-1), v)},\nonumber\\
        \theta(v, i \mid \mathrm{o}) &\coloneqq
        \begin{cases}
            \vartheta(v), \text{ if } i = 1, \\
            \theta(\mathrm{o}(i-2), \mathrm{o}(i-1), v), \text{ otherwise}.\!\!\!\!
        \end{cases}\label{eq:theta-greedy}
    \end{flalign}
    This method is a 1-depth variant of the utility greedy algorithm for \dsearch{} proposed in~\citep{Sarmiento2003}.
    Unlike the referenced work, where the algorithm directly solves \dsearch{}, here it is used exclusively to construct weight-defining regions for \gspt{} optimization.
    Fig.~\ref{fig:w5} illustrates the weight-defining regions with the greedy solution overlaid.

    \subsection{Ms-GVNS for \gspt{}}
    \label{subsec:msgvns}

    The Ms-GVNS metaheuristic, originally developed for the TDP in~\citep{Mikula2022}, was designed for mobile robotics, providing high-quality solutions within limited runtime budgets.
    Its time-efficient focus makes it well-suited for the \milaps{} framework, particularly in the replanning approach (Sec.~\ref{subsec:milaps-replanning}), where solving subproblems within constrained time budgets is crucial.
    Ms-GVNS is a multi-start variant of VNS~\citep{Mladenovic1997,Hansen2019}, comprising the following key components:
    \begin{enumerate}
        \item A deterministic greedy algorithm that generates initial solutions for each restart.
        \item Stochastic perturbation operators with increasing intensity to escape local optima.
        \item VND~\citep{Mladenovic1997,Duarte2018}, a deterministic local search algorithm that iteratively descends toward a local optimum within a predefined sequence of neighborhood structures.
        \item A carefully selected sequence of VND local search operators, each defining efficiently explorable neighborhood structures.
    \end{enumerate}

    To minimize additional parameters, we retain the original Ms-GVNS configuration from~\citep{Mikula2022}, making only essential adaptations for the \gspt{} problem.
    Since both the TDP and \gspt{} represent solutions as permutations of graph nodes, the definitions of the perturbation and local search operators remain unchanged.
    The main adaptations include:
    \begin{itemize}
        \item Adjusting the greedy algorithm to generate initial solutions aligned with the \gspt{} objective.
        \item Modifying local search computations to efficiently explore \gspt{} neighborhoods.
    \end{itemize}
    Details of these adaptations are provided below, while the full Ms-GVNS metaheuristic is described in~\citep{Mikula2022}.

    \subsubsection{Initial Solution Adaptation}

    In the TDP, the greedy algorithm selects the next node $v$ based on the minimum $d(u, v)$, where $u$ is the last node in the partial solution.
    For the \gspt{}, the algorithm is modified to prioritize nodes with a lower cost-to-weight ratio:
    \begin{flalign*}
        \argmin_{v \notin \sigma} \frac{\theta(v, i \mid \sigma) + d(\sigma(i - 1), v)}{w(v)},
    \end{flalign*}
    where $i$ is the index of the next node, and $\theta(v, i \cop{\mid} \sigma)$ is defined in Eq.~\eqref{eq:theta-greedy}.

    \subsubsection{Efficient Neighborhood Exploration (Definitions)}

    Efficient neighborhood exploration relies on local search operators.
    A local search operator $\mathrm{Op}$ is defined as a mapping $\mathrm{Op} \cop{:} \Pi \cop{\times} \mathbb{J}_{\mathrm{Op}} \cop{\mapsto} \Pi$, where $\Pi$ is the solution space and $\mathbb{J}_{\mathrm{Op}}$ is the parameter space.
    Exploring the $\mathrm{Op}$-neighborhood of a solution $\sigma$ involves finding the parameter $J^{\star}$ that minimizes the cost of the resulting solution:
    \begin{flalign}
        J^{\star} &\coloneqq \argmin_{J \in \mathbb{J}_{\mathrm{Op}}} \mathrm{Cost}(\mathrm{Op}(\sigma, J)) = \argmin_{J \in \mathbb{J}_{\mathrm{Op}}} \Delta_{J}^{\mathrm{Op}}, \nonumber\\
        \Delta_{J}^{\mathrm{Op}} &\coloneqq \mathrm{Cost}(\mathrm{Op}(\sigma, J)) - \mathrm{Cost}(\sigma), \label{eq:improvement}
    \end{flalign}
    where $\mathrm{Cost}$ represents the objective function (e.g., Eq.~\eqref{eq:gspt-objective} for \gspt{}), and $\Delta_{J}^{\mathrm{Op}}$ denotes the cost improvement from applying $\mathrm{Op}$ with parameter $J$.
    The complexity of neighborhood exploration is at most $\mathbb{O}(\eta^{j + c})$, where $\eta$ is the problem size, $j$ is the dimension of $\mathbb{J}_{\mathrm{Op}}$, and $c$ is the complexity of computing the objective function.
    Efficient exploration requires $\Delta_{J}^{\mathrm{Op}}$ be computable in $\mathbb{O}(1)$, reducing the overall complexity to $\mathbb{O}(\eta^{j})$.

    \subsubsection{Efficient Exploration in Ms-GVNS for \gspt{}}

    Ms-GVNS employs two local search operators: $\mathrm{2string}$\footnote{
        The $\mathrm{2string}$ operator is not directly used in Ms-GVNS but serves as the basis for defining $\mathrm{Or\text{-}opt}$ and other variants with two fixed parameters~\citep{Mikula2022}.
        Its improvement formula is derived for convenience, eliminating the need to derive separate formulas for each variant.
    } ($j \cop{=} 4$) and $\mathrm{2opt}$ ($j \cop{=} 2$).
    The \gspt{} objective (Eq.~\eqref{eq:gspt-objective}) requires $c \cop{=} 2$, resulting in unoptimized computational complexities of $\mathbb{O}(\eta^{6})$ for $\mathrm{2string}$ and $\mathbb{O}(\eta^{4})$ for $\mathrm{2opt}$, where $\eta \cop{\coloneqq} |V| \cop{=} n \cop{+} 1$.
    Efficient exploration reduces these complexities to $\mathbb{O}(\eta^{4})$ for $\mathrm{2string}$ and $\mathbb{O}(\eta^{2})$ for $\mathrm{2opt}$ by leveraging precomputed auxiliary structures.
    These auxiliary structures, definitions of the operators, and their $\mathbb{O}(1)$ improvement formulas are provided in Appx.~\ref{app:local-search}.
    The $\mathrm{2string}$ improvement formula supports both symmetric and asymmetric graphs, whereas $\mathrm{2opt}$ assumes symmetry.
    For asymmetric graphs, $\mathrm{2opt}$ requires computing the improvement using definition~\eqref{eq:improvement}, which may degrade performance.

    \subsection{Practical Considerations of Framework Implementation}
    \label{subsec:implementation}

    Apart from the choice of the SPP method, the proposed framework, in theoretical terms, imposes no specific constraints on the shape of the environment's boundary, allowing for arbitrary curves.
    In practice, \emph{polygonal} or \emph{rectilinear} boundaries are common approximations.
    Alternatively, the environment can be represented as a \emph{binary grid}, where each cell is classified as occupied or free, provided that other components, such as the visibility model, are adapted accordingly.
    For grid-based representations, the top-level decoupling scheme remains applicable.
    However, unlike in continuous representations, it does not act as a discretization step but still serves to construct a reduced solution space that satisfies problem constraints.

    We demonstrate the framework using a polygonal environment, as it is easier to implement than curved boundaries, avoids the resolution dependency of rectilinear or grid-based representations~\citep{Harabor2022}, and aligns with our chosen SPP method.
    A \emph{polygonal environment} $\mathcal{W}$ consists of one or more \emph{simple polygons}---regions bounded by connected line segments with no self-intersections.
    It has a single \emph{outer boundary} defining its exterior shape and zero or more inner boundaries \emph{(holes)} representing additional obstacles.
    The rest of the framework conforms to the polygonal representation: paths are \emph{polylines} (piecewise linear curves) in a polygonal $\mathbb{C}_{\mathrm{free}}$; the visibility model is approximated using a \emph{simple star-shaped polygon}; region clipping operations are performed with \emph{polygon clipping algorithms}; and the \search{} sensing policy is defined by discretizing line segments.
    Details are provided below.

    We approximate the computation of quickest paths between all pairs of guards by finding all-pairs shortest paths in the polygonal representation of $\mathbb{C}_{\mathrm{free}}$, where shortest paths are polylines that turn at reflex vertices.
    To compute shortest paths between all pairs $u, v \cop{\in} G$, we use a precomputed visibility graph for $G \cop{\cup} R$ (where $R$ are the reflex vertices) combined with Dijkstra's algorithm~\citep{Dijkstra1959}.
    The visibility graph is constructed using the \emph{triangular expansion algorithm} (TEA)~\citep{Bungiu2014,Xu2015}, implemented in the T\v{r}iVis C++ visibility library\footnote{Available at \url{https://github.com/janmikulacz/trivis}}~\citep{Mikula2024}.

    T\v{r}iVis is also used to efficiently compute visibility regions via TEA with early termination~\citep{Mikula2024}.
    For a limited range $r_{\mathrm{vis}}\cop{<}\infty$, we approximate the visibility region's boundary by sampling circular arcs equidistantly with a maximum distance $d_{\mathrm{circ}}$ along the arc.
    To balance approximation precision and runtime scalability, we define $d_{\mathrm{circ}}$ such that the number of samples $\big\lfloor \frac{2\pi r_{\mathrm{vis}}}{d_{\mathrm{circ}}} \big\rfloor$ for a full circle scales with $4r_{\mathrm{vis}}$, constrained to a minimum of 16 and a maximum of 64 samples: $d_{\mathrm{circ}} \cop{=} \max\big(\frac{\pi}{32}r_{\mathrm{vis}}, \min\big(\frac{\pi}{8}r_{\mathrm{vis}}, \frac{\pi}{2}\big)\big)$.
    For polygon clipping, we use Clipper2, an extended C++ implementation\footnote{Available at \url{https://github.com/AngusJohnson/Clipper2}} of the \emph{Vatti clipping algorithm}~\citep{Vatti1992}.
    Clipper2 also computes $\mathbb{C}_{\mathrm{free}}$ by inflating obstacles by $r_{\mathrm{fp}}$ and is used for various preprocessing tasks.

    The evaluated \search{} sensing policy for a polyline path $\tau$ samples each segment equidistantly with a maximum distance $d_{\mathrm{sens}}$, always including endpoints, ensuring alignment with the \dsearch{} sensing policy.
    This alignment guarantees that any feasible \dsearch{} solution is also feasible for \search{} (recall the discussion in Sec.~\ref{subsec:decoupling-framework}).
    To maintain scalability across different environment sizes and visibility radii, we define $d_{\mathrm{sens}} \cop{=} \min\big(\frac{1}{2}r_{\mathrm{vis}}, \frac{1}{100} \sqrt{x^2 \cop{+} y^2}\big)$, where $x$ and $y$ are the map's width and height.

    \section{Quantitative Evaluation}
    \label{sec:quantitative-evaluation}

    \subsection{Methodology}

    \subsubsection{Evaluation Objectives and Setup}

    Our quantitative evaluation aims to thoroughly assess the \search{} framework in terms of solution quality and computational efficiency, compare it against multiple baselines, and analyze its performance on a large-scale dataset of environments with varying characteristics.
    We select the classical scenarios (\dclassical{}/\classical{}) for evaluation, as they form the core of the \search{} problem, with parameters set to $r_{\mathrm{fp}} \cop{=} 0, r_{\mathrm{vis}} \cop{=} \infty, t_{\mathrm{lin}} \cop{=} 1, t_{\mathrm{ang}} \cop{=} 0$, and a uniform probability distribution for the object's location, represented by a single 1-weighted target region covering the entire environment: $P \cop{=} \{(p_1 \cop{=} 1, \mathcal{P}_1 \cop{=} \mathcal{W})\}$.
    With $r_{\mathrm{fp}} \cop{=} 0$, this setup ensures that any connected environment has a feasible solution, allowing us to focus on the optimization aspect of \search{}.
    As the only deviation from the classical scenarios, we set $\epsilon \cop{=} 10^{-5}$ to slightly relax the 100\% coverage constraint.

    \subsubsection{Dataset Generation}

\begin{figure}
        \centering
        \begin{subfigure}[t]{0.325\columnwidth}
            \centering
            \includegraphics[width=\linewidth]{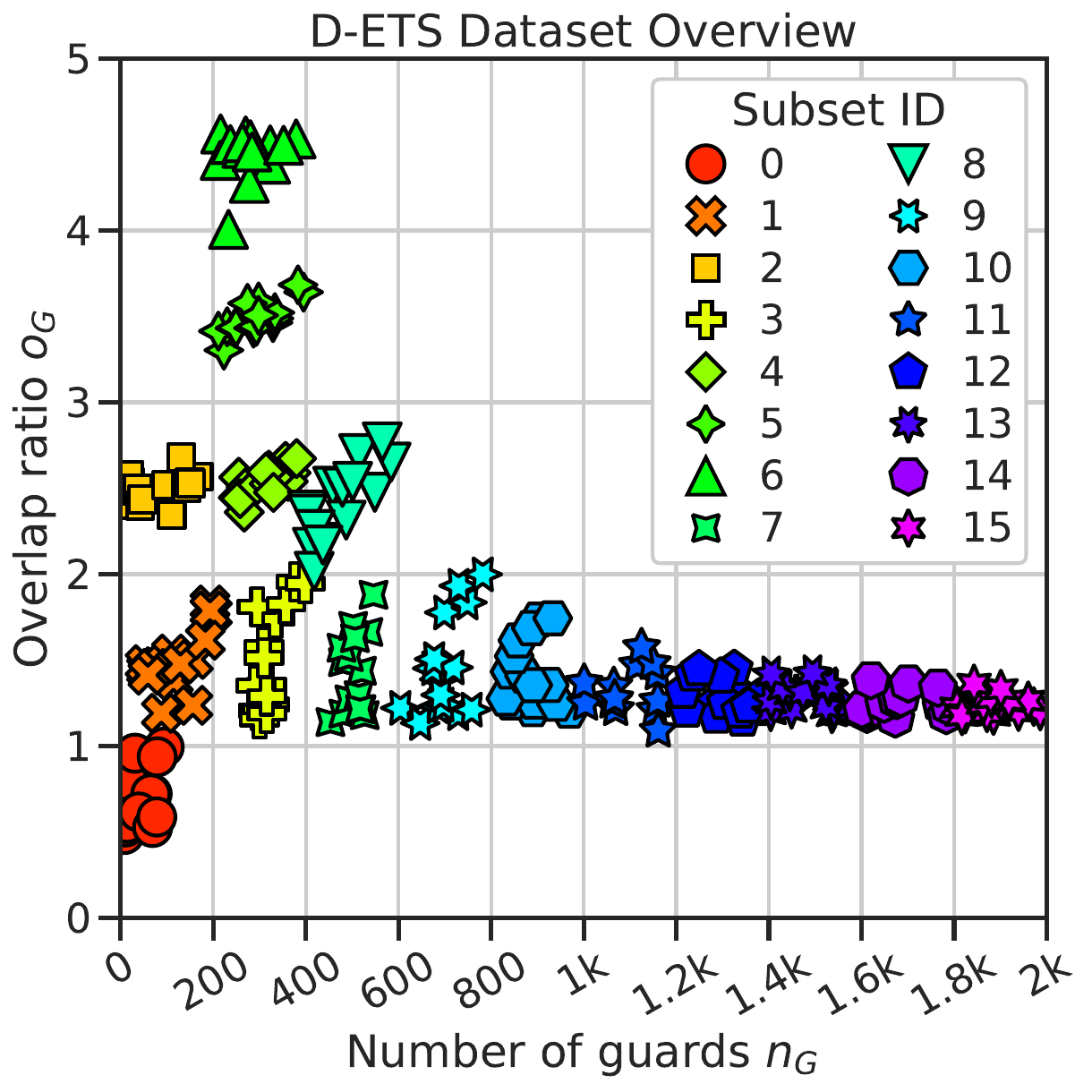}
            \caption{Metrics overview}
            \label{fig:dataset-overview}
        \end{subfigure}
        \hfill
        \begin{subfigure}[t]{0.325\columnwidth}
            \centering
            \includegraphics[width=\linewidth]{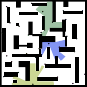}
            \caption{0-11: 31, 1.0}
            \label{fig:inst00-11}
        \end{subfigure}
        \hfill
        \begin{subfigure}[t]{0.325\columnwidth}
            \centering
            \includegraphics[width=\linewidth]{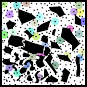}
            \caption{3-4: 316, 1.2}
            \label{fig:inst03-04}
        \end{subfigure}
        \\\vspace{2pt}
        \begin{subfigure}[t]{0.325\columnwidth}
            \centering
            \includegraphics[height=\linewidth]{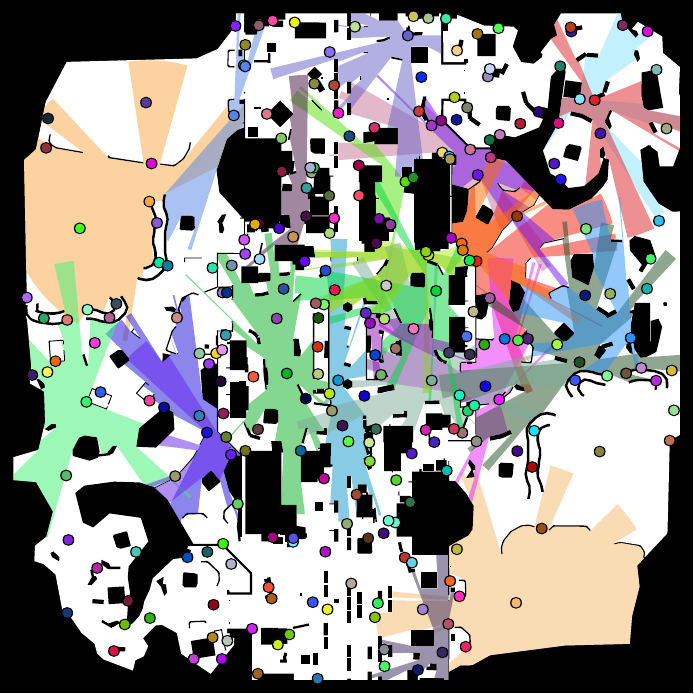}
            \caption{6-8: 262, 4.5}
            \label{fig:inst06-03}
        \end{subfigure}
        \hfill
        \begin{subfigure}[t]{0.325\columnwidth}
            \centering
            \includegraphics[height=\linewidth]{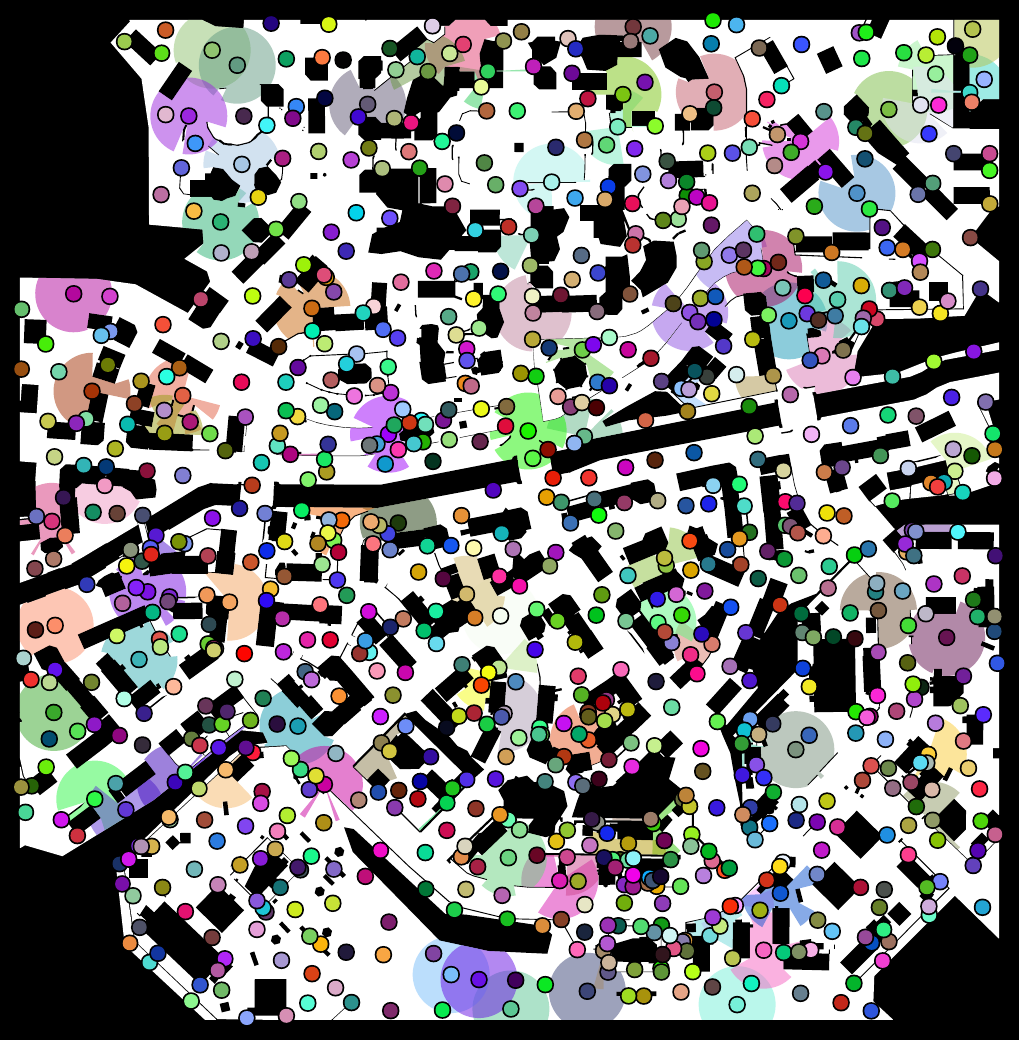}
            \caption{10-14: 933, 1.7}
            \label{fig:inst10-14}
        \end{subfigure}
        \hfill
        \begin{subfigure}[t]{0.325\columnwidth}
            \centering
            \includegraphics[height=\linewidth]{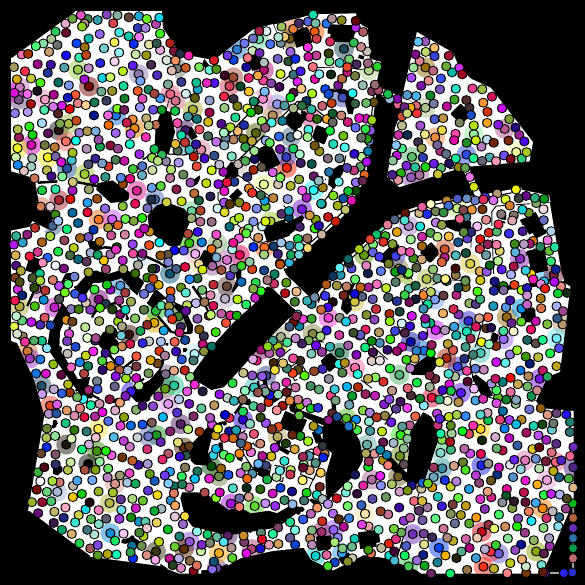}
            \caption{15-5: 1,986, 1.2}
            \label{fig:inst15-13}
        \end{subfigure}
        \caption{
            Dataset metrics overview~(\subref{fig:dataset-overview}) and example instances~(\subref{fig:inst00-11}--\subref{fig:inst15-13}) showing the map, all guards, and 10\% of visibility regions.
            Captions use the format SubsetID-InstanceID: $n_G$, $o_G$.
        }
        \label{fig:dataset}
    \end{figure}

    We generate a dataset of 240 \dsearch{} instances in the form $(\mathcal{W}, r_{\mathrm{vis}}, G)$, which also serves as \search{} instances by excluding $G$.
    The polygonal environments $\mathcal{W}$ are sourced from 22 maps in our private collection (OURS dataset\footnote{We conceal our identity for the double-blind review process.}) and 35 from the publicly available, large-scale dataset of the video game Iron Harvest, developed by KING Art Games and introduced to the research community in~\citep{Harabor2022} (IH dataset).
    Details on preprocessing and map characteristics, including the number of vertices, holes, and dimensions, are provided in Appx.~\ref{app:dataset}.
    The visibility range $r_{\mathrm{vis}}$ and guard set $G$ were generated using the process described in Appx.~\ref{app:dataset}, resulting in 240 \dsearch{} instances characterized by:
    \begin{flalign*}
        n_G \coloneqq |G|, \quad o_G \coloneqq \frac{\sum_{g \in G}\mathrm{Area}(\mathrm{Vis}(g))}{\mathrm{Area}(\bigcup_{g \in G}\mathrm{Vis}(g))} - 1,
    \end{flalign*}
    where $n_G$ represents the number of guards, and $o_G \cop{\in} \mathbb{R}_{\geq 0}$ is the overlap ratio, measuring the overlap between the guards' visibility regions ($o_G \cop{=} 0$ indicates no overlaps).
    The dataset is divided into 16 subsets of 15 instances, grouped by $n_G$ and $o_G$.
    Appx.~\ref{app:dataset} includes a summary table, and Fig.~\ref{fig:dataset} illustrates the metric distribution and provides examples.
    Three instances (IDs 2, 7, 12) were excluded from each subset for preliminary, informal experiments, leaving 12 instances (192 in total) for the evaluation presented here.
    To complete the benchmark, five sensor starting configurations $g_0 \cop{\in} \mathcal{W}$ were selected per instance: $(0,0),\allowbreak{}(x,0),\allowbreak{}\frac{1}{2}(x,y),\allowbreak{}(0,y),\allowbreak{}(x,y)$, where $x$ and $y$ are the map's width and height.\footnote{
        If $g_0$ fell inside an obstacle, it was shifted to the nearest vertex of the environment.
    }
    This results in 1,200 \dsearch{} instances (960 for evaluation).

    \subsubsection{Evaluated Framework Methods}

    Alg.~\ref{alg:milaps-replanning} is parameterized by a replanning schedule, defined as a sequence of boolean variables $\langle \mathrm{replan}_{i} \rangle_{i=1}^{n}$.
    The schedule is controlled by two parameters: the number of replanning steps $\mathrm{cnt} \cop{\in} \{1{:}n\}$ and the replanning period multiplier $\mathrm{coeff} \cop{\in} \mathbb{R}_{\geq 1}$.
    $\mathrm{cnt}$ specifies the total number of replanning steps, while $\mathrm{coeff}$ controls their distribution---larger values concentrate steps earlier, spacing them out progressively (when $\mathrm{coeff} \cop{=} 1$, the replanning intervals remain constant).
    The replanning sequence is given by $\mathrm{replan}_i \cop{=} \llbracket i \cop{\in} \mathrm{RS} \rrbracket$, where $\mathrm{RS}$ is the set of replanning indices: $\mathrm{RS} \cop{\coloneqq} \{i_k \cop{:} k \cop{\in} \{1{:}n\} \cop{\land} i_k \cop{\leq} n\}$, where $i_k$ represents the index of the $k$-th replanning step.
    The value of $i_k$ is computed based on the initial replanning period $T_0$ and the value of $\mathrm{coeff}$:
    \begin{flalign*}
        T_0 \coloneqq \frac{n + 1}{\sum_{j=0}^{\mathrm{cnt}} \mathrm{coeff}^{j}}, \quad i_k \coloneqq \bigg\lfloor {\sum}_{j=0}^{k} T_0 \mathrm{coeff}^{j} \bigg\rceil,
    \end{flalign*}
    where $\lfloor {.} \rceil$ rounds to the nearest integer.

    We encode method variants using the following naming convention: \milaps{}-$\mathrm{wtype}$[-R$\mathrm{cnt}$[-$\mathrm{coeff}$][-$\mathrm{sens}$]][+].
    The brackets indicate optional components, and the conventions are as follows:
    \begin{enumerate*}[label=(\alph*)]
        \item $\mathrm{wtype}$: Weight type in Alg.~\ref{alg:milaps} and~\ref{alg:milaps-replanning} (Const, Vis, DisSplit, DisMaxW, or DisGreedy).
        \item $\mathrm{cnt}$: Number of replanning steps.
        If omitted, defaults to Alg.~\ref{alg:milaps}; otherwise, uses Alg.~\ref{alg:milaps-replanning}.
        \item $\mathrm{coeff}$: Replanning period multiplier.
        Defaults to $\mathrm{coeff} \cop{=} 1$ if omitted.
        \item $\mathrm{sens}$: Sensing policy in Alg.~\ref{alg:milaps-replanning} ($\text{\search{}}$ or $\text{\dsearch{}}$ if omitted).
        \item +: Optional step inclusion.
        For \dsearch{}, refers to line~\ref{alg:milaps:bestdetms} in Alg.~\ref{alg:milaps}; for \search{}, line~\ref{alg:decoupling:final} in Alg.~\ref{alg:decoupling}. Presence of + indicates inclusion; omission denotes exclusion.
    \end{enumerate*}

    Finally, we take advantage of the anytime property of Ms-GVNS and enforce linear scaling of the computational budget with the number of guards: $t_{\mathrm{max}} \cop{=} n_G \cop{/} 10$ (in seconds).
    We retain the remaining Ms-GVNS parameters from~\citep{Mikula2022}.

    \subsubsection{Baseline Methods}

    As baselines, we consider the adaptive-depth utility greedy algorithm (UGreedy-A) from the seminal work on \search{}~\citep{Sarmiento2003} and its single-step variant (UGreedy-1).
    These methods are allocated twice the computational budget ($2t_{\mathrm{max}}$) before termination, recording a \emph{timeout} if exceeded.
    We also include the original metaheuristic approaches~\citep{Kulich2014,Kulich2017}, whose key concepts are integrated into our framework.
    These methods correspond to specific parameterizations: \milaps{}-Const~\citep{Kulich2014} and \milaps{}-Vis~\citep{Kulich2017}.

    \subsubsection{Implementation Details}

    The proposed methods and baselines are implemented in C++ with the {C++17} standard, utilizing a shared codebase.
    The implementation is single-threaded and compiled in Release mode using GCC compiler version 12.3.0.
    It was executed on a Lenovo Legion 5 Pro 16\allowbreak{}IT\allowbreak{}H6H laptop equipped with an Intel Core i7-11800H processor (4.60GHz), 16GB of RAM, and running Ubuntu 20.04 LTS\@.

    \subsubsection{Evaluation Metrics}

    We evaluate the methods using two metrics: the objective value $\objective{}(\tau,\mathrm{sens})$, where $\mathrm{sens}$ depends on the problem (\dsearch{} or \search{}), and the CPU time $t$.
    To ensure comparability across instances and methods, metrics are \emph{normalized} by the best-known solution (BKS) and time budget $t_{\mathrm{max}}$: $\mathrm{Gap}(\objective{}) \cop{\coloneqq} 100\%\frac{\objective{} - \mathrm{BKS}}{\mathrm{BKS}}$ and $\mathrm{Rel}(t) \cop{\coloneqq} \frac{t}{t_{\mathrm{max}}}$.
    For \milaps{}, we also record the runtimes and routes of each intermediate best solution found by Ms-GVNS, enabling us to plot solution quality as a function of runtime.

    \subsection{Results and Discussion}

    \subsubsection{Parameter Tuning}

    In preliminary informal experiments on a separate dataset subset, disjoint from the main evaluation, we determined representative replanning parameters for the \milaps{} methods.
    \milaps{}-Vis and \milaps{}-DisSplit performed best with 32 replanning steps and a coefficient of 1.05, while \milaps{}-DisMaxW and \milaps{}-DisGreedy achieved the best results with 8 steps and a coefficient of 1.25.
    Increasing the number of replanning steps significantly reduced solution quality, as the limited computational budget was distributed less efficiently.

    \subsubsection{Main Evaluation---\dsearch{} Objective}

\begin{figure*}
        \raggedright
        \includegraphics[height=0.30556754423\textwidth]{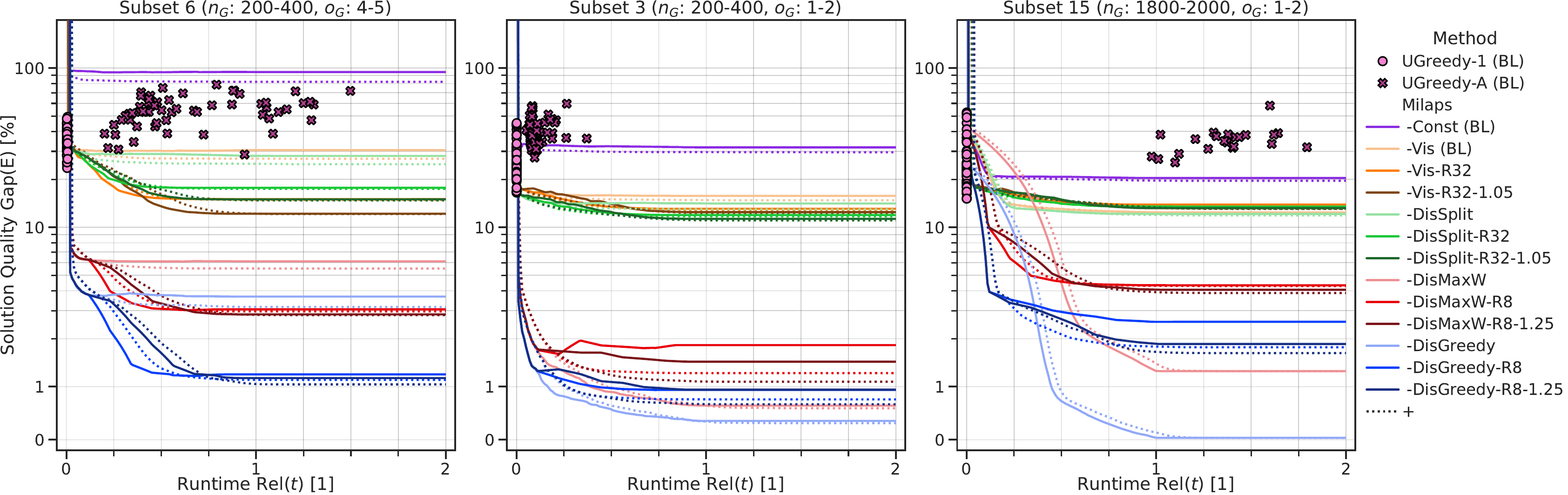}
        \caption{\dsearch{} results for subsets 6, 3, and 15.}
        \label{fig:dets-results}
    \end{figure*}

    We tested 28 methods in total: UGreedy-\{1\textbf{\textbar{}}\allowbreak{}A\}\,(2), \milaps{}-Const\allowbreak{}[+]\,(2), \milaps{}-\{Vis\textbf{\textbar{}}\allowbreak{}DisSplit\}\allowbreak{}[-R32[-1.05]]\allowbreak{}[+]\,(12), and \milaps{}-\{DisMaxW\textbf{\textbar{}}\allowbreak{}DisGreedy\}\allowbreak{}[-R8[-1.25]]\allowbreak{}[+]\,(12).
    Experiments were conducted on all 16 subsets, each containing 60 instances, with each method run once per instance.
    Fig.~\ref{fig:dets-results} presents results for representative subsets 6, 3, and 15, where (6, 3) examines the impact of the overlap ratio with similar guard counts, while (3, 15) evaluates scalability with an increasing number of guards.
    The plots show runtime in the interval $[0, 2t_{\mathrm{max}}{=}n_G/5]$ mapped onto $[0,2]$ on the horizontal axis (linear scale) and the percentage BKS gap on the vertical axis (linear scale between 0 and 1\%, logarithmic scale above 1\%).
    The baseline methods are marked as BL in the legend, and subset characteristics are provided in the plot titles.
    The \milaps{} runs are aggregated using mean values along the runtime axis, while the UGreedy runs are shown individually.
    UGreedy-A timed out 4, 11, 19, and 38 times on subsets 12, 13, 14, and 15, respectively, which is not shown in the plots.

    The key result is that \milaps{}-DisGreedy consistently outperforms all other methods, achieving near-zero BKS gaps across all subsets, which is approximately 10--20\% better than the best baseline at the time limit.
    The basic variant excels for smaller overlap ratios, while the replanning variants provide significant improvements for larger overlap ratios, validating the novel ideas introduced in this work.
    DisMaxW also performs well, exhibiting similar characteristics to DisGreedy, whereas all other methods are less competitive.
    The DisSplit and Vis variants perform similarly but fall significantly behind DisGreedy and DisMaxW, while the Const variant is by far the least effective.
    The utility-greedy baselines typically produce BKS gaps exceeding 20\%.
    While UGreedy-1 returns solutions rapidly, UGreedy-A is considerably slower without improving solution quality.
    The increased runtime of UGreedy-A is due to its lookahead mechanism, and its failure to enhance solution quality may stem from its persistently greedy nature and the pruning of the search space during the lookahead phase.
    These results suggest that assigning static weights to guards based on their visibility regions is an effective global optimization strategy for the computationally expensive \dsearch{} objective, outperforming utility-greedy methods that use dynamic weights to address the same objective directly.
    However, the choice of the static weight assignment method is critical for solution quality.

    Secondary observations include the following:
    \begin{enumerate*}[label=(\roman*)]
        \item The \milaps{} framework introduces some runtime overhead, leading to delayed convergence beyond $t_{\mathrm{max}}$, particularly for larger instances.
        \item Distributing more replanning steps toward the beginning of the route slightly improves solution quality for larger overlap ratios, with the strongest effect on \milaps{}-Vis and less impact on top-performing variants.
        \item Evaluating intermediate solutions against the true \dsearch{} objective (the `+' variants) enhances solution quality but slows convergence, with diminishing benefits for the best-performing methods.
        Without this feature, some methods may finalize worse solutions than certain intermediate ones, as observed for DisMaxW-R8 on subset 3, though this does not affect top-performing variants.
        \item For smaller overlap ratios, the replanning variants of DisGreedy may yield better early solutions but eventually converge to worse ones than the basic variant.
        \item Larger instances require more relative time to converge.
    \end{enumerate*}

    \subsubsection{Main Evaluation---\search{} Objective}

\begin{figure*}
        \raggedright
        \includegraphics[width=\textwidth]{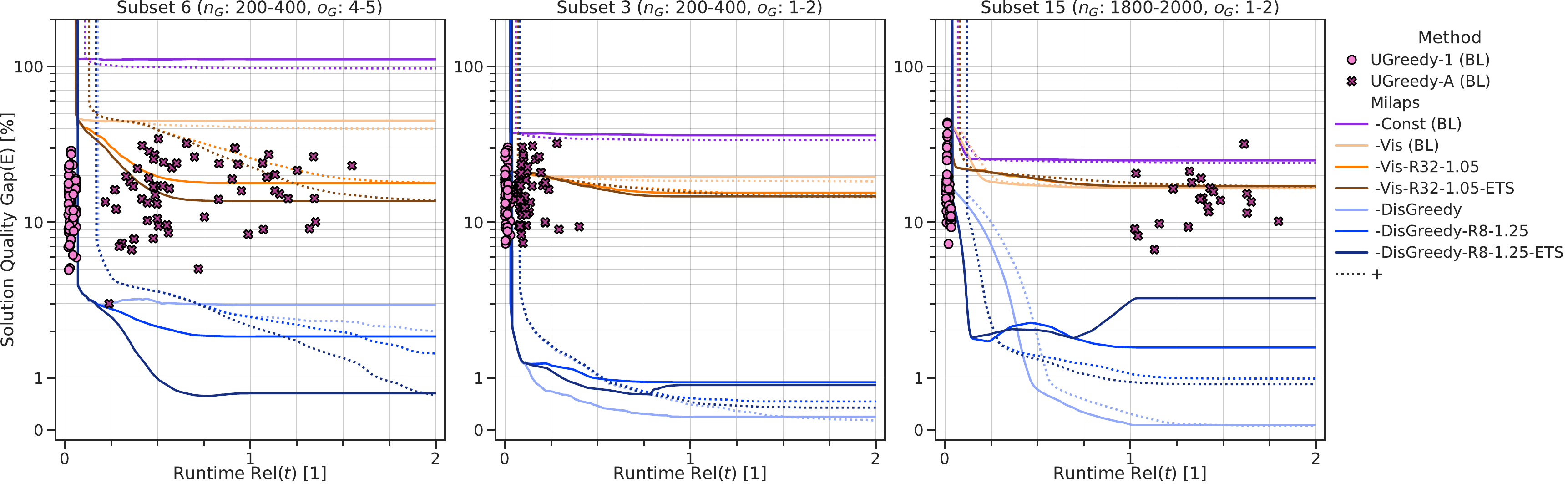}
        \caption{\search{} results for subsets 6, 3, and 15.}
        \label{fig:ets-results}
    \end{figure*}

    We conducted a similar evaluation for the computationally even more expensive \search{} objective, testing a representative subset of methods from the \dsearch{} evaluation alongside additional variants incorporating the \search{} sensing policy during replanning (`\mbox{-\search{}}').
    In total, we evaluated 16 methods: UGreedy-\allowbreak{}\{1\textbf{\textbar{}}\allowbreak{}A\}\,(2), \milaps{}-\allowbreak{}Const\allowbreak{}[+]\,(2), \milaps{}-\allowbreak{}Vis\allowbreak{}[-R32-1.05[-\search{}]]\allowbreak{}[+]\,(6), and \milaps{}-\allowbreak{}DisGreedy\allowbreak{}[-R8-1.25[-\search{}]]\allowbreak{}[+]\,(6).
    The~measured runtime now includes guard generation.
    Results are in Fig.~\ref{fig:ets-results}.

    The best-quality solutions are again achieved by \milaps{}-DisGreedy, which consistently outperforms all other methods, following similar trends to the \dsearch{} evaluation: the basic variant excels for small overlap ratios, while the replanning variants improve solutions for larger overlap ratios.
    A new observation is that incorporating the \search{} sensing policy during replanning further enhances performance for larger overlap ratios, validating its inclusion in the framework.
    Additionally, evaluating intermediate solutions against the true \search{} objective now imposes a significant computational burden without notable benefits for the best-performing methods.
    This is a key insight: the proposed framework effectively optimizes the \search{} objective without ever evaluating it directly.
    Naturally, this approach has inherent limitations, and a hypothetical metaheuristic---though none has been proposed yet---capable of directly optimizing \search{} through dynamic weight adjustments might achieve superior solution quality.
    However, such a method would likely be practical only for small instances, with scalability to large-scale settings, as considered in this work, remaining a major challenge.

    \subsubsection{Ablation Study (Sensor-Placement Method)}

\begin{figure}[t]
        \centering
        \includegraphics[width=0.75\columnwidth]{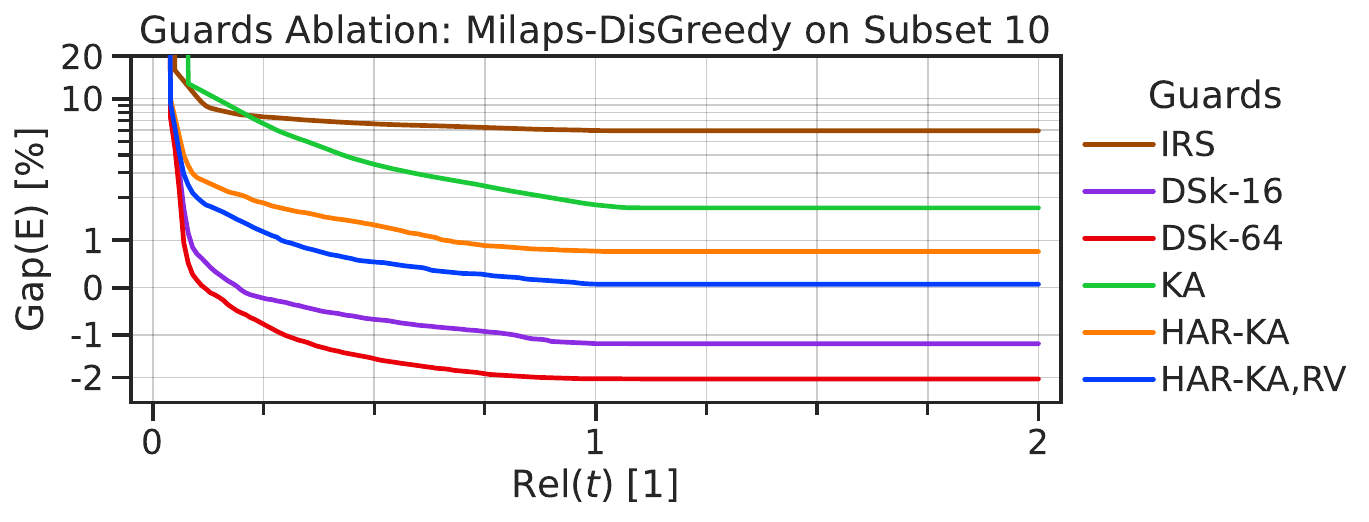}
        \caption{Results of the SPP ablation study for subset 10.}
        \label{fig:ets-ablation}
    \end{figure}

    The final part of our quantitative evaluation examines the impact of the selected SPP method, HAR-KA,RV, by replacing it with alternative guard generation approaches to assess their effect on the \search{} objective.
    We consider four alternatives, all previously evaluated for minimum guard generation in~\citep{Mikula2024}: \emph{informed random sampling} (IRS, placing new guards in uncovered regions), \emph{dual sampling} (DS)~\cite{Gonzalez-Banos1998}, the standalone KA method~\cite{Kazazakis2002}, and a refined KA variant (HAR-KA)~\cite{Mikula2024}.
    DS is tested in two variants, using 16 and 64 dual samples, respectively.
    The study is conducted with a fixed \milaps{}-DisGreedy method on dataset subsets 3, 6, 10, and 15, yielding consistent results.
    Subset 10 is selected as representative for presentation in Fig.~\ref{fig:ets-ablation}.

    The results show that HAR-KA,RV is slightly outperformed by the DS methods, which achieved negative gaps of \mbox{\textminus{}1\%} and \mbox{\textminus{}2\%} for 16 and 64 dual samples, respectively.
    These negative values arise because BKSs were taken exclusively from the main evaluation, where HAR-KA,RV was used.
    This is a noteworthy finding, suggesting that minimizing the number of guards does not necessarily lead to better \search{} solutions, even though smaller \dsearch{} instances are generally easier to optimize under limited runtime budgets.
    The key reason is that the decoupling scheme in our framework is inherently suboptimal---a common property of problem decoupling---whereas an optimal \search{} solution would require a fully coupled approach integrating guard generation and route optimization.
    However, achieving such a coupled method remains an open challenge in \search{} optimization.
    Nonetheless, our results suggest that DS-generated guard sets better approximate an ideal coupled solution than those from HAR-KA,RV, offering valuable insights for future research and applications.

    \section{Qualitative Study}
    \label{sec:qualitative-evaluation}

    Our qualitative study demonstrates the flexibility of the proposed framework across diverse scenarios, providing visual examples alongside true objective function values.
    We primarily use a single environment from the OURS dataset, \textsc{jari-huge}, a moderately complex indoor environment with rooms, corridors, and open areas.
    The map measures $21\,\mathrm{m} \cop{\times} 23\,\mathrm{m}$, featuring 279 vertices, 9 holes, and a free space area of $459\,\mathrm{m}^2$.
    The initial configuration $g_0$ is set at the map's center, while other scenario parameters vary based on the studied aspect.
    Unless stated otherwise, the solution method is fixed to \milaps{}-DisGreedy.
    We begin with the \classical{} scenario from our quantitative evaluation.

    \subsubsection{Varying the Footprint Radius}

\begin{figure}
        \centering
        \begin{subfigure}[t]{0.325\columnwidth}
            \centering
            \includegraphics[width=\linewidth]{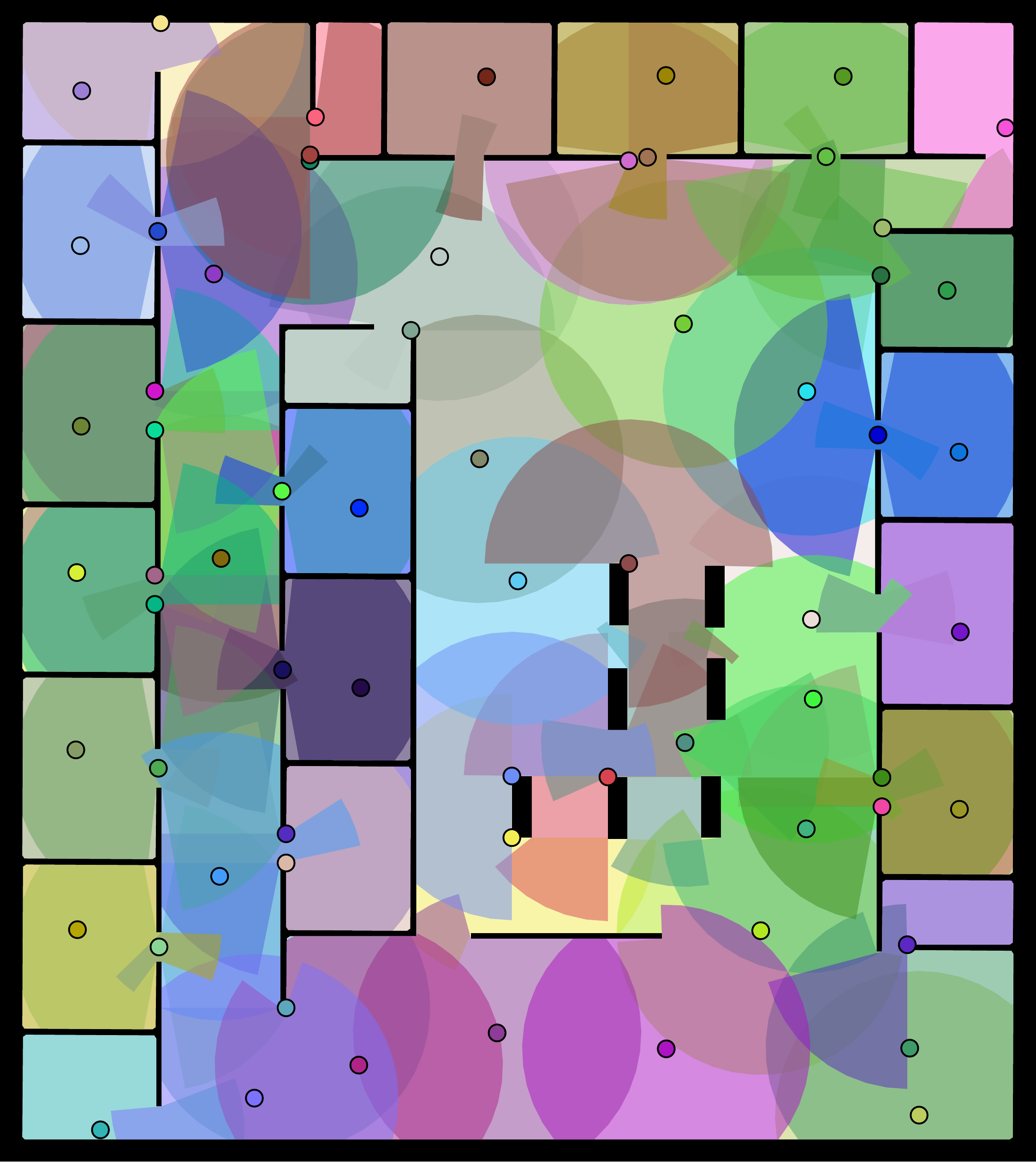}
            \caption{$r_{\mathrm{fp}} \cop{=} 0, \epsilon \cop{=} 10^{-5}$}
            \label{fig:cov-fp0.00}
        \end{subfigure}
        \hfill
        \begin{subfigure}[t]{0.325\columnwidth}
            \centering
            \includegraphics[width=\linewidth]{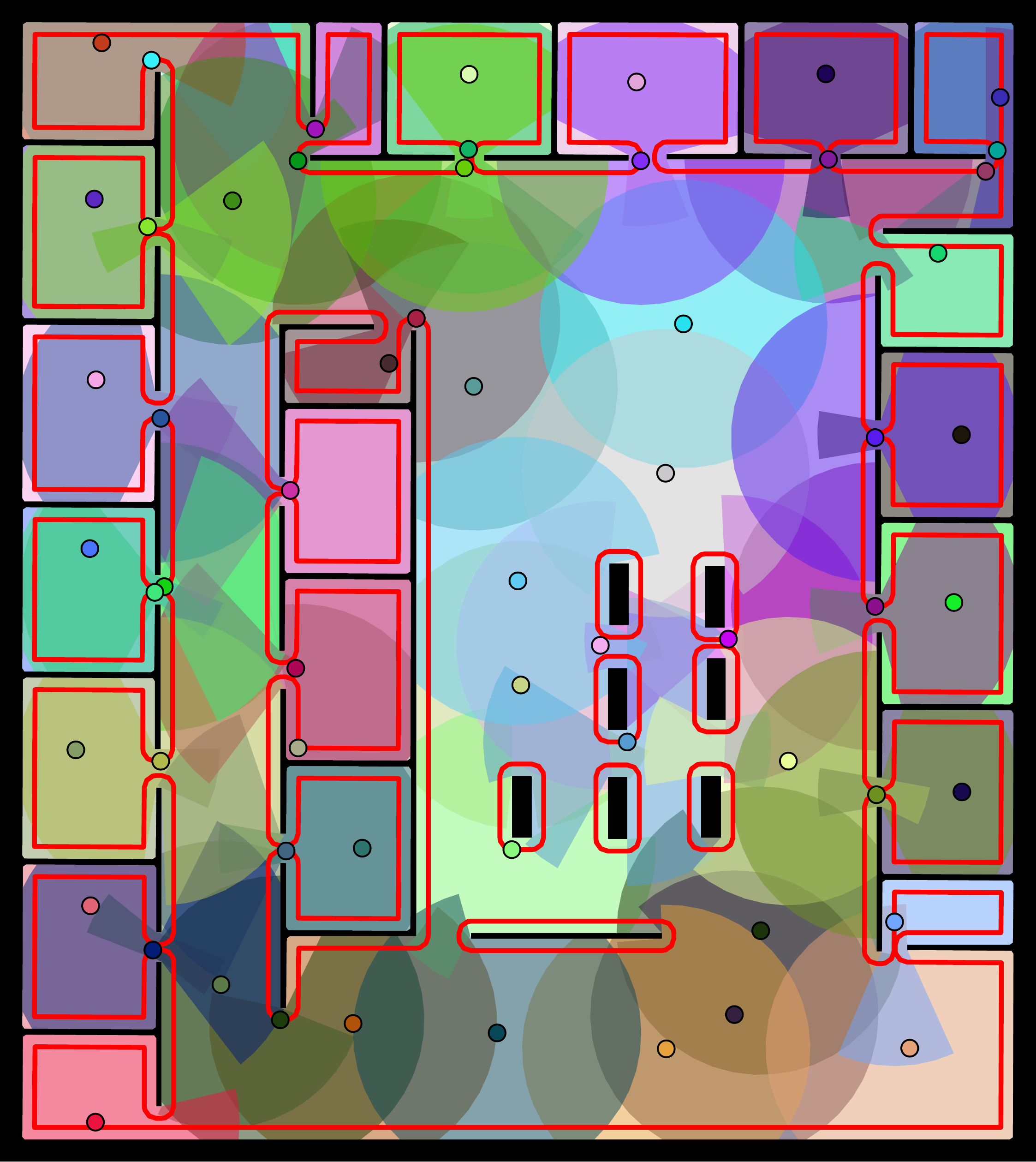}
            \caption{$r_{\mathrm{fp}} \cop{=} 0.25, \epsilon \cop{=} 10^{-5}$}
            \label{fig:cov-fp0.25}
        \end{subfigure}
        \hfill
        \begin{subfigure}[t]{0.325\columnwidth}
            \centering
            \includegraphics[width=\linewidth]{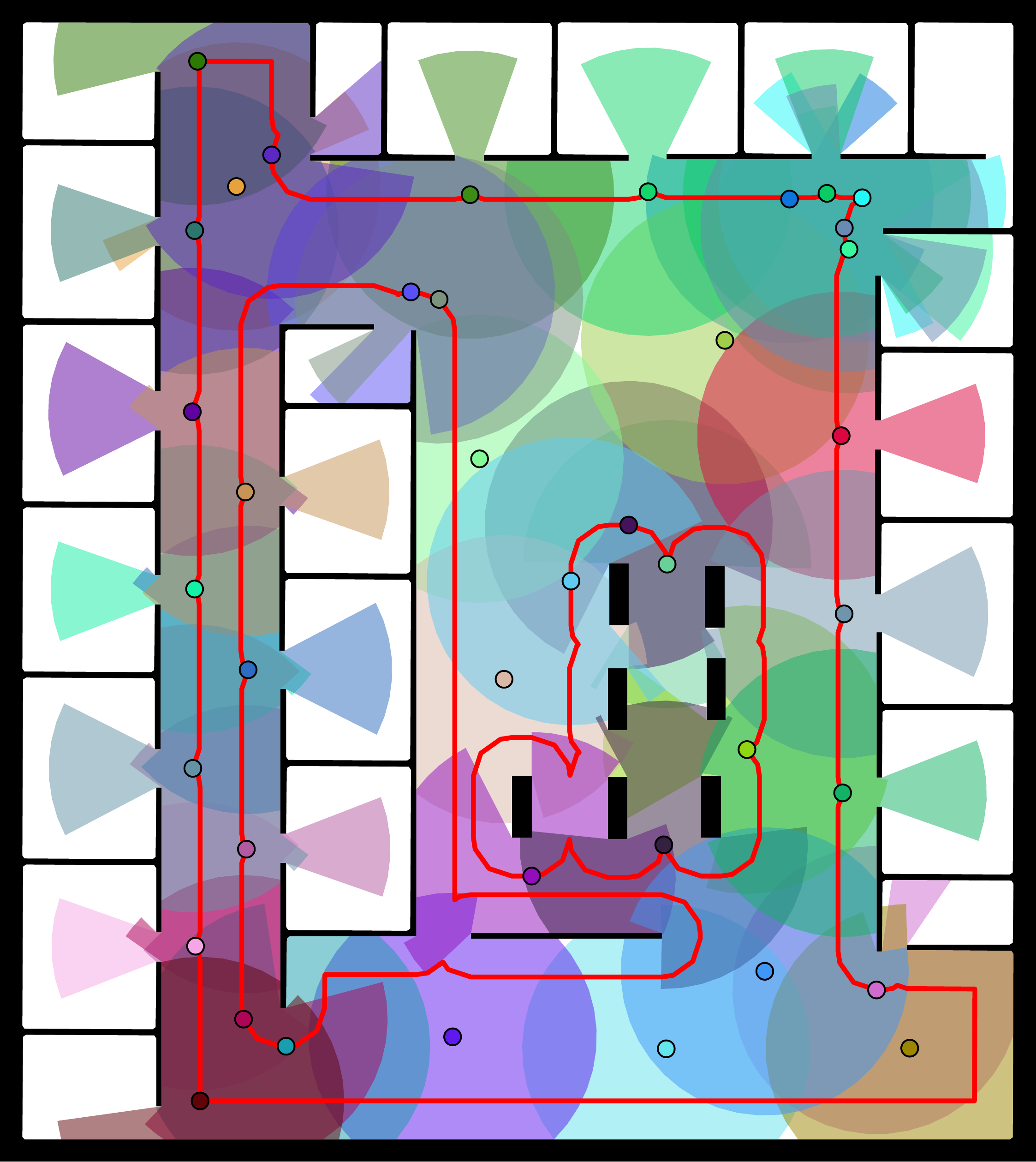}
            \caption{$r_{\mathrm{fp}} \cop{=} 0.8, \epsilon \cop{=} 0.24$}
            \label{fig:cov-fp0.80}
        \end{subfigure}
        \caption{
            Varying the footprint radius $r_{\mathrm{fp}}$.
        }
        \label{fig:coverage-fp}
    \end{figure}

    We evaluate the framework's handling of nonzero footprint radii $r_{\mathrm{fp}}$, managed by the modified HAR-KA,RV method for the SPP, with visual examples in Fig.~\ref{fig:coverage-fp}.
    The figures depict guards with $r_{\mathrm{vis}} \cop{=} 3\,\mathrm{m}$ and their visibility regions forming the environment's coverage, while the red boundary of $\mathbb{C}_{\mathrm{free}}$ indicates movement constraints.
    For the first two scenarios---one with $r_{\mathrm{fp}} \cop{=} 0$ and the other with $r_{\mathrm{fp}} \cop{=} 0.25\,\mathrm{m}$---we set $\epsilon \cop{=} 10^{-5}$ to achieve $99.999\%$ coverage.
    Both cases show no visible gaps, though the latter lacks formal feasibility guarantees.
    A key factor in this practical success is that no corridor is narrower than $2r_{\mathrm{fp}}$, preserving the topologies of $\mathcal{W}$ and $\mathbb{C}_{\mathrm{free}}$.
    When $r_{\mathrm{fp}}$ increases to $0.80\,\mathrm{m}$, the topology of $\mathbb{C}_{\mathrm{free}}$ changes, leaving some areas inaccessible.
    To compensate, we relax the coverage constraint to $\epsilon \cop{=} 0.24$, achieving $76\%$ coverage.
    As shown in the final figure, the modified HAR-KA,RV method adapts effectively, leveraging the sensor's range to cover some inaccessible areas while maintaining a low guard count.

    \subsubsection{Impact of Turning Costs} 

    \begin{figure}[b]
        \centering
        \begin{subfigure}[t]{0.325\columnwidth}
            \centering
            \includegraphics[width=\linewidth]{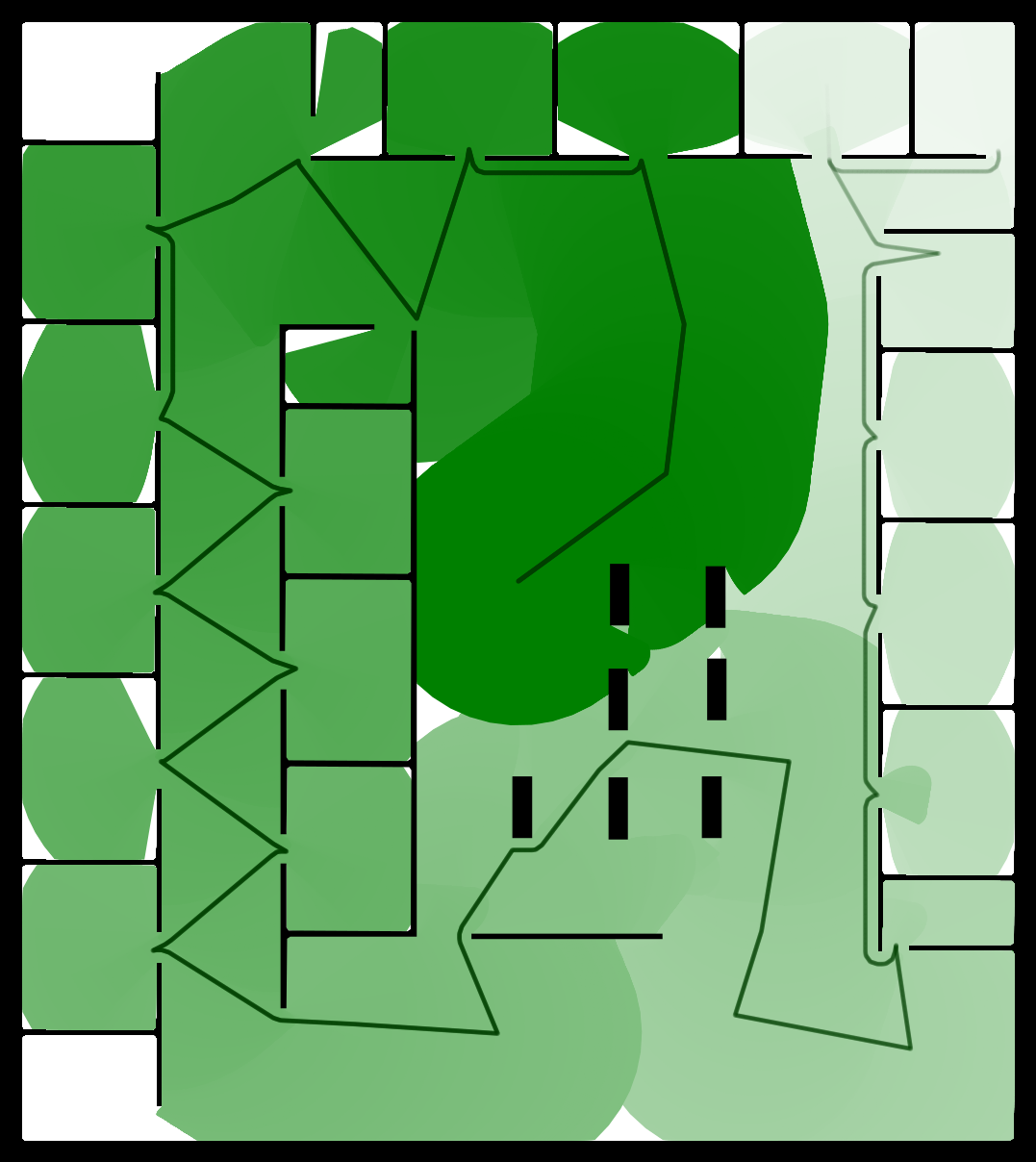}
            \caption{GSP: $\mathrm{ET} \cop{=} 67.7\,\mathrm{s}$}
            \label{fig:gsp-3.0}
        \end{subfigure}
        \hfill
        \begin{subfigure}[t]{0.325\columnwidth}
            \centering
            \includegraphics[width=\linewidth]{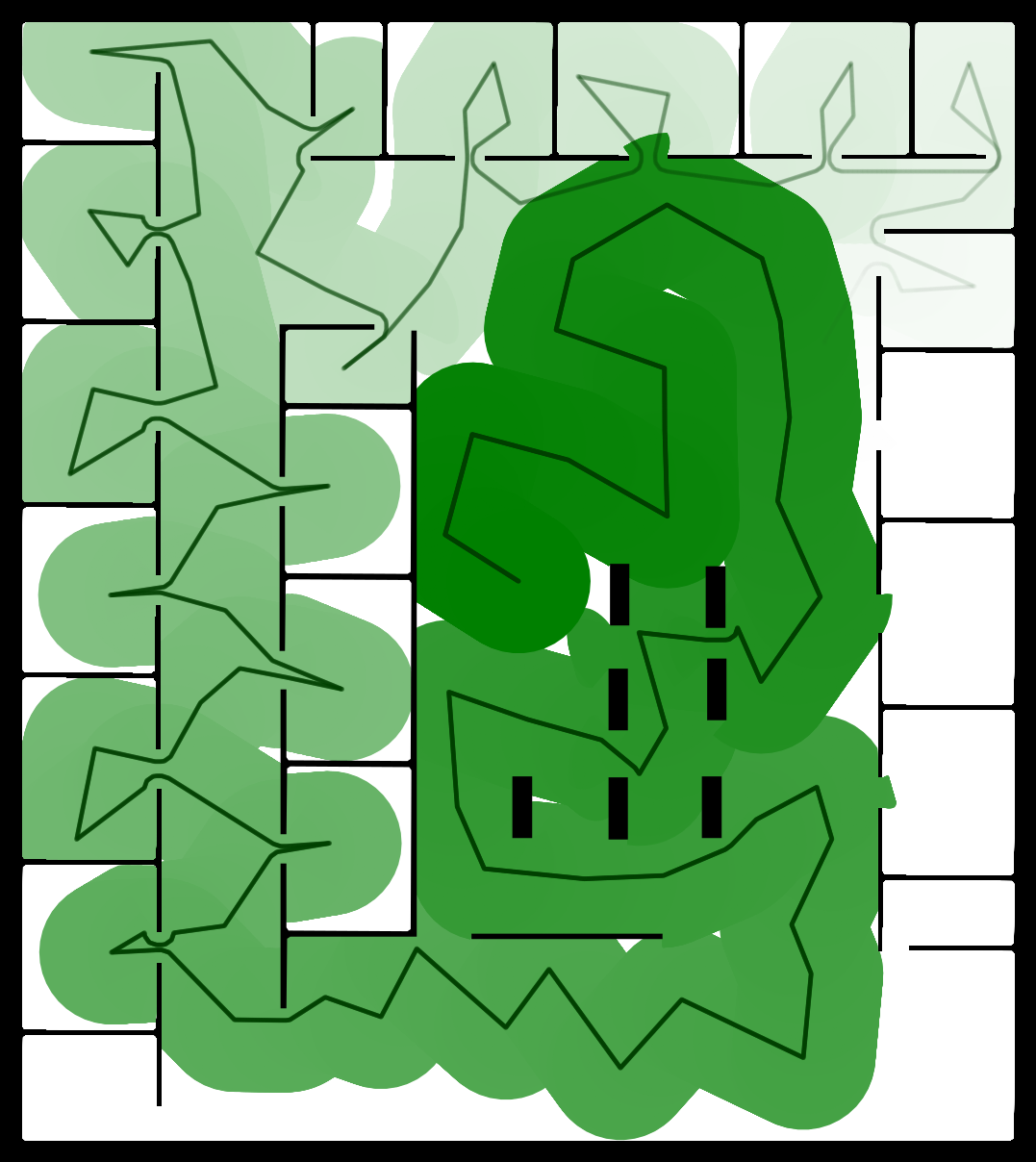}
            \caption{GSP: $\mathrm{ET} \cop{=} 155.8\,\mathrm{s}$}
            \label{fig:gsp-1.5}
        \end{subfigure}
        \hfill
        \begin{subfigure}[t]{0.325\columnwidth}
            \centering
            \includegraphics[width=\linewidth]{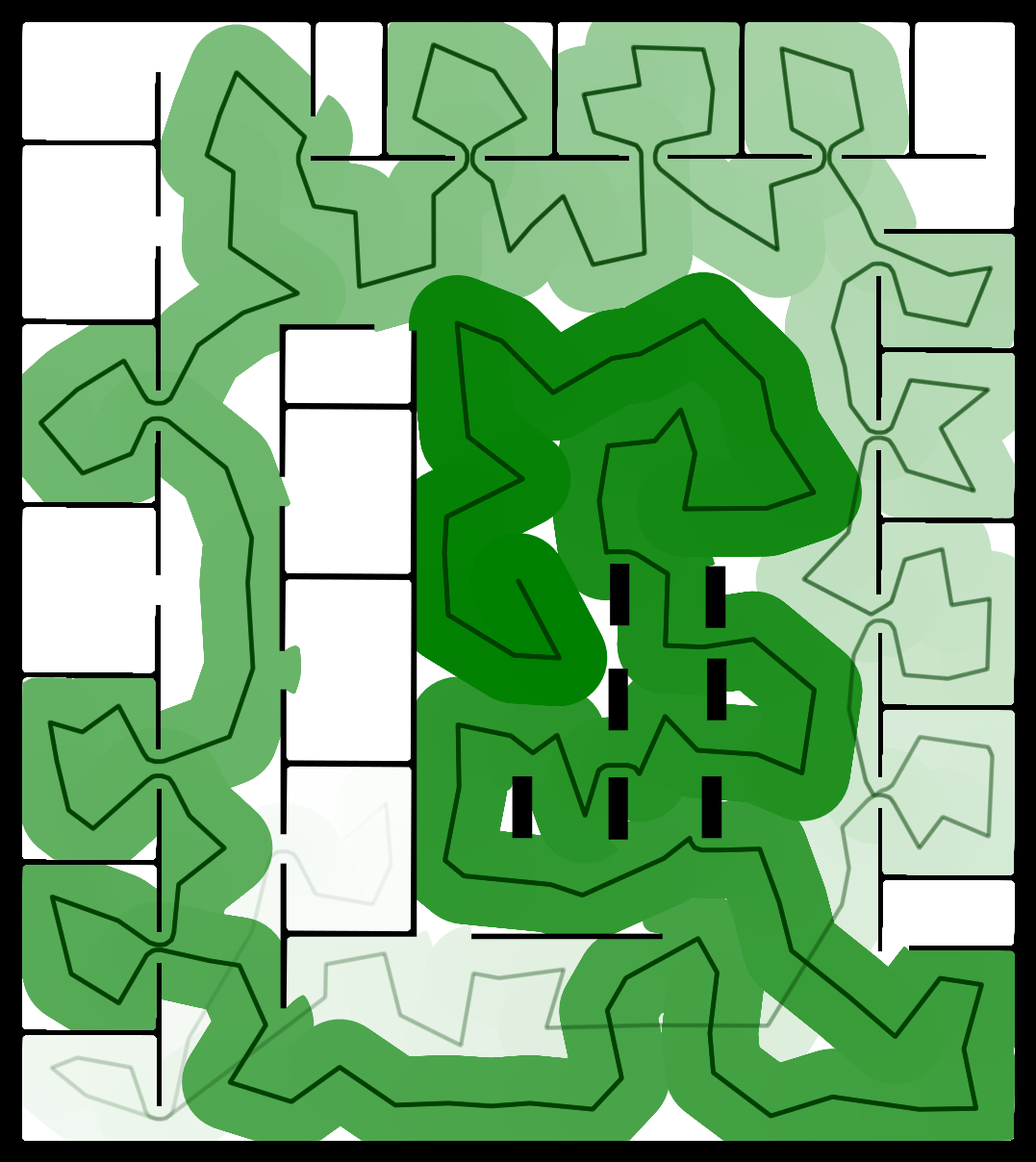}
            \caption{GSP: $\mathrm{ET} \cop{=} 301.8\,\mathrm{s}$}
            \label{fig:gsp-1.0}
        \end{subfigure}
        \\
        \begin{subfigure}[t]{0.325\columnwidth}
            \centering
            \includegraphics[width=\linewidth]{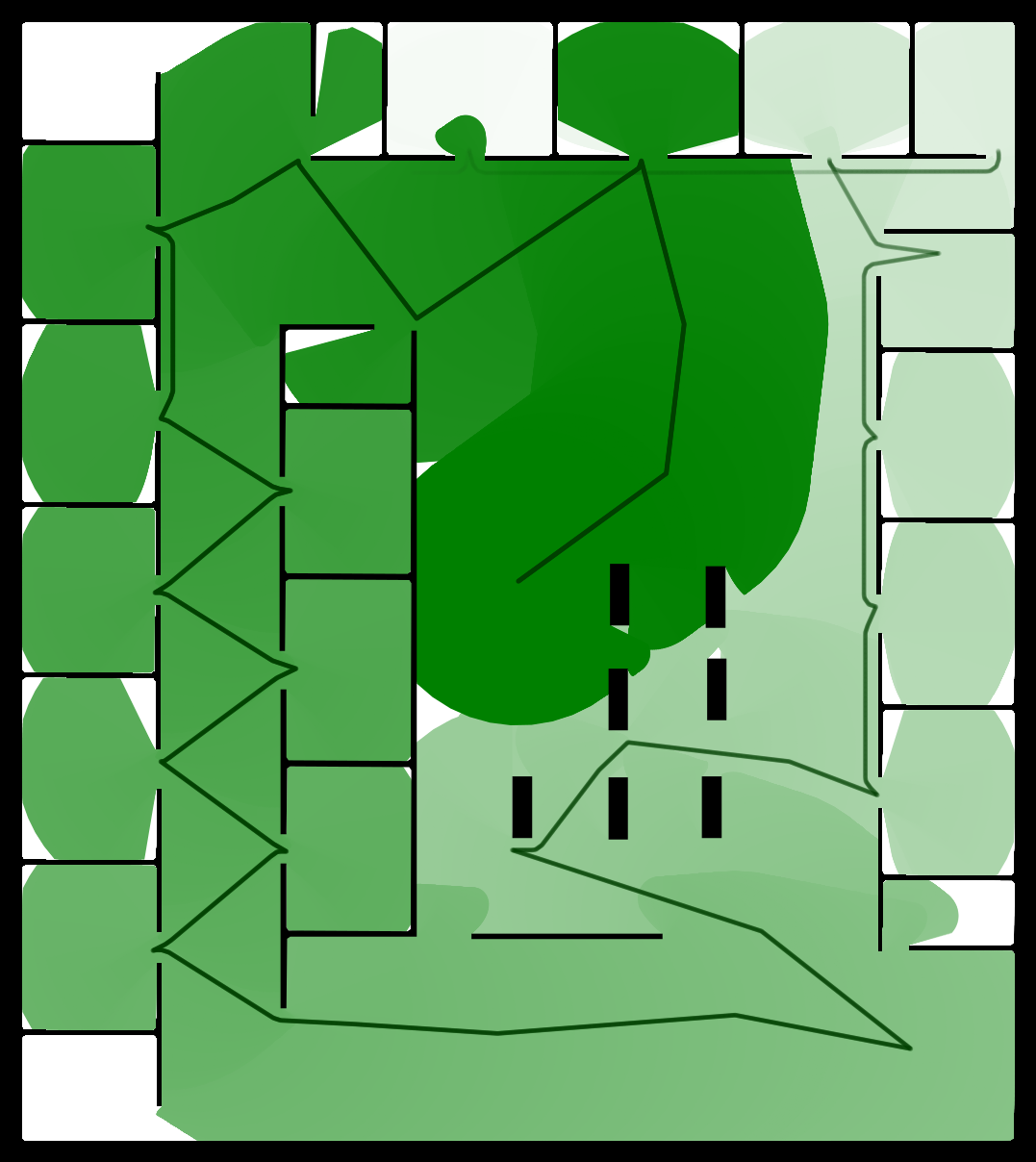}
            \caption{\gspt{}: $\mathrm{ET} \cop{=} 66.0\,\mathrm{s}$}
            \label{fig:gspt-3.0}
        \end{subfigure}
        \hfill
        \begin{subfigure}[t]{0.325\columnwidth}
            \centering
            \includegraphics[width=\linewidth]{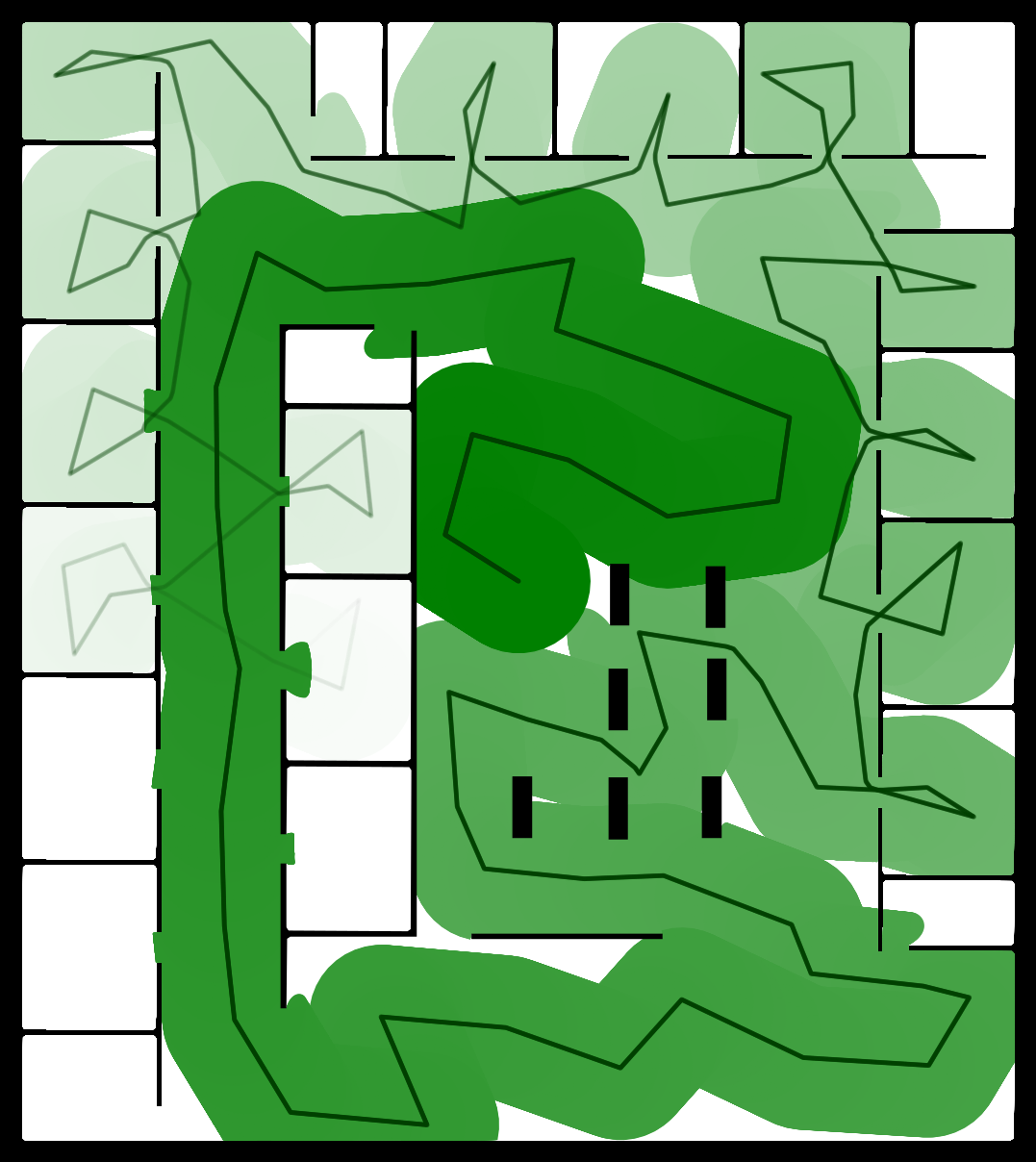}
            \caption{\gspt{}: $\mathrm{ET} \cop{=} 144.7\,\mathrm{s}$}
            \label{fig:gspt-1.5}
        \end{subfigure}
        \hfill
        \begin{subfigure}[t]{0.325\columnwidth}
            \centering
            \includegraphics[width=\linewidth]{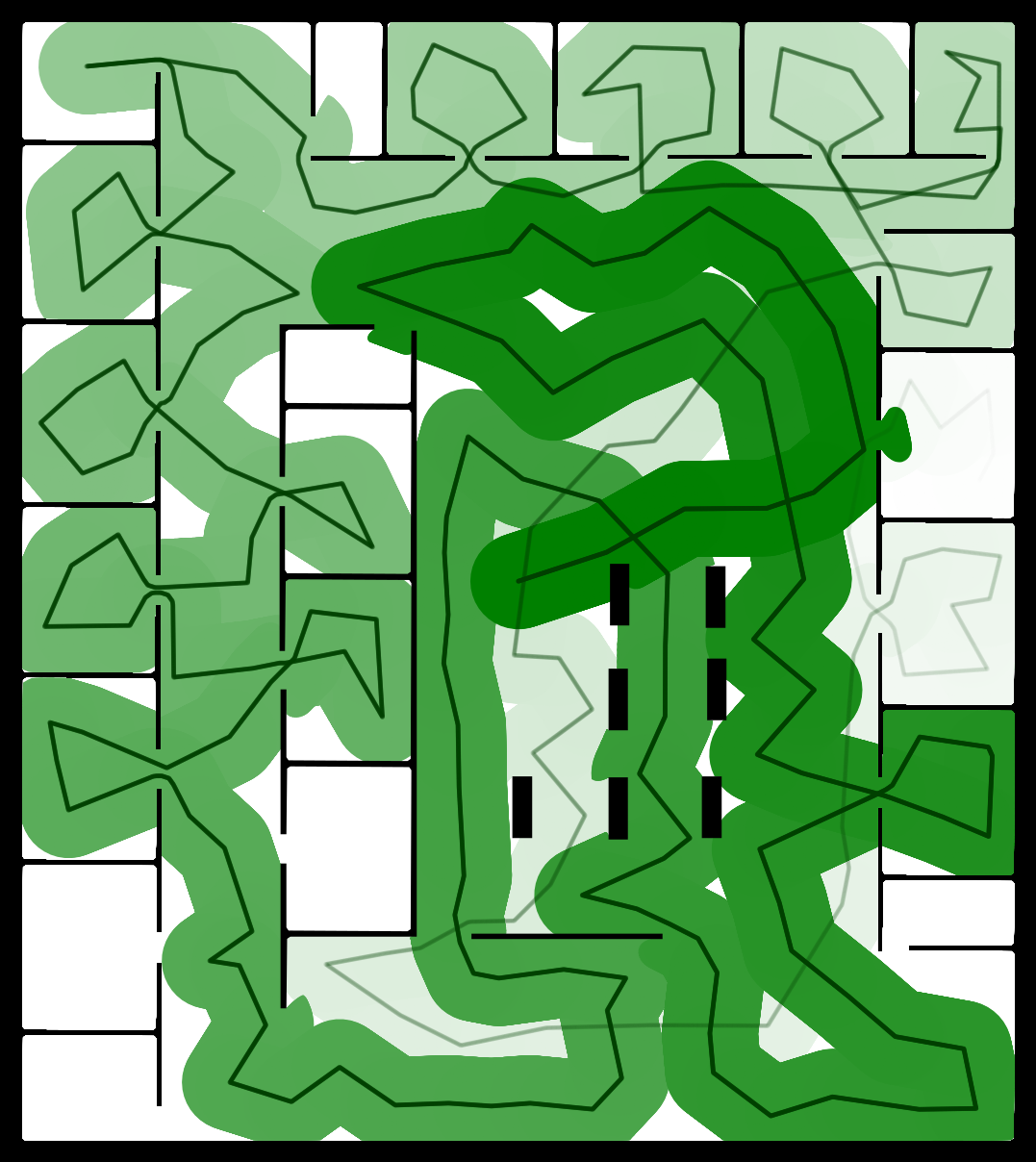}
            \caption{\gspt{}: $\mathrm{ET} \cop{=} 270.1\,\mathrm{s}$}
            \label{fig:gspt-1.0}
        \end{subfigure}
        \caption{
            Impact of turning costs on the solution quality.
        }
        \label{fig:turning-costs}
    \end{figure}

    We examine the impact of turning costs in the optimization process using the \search{} objective ($\mathrm{ET}$).
    We compare two framework variants: the default \gspt{}, which accounts for turning costs, and GSP, which ignores them despite their inclusion in the objective.
    Fig.~\ref{fig:turning-costs} presents three scenarios defined by $(r_{\mathrm{vis}}, t_{\mathrm{lin}}, t_{\mathrm{ang}})$: (\subref{fig:gsp-3.0},\,\subref{fig:gspt-3.0}): $(3, 1, 0.5)$, (\subref{fig:gsp-1.5},\,\subref{fig:gspt-1.5}): $(1.5, 1, 0.5)$, and (\subref{fig:gsp-1.0},\,\subref{fig:gspt-1.0}): $(1, 1, 1)$.
    The inverse values of the travel time parameters, $t_{\mathrm{lin}}^{-1}$ and $t_{\mathrm{ang}}^{-1}$, represent the sensor’s average linear and angular velocities, set to realistic values ($1\,\mathrm{m}{/}\mathrm{s}$, $2\,\mathrm{rad}{/}\mathrm{s}$, and $1\,\mathrm{rad}{/}\mathrm{s}$).
    Routes computed by GSP appear in (\subref{fig:gsp-3.0}--\subref{fig:gsp-1.0}), while \gspt{} routes are in (\subref{fig:gspt-3.0}--\subref{fig:gspt-1.0}), with respective $\mathrm{ET}$ values in the captions.
    The color gradient from green to white represents newly seen regions, with the route shown as a darker polyline, mapped onto the execution time interval $[0, \frac{1}{3} \mathrm{TT}_{\text{\gspt{}}}]$, where $\mathrm{TT}_{\text{\gspt{}}}$ denotes the total traversal time of the respective \gspt{} solution.
    To reduce visual clutter, route endings are omitted, as they primarily cover leftover uncovered areas.
    Results show \gspt{} improving $\mathrm{ET}$ over GSP by 2.5\%, 7.1\%, and 11.7\%, respectively, with the effect increasing for longer routes and higher $t_{\mathrm{ang}} \cop{/} t_{\mathrm{lin}}$ ratios.
    Intuitively, \gspt{} produces qualitatively different routes, favoring straighter paths, particularly at the start of the search.
    This is evident in (\subref{fig:gsp-1.5},\,\subref{fig:gspt-1.5}), where \gspt{} prioritizes the straight corridor on the left before entering its rooms, whereas GSP visits the rooms earlier, incurring higher turning costs.

    \subsubsection{Varying the Object's Probability Distribution} 

    \begin{figure}[b]
        \centering
        \begin{subfigure}[t]{0.325\columnwidth}
            \centering
            \includegraphics[width=\linewidth]{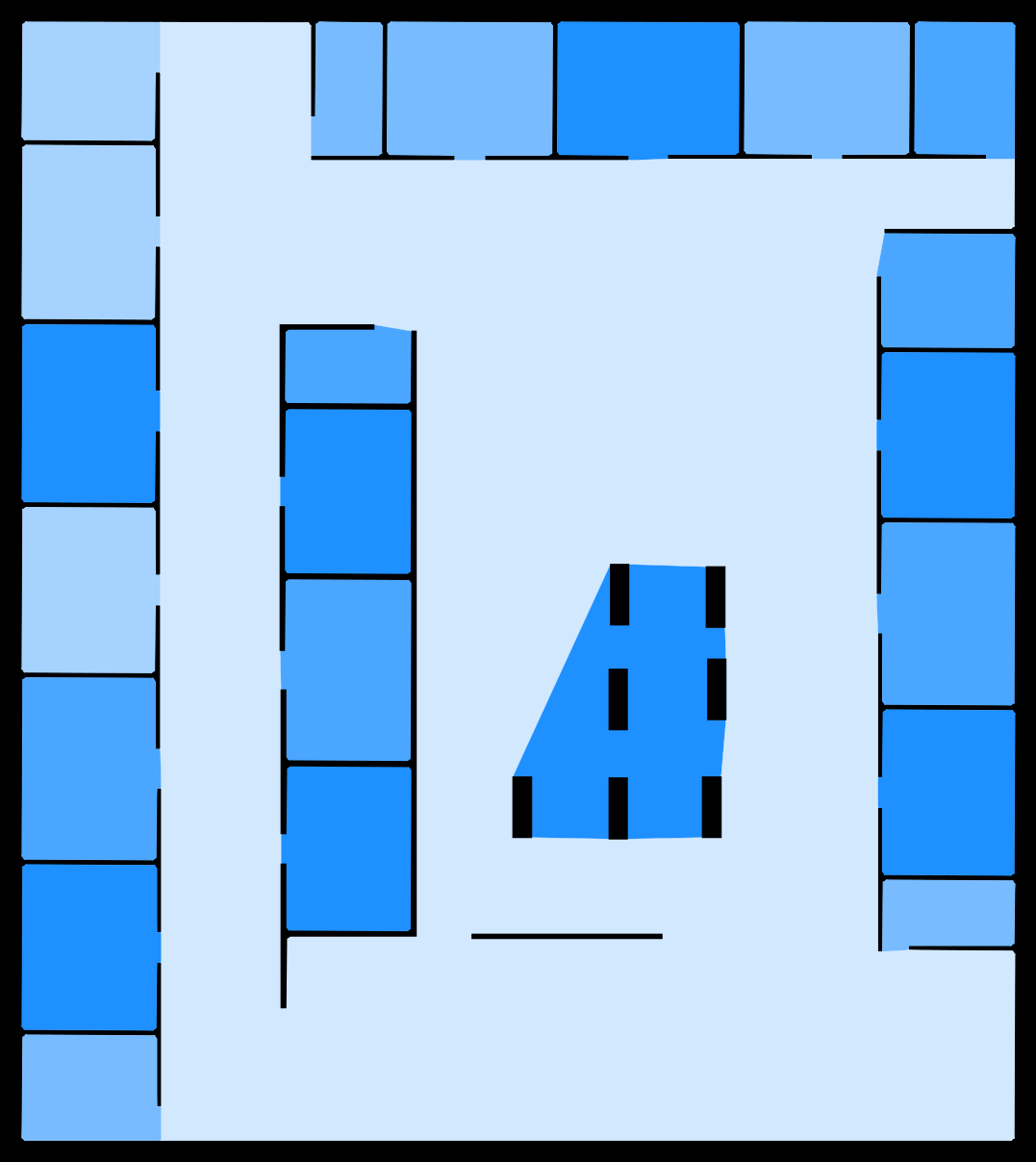}
            \caption{$P_1$, UG1: $\mathrm{ET} \cop{=} 236.6\,\mathrm{s}$}
            \label{fig:target_regions_1}
        \end{subfigure}
        \hfill
        \begin{subfigure}[t]{0.325\columnwidth}
            \centering
            \includegraphics[width=\linewidth]{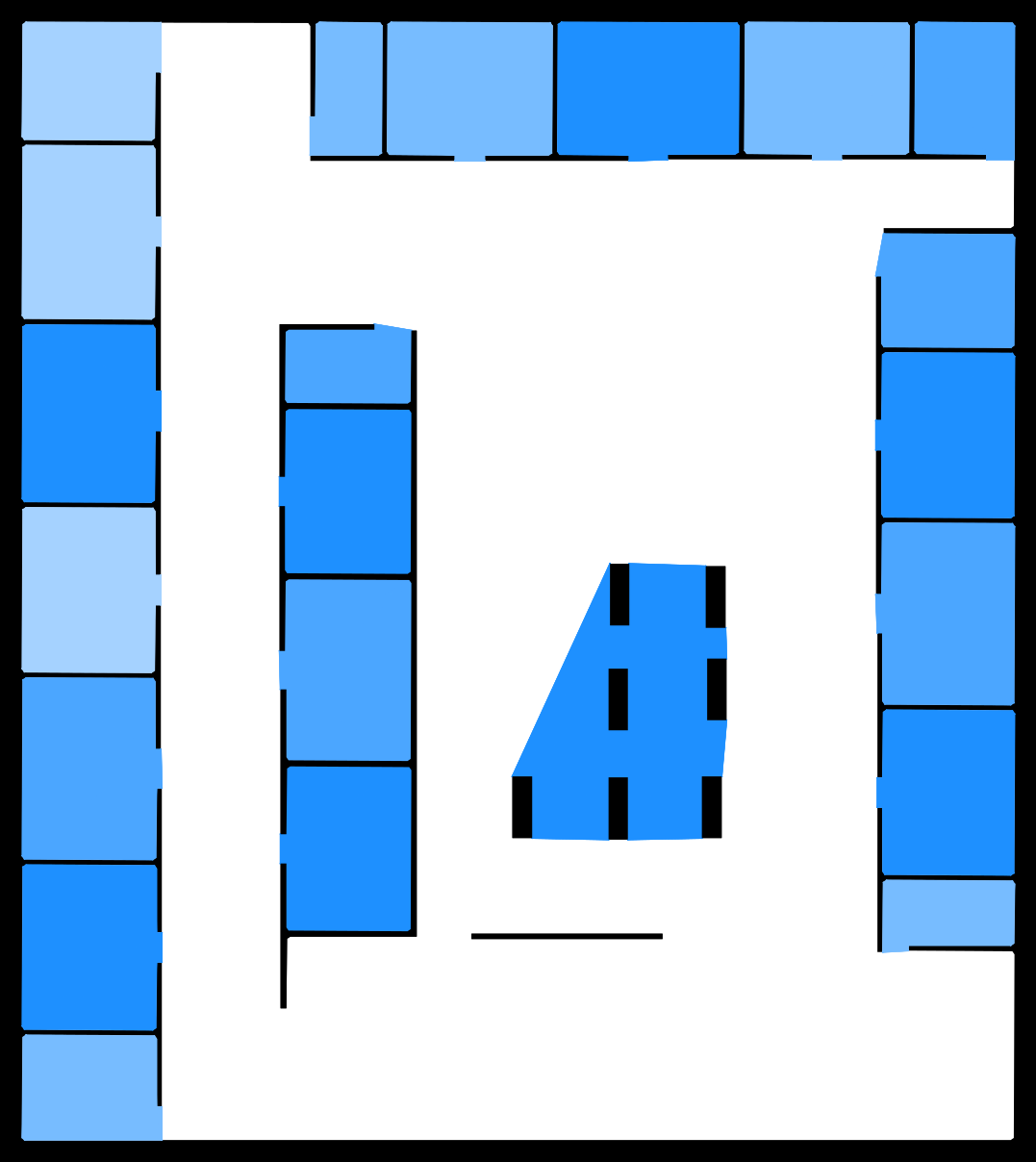}
            \caption{$P_2$, UG1: $\mathrm{ET} \cop{=} 179.2\,\mathrm{s}$}
            \label{fig:target_regions_2}
        \end{subfigure}
        \hfill
        \begin{subfigure}[t]{0.325\columnwidth}
            \centering
            \includegraphics[width=\linewidth]{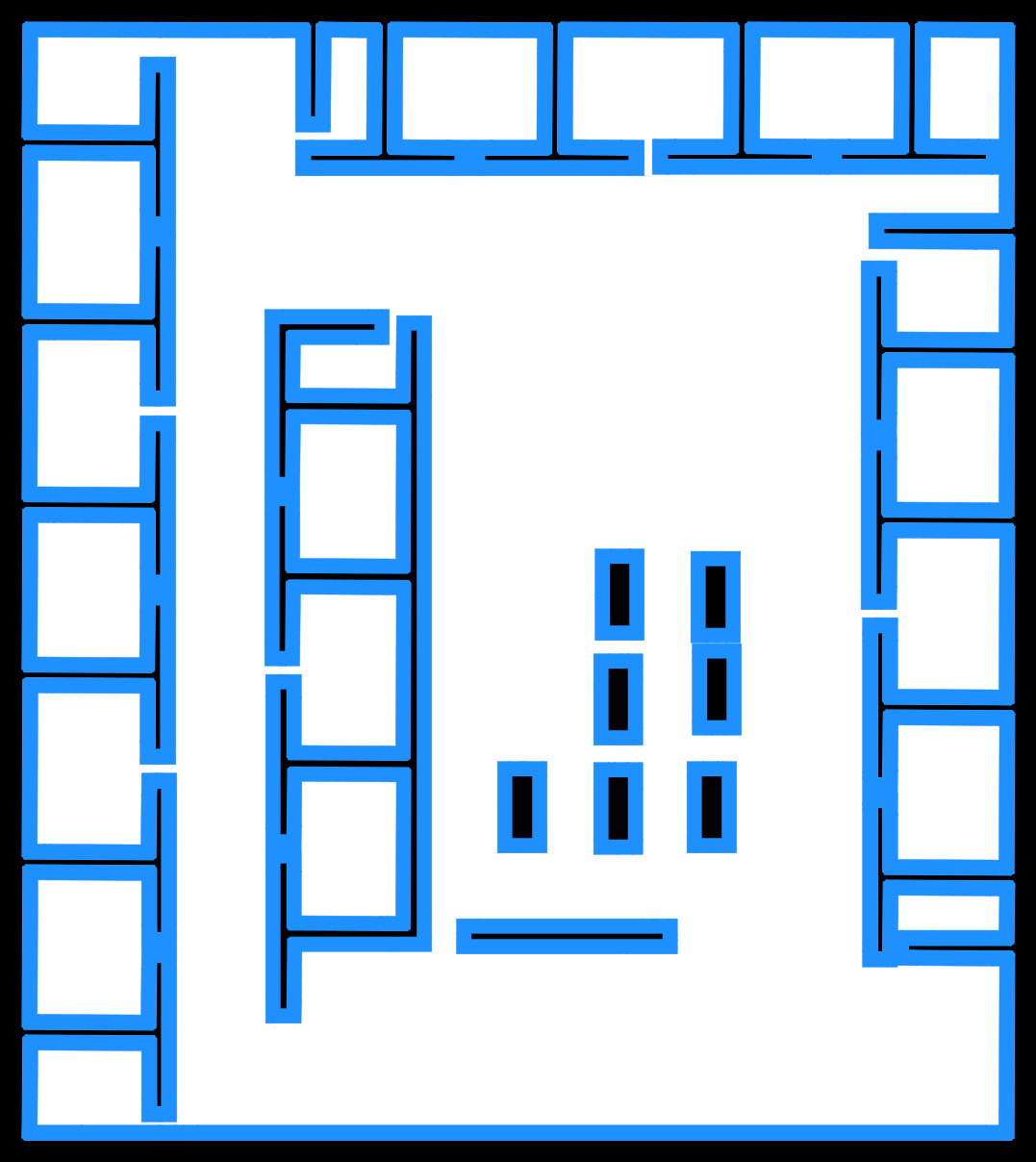}
            \caption{$P_3$, UG1: $\mathrm{ET} \cop{=} 214.6\,\mathrm{s}$}
            \label{fig:target_regions_3}
        \end{subfigure}
        \\
        \begin{subfigure}[t]{0.325\columnwidth}
            \centering
            \includegraphics[width=\linewidth]{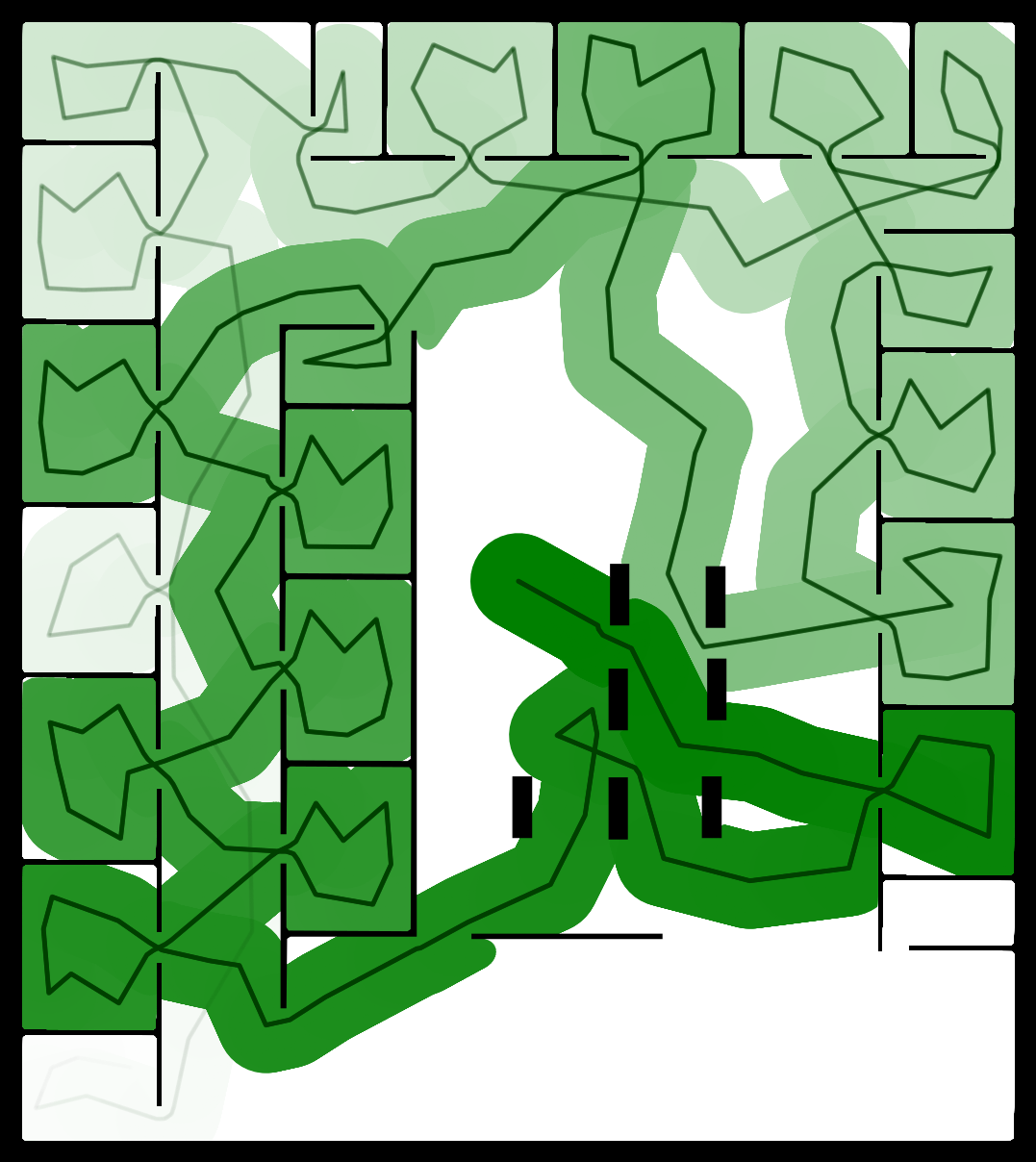}
            \caption{$\mathrm{ET} \cop{=} 204.0\,\mathrm{s}$}
            \label{fig:search_path_1}
        \end{subfigure}
        \hfill
        \begin{subfigure}[t]{0.325\columnwidth}
            \centering
            \includegraphics[width=\linewidth]{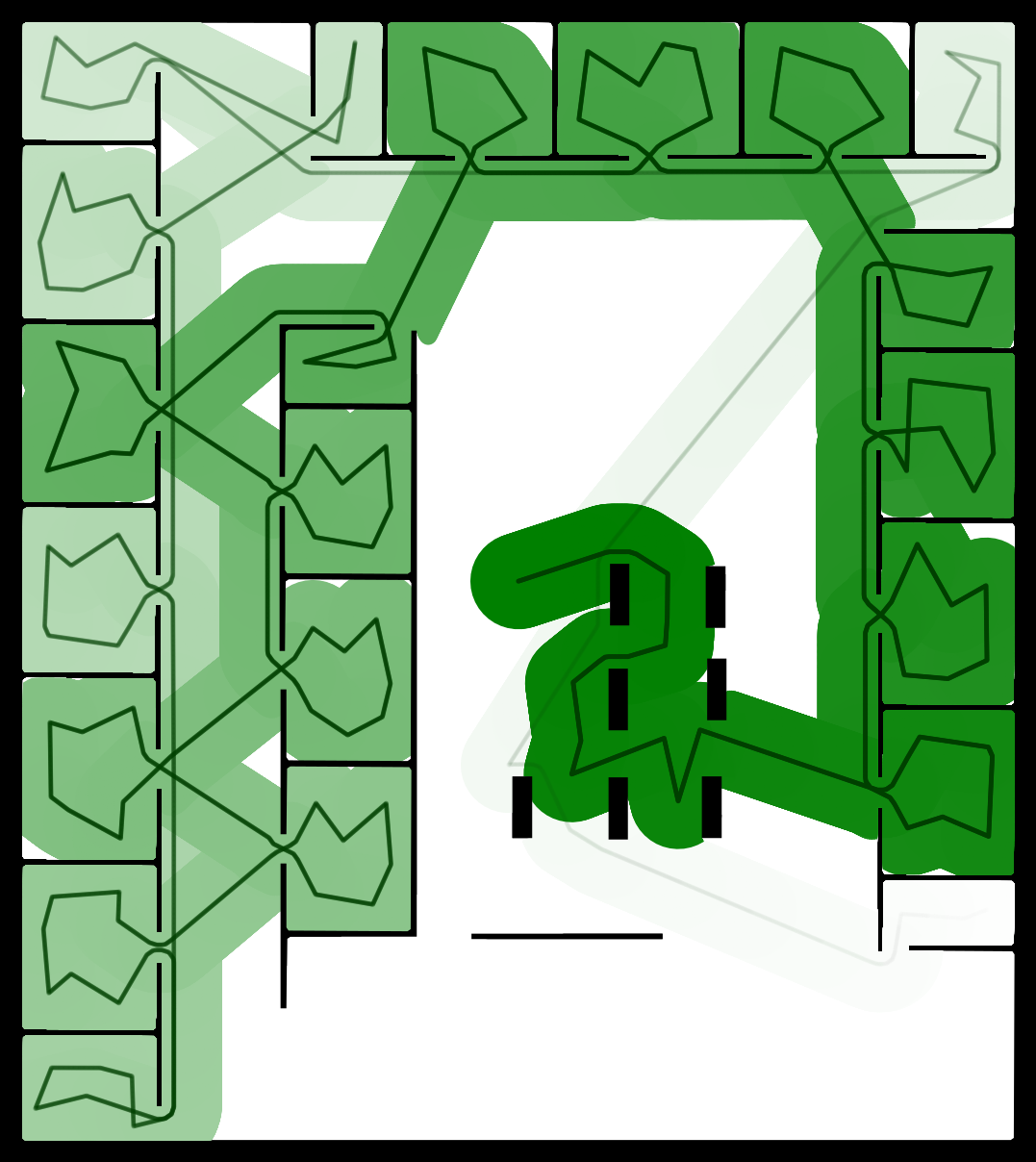}
            \caption{$\mathrm{ET} \cop{=} 156.5\,\mathrm{s}$}
            \label{fig:search_path_2}
        \end{subfigure}
        \hfill
        \begin{subfigure}[t]{0.325\columnwidth}
            \centering
            \includegraphics[width=\linewidth]{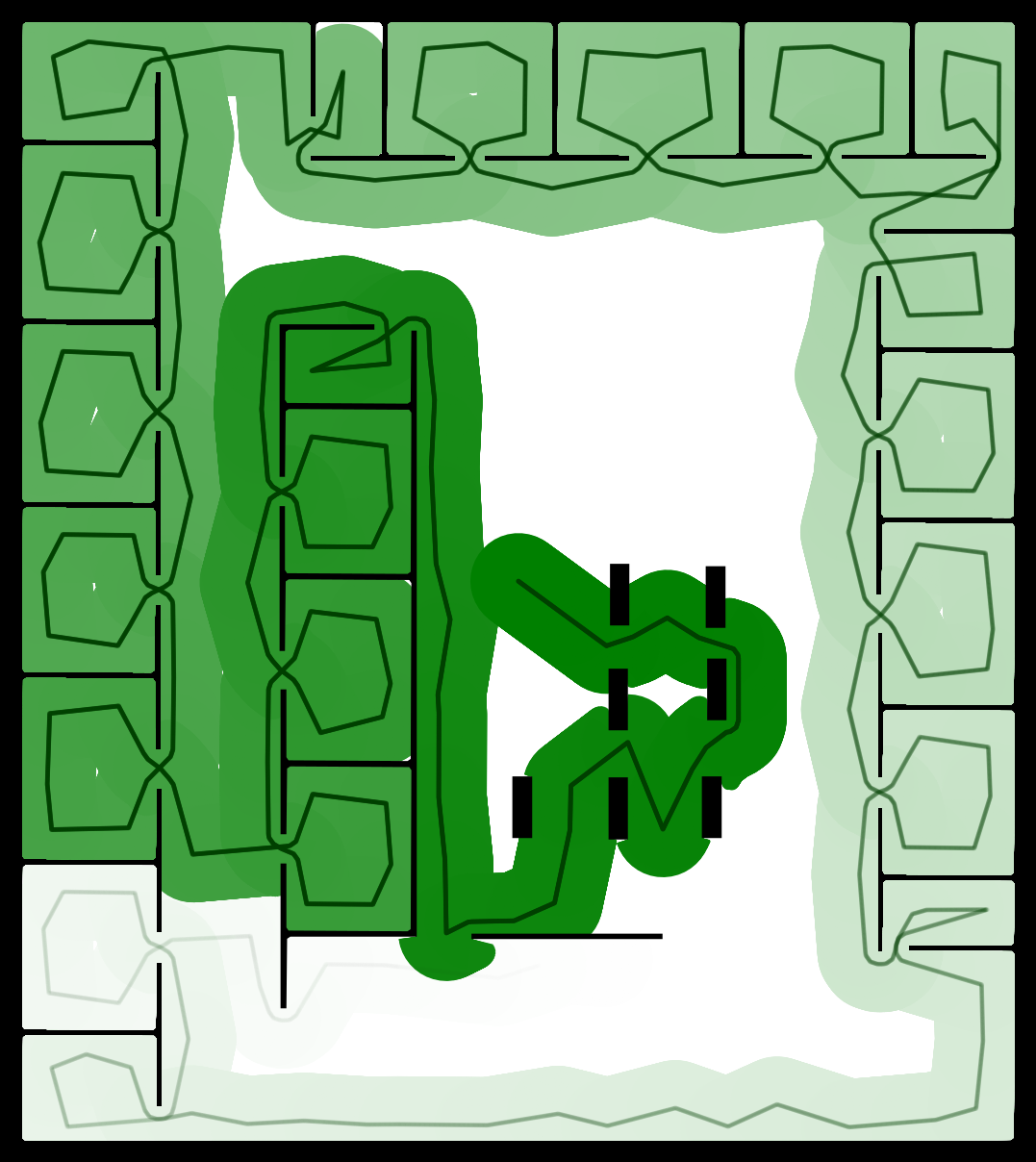}
            \caption{$\mathrm{ET} \cop{=} 203.8\,\mathrm{s}$}
            \label{fig:search_path_3}
        \end{subfigure}
        \caption{
            Varying the object's probability distribution.
        }
        \label{fig:target_regions}
    \end{figure}

    We demonstrate the framework's ability to handle varying object probability distributions in Fig.~\ref{fig:target_regions}.
    The first row presents three scenarios with different weighted target regions, $P_1$, $P_2$, and $P_3$, each defining a unique probability distribution.
    $P_1$ assigns weights between 2 and 5 to rooms and the central region, with weight 1 in the remaining open areas.
    $P_2$ follows $P_1$ but restricts the probability distribution to the rooms and central region, setting it to zero elsewhere.
    $P_3$ concentrates all probability mass near the environment boundaries, leaving the rest at zero.
    For reference, we include the objective value $\mathrm{ET}$ achieved by the UGreedy-1 baseline (UG1) for each scenario, adapted to also consider the non-uniform object distribution.
    The second row shows the search routes generated by \milaps{}-DisGreedy, along with the corresponding $\mathrm{ET}$ values, using the same visual representation as previously.
    The results highlight the qualitative differences between scenarios, with routes adapting to the probability distributions.
    In the first case, high-probability rooms are prioritized, but open areas are not necessarily avoided.
    In the second case, the route more strongly avoids open areas due to their zero probability mass, except when necessary for moving between uncovered regions.
    In the third case, the route is pushed toward the boundaries, reflecting the probability concentration in those regions.

    \subsubsection{Unstructured Environments}

\begin{figure}
        \centering
        \begin{subfigure}[t]{0.325\columnwidth}
            \centering
            \includegraphics[width=\linewidth]{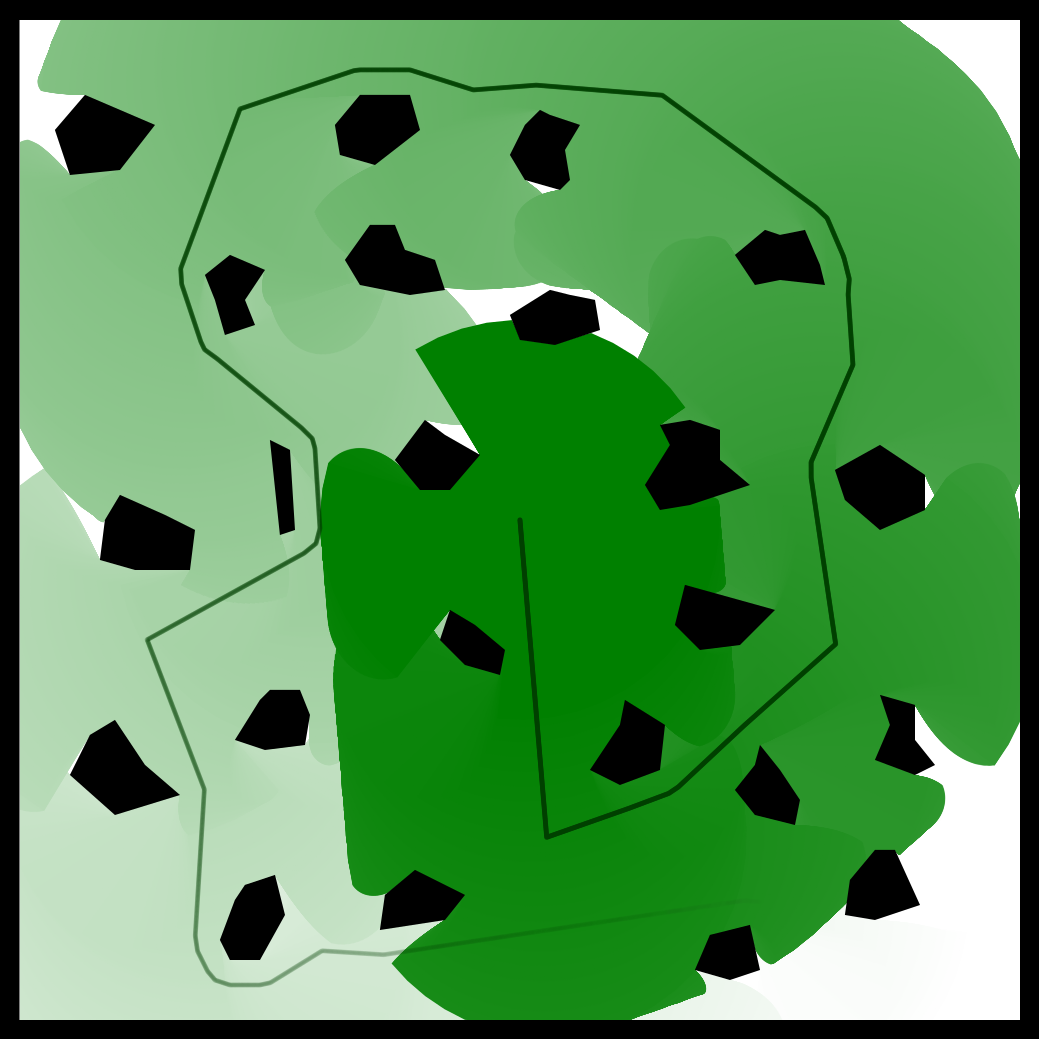}
            \caption{\textsc{potholes}}
        \end{subfigure}
        \hfill
        \begin{subfigure}[t]{0.325\columnwidth}
            \centering
            \includegraphics[width=\linewidth]{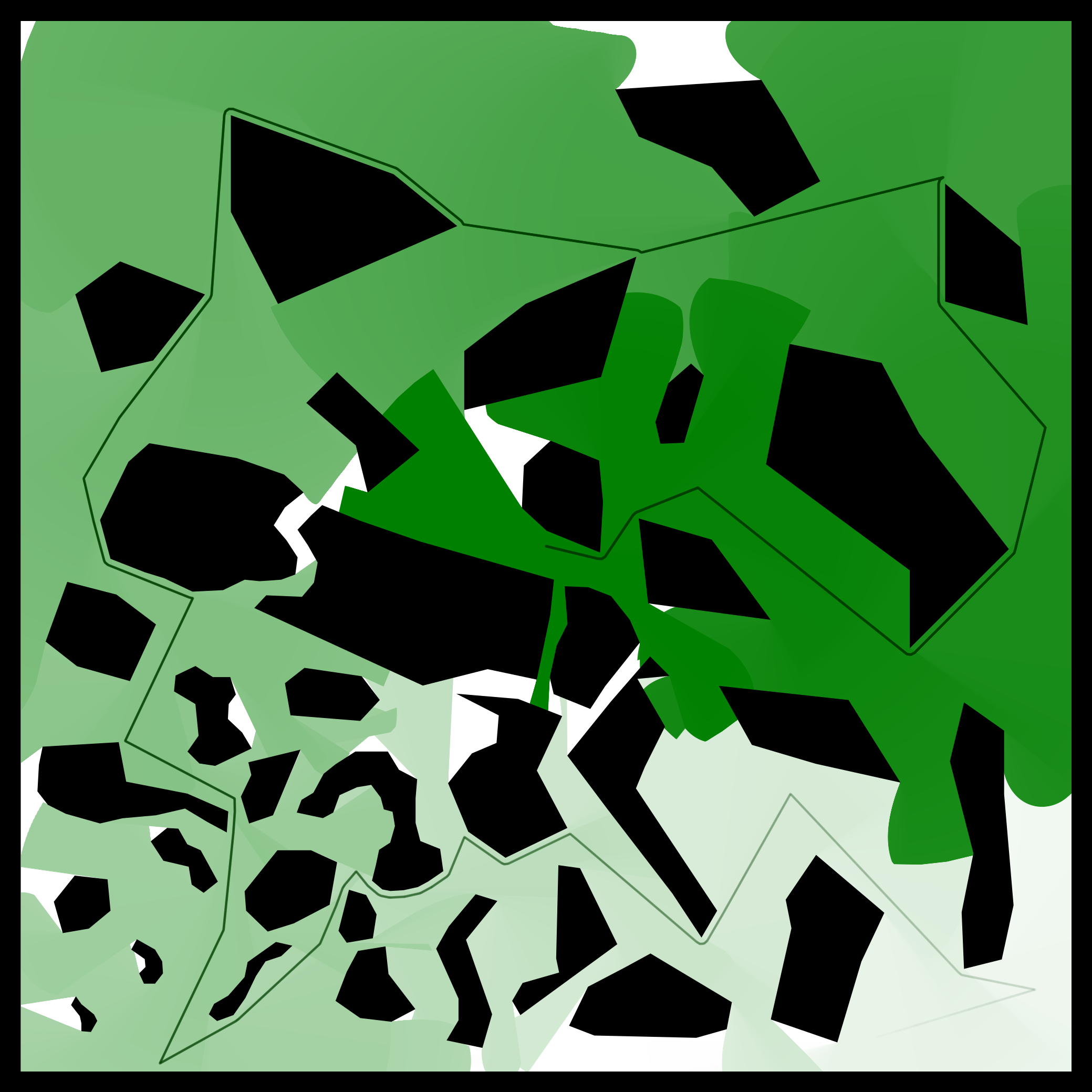}
            \caption{\textsc{large}}
        \end{subfigure}
        \hfill
        \begin{subfigure}[t]{0.325\columnwidth}
            \centering
            \includegraphics[width=\linewidth]{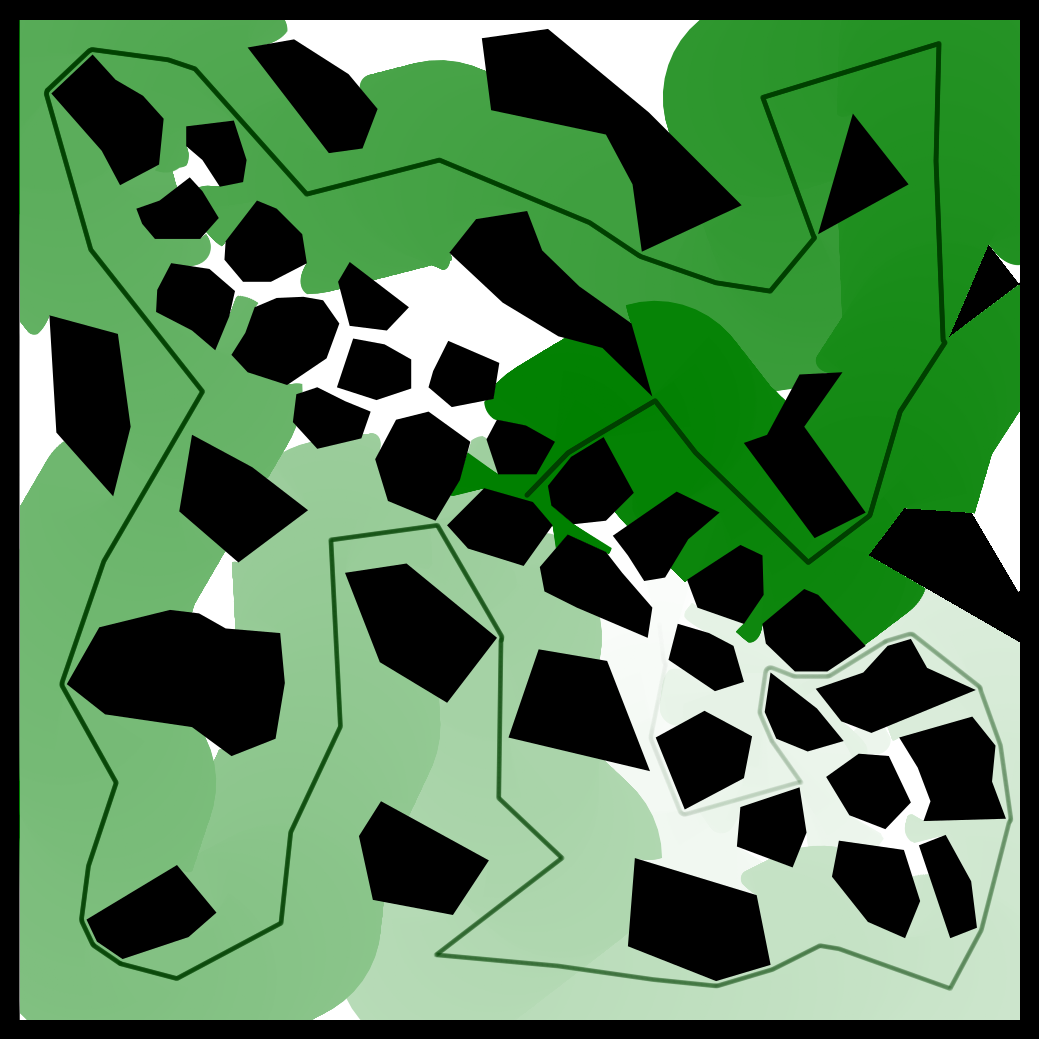}
            \caption{\textsc{var density}}
        \end{subfigure}
        \caption{
            Unstructured environments.
        }
        \label{fig:unstructured}
    \end{figure}

    Finally, to demonstrate the framework's native support for unstructured environments, we present three scenarios in Fig.~\ref{fig:unstructured}.

    \section{Conclusion and Future Work}
    \label{sec:conclusions}

    How to most effectively and efficiently optimize the challenging \search{} objective in large-scale continuous environments remains an open question.
    This work provides strong new evidence that indirect global optimization via static weight assignment is both a practically effective and scalable approach, with scalability achieved by avoiding direct evaluation of the expensive \search{} objective.
    Our key insight is that the method of weight assignment is pivotal to achieving high-quality solutions.
    Within the proposed \milaps{} framework, our experiments identify the best-performing strategy as a two-phase process: first, rapidly generating a preliminary solution route and fixing weights based on newly observed regions, then metaheuristically optimizing the global costs associated with these weights.
    This approach outperforms state-of-the-art utility-greedy heuristics and \milaps{}-integrated metaheuristic strategies with simpler weight assignments by approximately 10--20\% in terms of the best-known solution gap when runtime budgets scale linearly with problem size.
    Additionally, the anytime property of the embedded metaheuristic is preserved, allowing control over the trade-off between solution quality and runtime.
    Furthermore, we significantly expand the range of search scenarios that \milaps{} can address, demonstrating its adaptability to varying travel time models, footprint radii, and object probability distributions.

    Looking ahead, further progress can be made by addressing practical challenges and exploring new research avenues.
    Adapting \milaps{} for directional sensors and multi-agent scenarios stands out as an immediate opportunity.
    The current formulation is already close to supporting directional sensors, requiring only modifications to the visibility model, sensor placement method, and the handling of non-symmetric travel times.
    Multi-agent adaptation, on the other hand, will demand extending the optimization metaheuristic with novel operators and strategies to coordinate multiple agents.
    On a more ambitious front, fully coupled optimization of the \search{} objective---integrating guard generation with route planning---remains a compelling goal, as does exploring dynamic weight assignment strategies.

    \appendix

    \subsection{Ms-GVNS Operators and $\mathbb{O}(1)$ Improvements for \gspt{}}
    \label{app:local-search}

    For brevity, we introduce the following notation:
    \begin{flalign*}
    (\mathrm{cond}?\mathrm{expr}_1{:}\mathrm{expr}_2)
        &\coloneqq
        \begin{cases}
            \mathrm{expr}_1, \text{ if } \mathrm{cond}, \\
            \mathrm{expr}_2, \text{ otherwise}.
        \end{cases}
    \end{flalign*}

    \subsubsection{Auxiliary Structures}
    First, we define the auxiliary structures $\delta_{k},\gamma_{k},\omega_{k},f_{k},\psi_{k}$ for a fixed \gspt{} solution $\sigma \cop{:} \{0{:}\eta{-}1\} \cop{\mapsto} V$, where $\eta \cop{=} |V|$ represents the number of graph nodes.
    Constructing these structures requires $\mathcal{O}(\eta)$ and facilitates efficient improvement computations for the $\mathrm{2string}$ and $\mathrm{2opt}$ operators.
    \begin{flalign*}
        w_{k} &\cop{\coloneqq} w(\sigma(k)), \\
        d_{j,k} &\cop{\coloneqq} d(\sigma(j), \sigma(k)), \\
        \theta_{i,j,k} &\cop{\coloneqq} (i {=} {-}1{?}\vartheta(\sigma(k)){:}\theta(\sigma(i), \sigma(j), \sigma(k))), \\
        \xi_{i,j,k} &\cop{\coloneqq} \theta_{i,j,k} \cop{+} d_{j,k}, \\
        \delta_{k} &\cop{\coloneqq} (k {=} 0{?}0{:}\delta_{k{-}1} \cop{+} \xi_{k{-}2,k{-}1,k}), \\
        \gamma_{k} &\cop{\coloneqq} (k {=} 0{?}0{:}\gamma_{k{-}1} \cop{+} w_{k}), \\
        \omega_{k} &\cop{\coloneqq} (k {=} 0{?}0{:}\delta_{k}  w_{k}), \\
        f_{k} &\cop{\coloneqq} (k {=} 0{?}0{:}f_{k{-}1} \cop{+} \omega_{k}), \\
        \psi_{k} &\cop{\coloneqq} (k {=} 0 {\lor} k {=} \eta{-}1{?}0{:}\psi_{k{-}1} \cop{+} w_{k}  \theta_{k{-}1,k,k{+}1}), \\
        &\forall i \cop{\in} \{-1{:}\eta\cop{-}3\},\; \forall j \cop{\in} \{0{:}\eta\cop{-}2\},\; \forall k \cop{\in} \{1{:}\eta\cop{-}1\}.\nonumber%
    \end{flalign*}

    \subsubsection{2string Definition and Improvement Formula}

    Let the current solution $\sigma$ be represented as: $\sigma \cop{=} \langle \chi_{0}{=}v_{\mathrm{s}},\allowbreak \chi_{1},\allowbreak \ldots,\allowbreak \chi_{\eta{-}1} \rangle$.
    Consider the following parameter domains and constraints: $x, y, i, j$: $x \cop{\in} \{0{:}\eta{-}1\}$, $y \cop{\in} \{0{:}\eta{-}1\}$ s.t. $x {+} y \cop{\leq} \eta {-} 1$; $i \cop{\in} \{0{:}\eta{-}x{-}1\}$; $j \cop{\in} \{0{:}\eta{-}y{-}1\}$ s.t. $j {-} i \cop{\geq} x \cop{\lor} i {-} j \cop{\geq} y$.
    The $\mathrm{2string}$ operator swaps a string of length $x$ following $\chi_{i}$ (exclusive) with a string of length $y$ following $\chi_{j}$ (exclusive), formally defined as:
    \begin{flalign*}
        \begin{split}
            &\mathrm{2string}(\sigma, i, j, x, y) \coloneqq \langle \chi_{0}, \ldots, \chi_{i}, \chi_{j+1}, \ldots, \chi_{j+y}, \\
            &\chi_{i+x+1}, \ldots, \chi_{j}, \chi_{i+1}, \ldots, \chi_{i+x}, \chi_{j+y+1}, \ldots, \chi_{\eta-1} \rangle.
        \end{split}
    \end{flalign*}
    The operator is symmetric, meaning $\mathrm{2string}(\sigma,\allowbreak i,\allowbreak j,\allowbreak x,\allowbreak y) \cop{=} \mathrm{2string}(\sigma,\allowbreak j,\allowbreak i,\allowbreak y,\allowbreak x)$.
    Thus, we assume $i \cop{<} j$ and derive the improvement using the technique from the GSP's $\mathrm{swap}$ operator in~\citep{Kulich2022}.
    For brevity, we present only the final result, as the derivation is complex and offers no additional insight:
    \begin{flalign*}
        \Delta_{i,j,x,y}^{\mathrm{2string}} &\cop{=}
        (y{=}0?0{:}\Lambda^{\!1}_{i,j,x,y}  w_{j {+} 1} \cop{-} \omega_{j {+} 1})
        \\
        &\; {+}
        (y{\leq}1?0{:}\Lambda^{\!2}_{i,j,x,y}  ( \gamma_{j {+} y} \cop{-} \gamma_{j {+} 1} ))
        \\
        &\; {+}
        (i{+}x{=}j?0{:}\Lambda^{\!3}_{i,j,x,y}  w_{i {+} x {+} 1} \cop{-} \omega_{i {+} x {+} 1})
        \\
        &\; {+}
        (i{+}x{+}1{\geq}j?0{:}\Lambda^{\!4}_{i,j,x,y}  ( \gamma_{j} \cop{-} \gamma_{i {+} x {+} 1} ))
        \\
        &\; {+}
        (x{=}0?0{:}\Lambda^{\!5}_{i,j,x,y}  w_{i {+} 1} \cop{-} \omega_{i {+} 1})
        \\
        &\; {+}
        (x{\leq}1?0{:}\Lambda^{\!6}_{i,j,x,y}  ( \gamma_{i {+} x} \cop{-} \gamma_{i {+} 1} ))
        \\
        &\; {+}
        (j{+}y{=}\eta{-}1?0{:}\Lambda^{\!7}_{i,j,x,y}  w_{j {+} y {+} 1} \cop{-} \omega_{j {+} y {+} 1})
        \\
        &\; {+}
        (j{+}y{+}1{\geq}\eta{-}1?0{:}\Lambda^{\!8}_{i,j,x,y}  ( \gamma_{\eta {-} 1} \cop{-} \gamma_{j {+} y {+} 1} )),
        \\
        \Lambda^{\!1}_{i,j,x,y} &\cop{\coloneqq}
        (y{=}0?0{:}\delta_{i} \cop{+} \xi_{i {-} 1, i, j {+} 1}),
        \\
        \Lambda^{\!2}_{i,j,x,y} &\cop{\coloneqq} \Lambda^{\!1}_{i,j,x,y} \cop{+}
        (y{\leq}1?0{:}\xi_{i, j {+} 1, j {+} 2} \cop{-} \delta_{j {+} 2}),
        \\
        \Lambda^{\!3}_{i,j,x,y} &\cop{\coloneqq} \Lambda^{\!2}_{i,j,x,y} \cop{+}\!
        \begin{cases}
            0, \text{ if } i {+} x {=} j, \\
            \delta_{i} \cop{+} \xi_{i {-} 1, i, i {+} x {+} 1}, \text{ if } y {=} 0, \\
            \xi_{i, j {+} y, i {+} x {+} 1}, \text{ if } y {=} 1, \\
            \delta_{j {+} y} \cop{+} \xi_{j {+} y {-} 1, j {+} y, i {+} x {+} 1}, \text{ otherwise},
        \end{cases}
        \\
        \Lambda^{\!4}_{i,j,x,y} &\cop{\coloneqq} \Lambda^{\!3}_{i,j,x,y} \cop{+}\!
        \begin{cases}
            0, \text{ if } i {+} x {+} 1 {\geq} j, \\
            \xi_{i, i {+} x {+} 1, i {+} x {+} 2} \cop{-} \delta_{i {+} x {+} 2}, \text{ if } y {=} 0, \\
            \xi_{j {+} y, i {+} x {+} 1, i {+} x {+} 2} \cop{-} \delta_{i {+} x {+} 2}, \text{ otherwise},
        \end{cases}
        \\
        \Lambda^{\!5}_{i,j,x,y} &\cop{\coloneqq} \Lambda^{\!4}_{i,j,x,y} \cop{+}\!
        \begin{cases}
            0, \text{ if } x {=} 0, \\
            \xi_{i, j {+} y, i {+} 1}, \text{ if } y {=} 1 \cop{\land} i {+} x {=} j, \\
            \delta_{j {+} y} \cop{+} \xi_{j {+} y {-} 1, j {+} y, i {+} 1}, \text{ if } i {+} x {=} j \\
            \xi_{i, j, i {+} 1}, \text{ if } y {=} 0 \cop{\land} i {+} x {+} 1 {=} j, \\
            \xi_{j {+} y, j, i {+} 1} , \text{ if } i {+} x {+} 1 {=} j, \\
            \delta_{j} \cop{+} \xi_{j {-} 1, j, i {+} 1}, \text{ otherwise},
        \end{cases}
        \\
        \Lambda^{\!6}_{i,j,x,y} &\cop{\coloneqq} \Lambda^{\!5}_{i,j,x,y} \cop{+}\!
        \begin{cases}
            0, \text{ if } x {\leq} 1, \\
            \xi_{j {+} y, i {+} 1, i {+} 2} \cop{-} \delta_{i {+} 2} \text{ if } i {+} x {=} j, \\
            \xi_{j, i {+} 1, i {+} 2} \cop{-} \delta_{i {+} 2}, \text{ otherwise},
        \end{cases}
        \\
        \Lambda^{\!7}_{i,j,x,y} &\cop{\coloneqq} \Lambda^{\!6}_{i,j,x,y} \cop{+}\!
        \begin{cases}
            0, \text{ if } j {+} y {=} \eta {-} 1, \\
            \xi_{j {+} y, j, j {+} y {+} 1}, \text{ if } i {+} x {+} 1 {=} j \cop{\land} x {=} 0, \\
            \delta_{j} \cop{+} \xi_{j {-} 1, j, j {+} y {+} 1}, \text{ if } x {=} 0, \\
            \xi_{j {+} y, i {+} x, j {+} y {+} 1}, \text{ if } i {+} x {=} j \cop{\land} x {=} 1, \\
            \xi_{j, i {+} x, j {+} y {+} 1}, \text{ if } x {=} 1, \\
            \delta_{i {+} x} \cop{+} \xi_{i {+} x {-} 1, i {+} x, j {+} y {+} 1}, \text{ otherwise},
        \end{cases}
        \\
        \Lambda^{\!8}_{i,j,x,y} &\cop{\coloneqq} \Lambda^{\!7}_{i,j,x,y} \cop{+}\!
        \begin{cases}
            0, \text{ if } j {+} y {+} 1 {\geq} \eta {-} 1, \\
            \xi_{j, j {+} y {+} 1, j {+} y {+} 2} \cop{-} \delta_{j {+} y {+} 2}, \text{ if } x {=} 0, \\
            \xi_{i {+} x, j {+} y {+} 1, j {+} y {+} 2} \cop{-} \delta_{j {+} y {+} 2}, \text{ otherwise}.
        \end{cases}
    \end{flalign*}

    \subsubsection{2opt Definition and Improvement Formula}

    Let the current solution $\sigma$ be represented as: $\sigma \cop{=} \langle \chi_{0}{=}v_{\mathrm{s}},\allowbreak \chi_{1},\allowbreak \ldots,\allowbreak \chi_{\eta{-}1} \rangle$.
    Consider the parameter domains: $i, j$: $i \cop{\in} \{1{:}\eta\cop{-}2\}$, $j \cop{\in} \{i\cop{+}1{:}\eta\cop{-}1\}$.
    The $\mathrm{2opt}$ operator reverses the order of nodes between $\chi_i$ and $\chi_j$ (inclusive), formally defined as:
    \begin{flalign*}
        \begin{split}
            &\mathrm{2opt}(\sigma, i, j) \coloneqq \langle \chi_{0}, \ldots, \chi_{i-1}, \chi_{j}, \chi_{j-1}, \ldots, \chi_{i+1}, \chi_{i}, \\
            &\chi_{j+1}, \ldots, \chi_{\eta-1} \rangle.
        \end{split}
    \end{flalign*}
    We derive the improvement using the technique applied to the TDP’s and GSP’s $\mathrm{2opt}$ operator in~\citep{Mladenovic2013} and~\citep{Kulich2022}, respectively.
    This approach assumes a symmetric graph; for asymmetric cases, direct computation via Eq.~\eqref{eq:improvement} is required.
    For brevity, we present only the final result, as the derivation is complex and offers no additional insight:
    \begin{flalign*}
        \Delta_{i,j}^{\mathrm{2opt}} &\cop{=} 2 f_{i{-}1} \cop{+} w_j  ( \delta_{i{-}1} \cop{+} \Lambda^{\!1}_{i,j} \cop{+} \delta_{j} ) \cop{+} \psi_{i {-} 1} \cop{-} \psi_{j {-} 1}
        \\
        &\cop{+} ( \delta_{i {-} 1} \cop{+} \Lambda^{\!2}_{i,j} \cop{+} \delta_{j} )  ( \gamma_{j {-} 1} \cop{-} \gamma_{i {-} 1} )\cop{+}(j {=} \eta {-} 1?{-}2 f_{\eta {-} 1}
        \\
        &\quad{:}w_{j {+} 1}  ( 2 \delta_{j {+} 1} \cop{+} \Lambda^{\!3}_{i,j} ) \cop{+} \Lambda^{\!4}_{i,j}  ( \gamma_{\eta {-} 1} \cop{-} \gamma_{j {+} 1} ) \cop{-} 2 f_{j {+} 1})
        \\
        \Lambda^{\!1}_{i,j} &\cop{\coloneqq} \xi_{i {-} 2, i {-} 1, j},
        \\
        \Lambda^{\!2}_{i,j} &\cop{\coloneqq} \Lambda^{\!1}_{i,j} \cop{+} \theta_{i {-} 1, j, j {-} 1},
        \\
        \Lambda^{\!3}_{i,j} &\cop{\coloneqq} \Lambda^{\!2}_{i,j} \cop{-} \xi_{i {-} 2, i {-} 1, i} \cop{-} \theta_{i {-} 1, i, i {+} 1} \cop{+} \xi_{i {+} 1, i, j {+} 1} \cop{-} \xi_{j {-} 1, j, j {+} 1},
        \\
        \Lambda^{\!4}_{i,j} &\cop{\coloneqq} \Lambda^{\!3}_{i,j} \cop{+}
        (j {=} \eta {-} 2?0{:}\theta_{i, j {+} 1, j {+} 2} \cop{-} \theta_{j, j {+} 1, j {+} 2})
    \end{flalign*}

    \subsection{Dataset Details}
    \label{app:dataset}

    We preprocessed the OURS and IH map datasets to generate well-formed, connected polygons with holes, ensuring they were free of self-intersections, overlapping holes, and redundant vertices.
    The process involved applying the Ramer-Douglas-Peucker (RDP) algorithm~\citep{Douglas1973} with a 0.1\,m tolerance, followed by inflation–deflation with a 0.2\,m radius, a final inflation with a 0.01\,m radius, and reapplying the RDP algorithm (implemented in Clipper2).
    We retained the largest polygon along with its enclosed holes, discarding disconnected artifacts.
    Tab.~\ref{tab:maps} summarizes the processed maps' properties.

    The \dsearch{} dataset was generated by fixing the SPP method to HAR-KA,RV and varying the visibility range $r_{\mathrm{vis}}$ in small increments for each map in Tab.~\ref{tab:maps}.
    The resulting superset was refined by partitioning the plane defined by $n_G$ and $o_G$ into rectangular tiles and discarding those with fewer than 15 unique maps, leaving 16 tiles.
    From these, 15 instances were selected per tile, ensuring unique maps and metrics closest to the tile center, forming the final 16 subsets.
    A summary of the dataset is provided in Tab.~\ref{tab:dataset}.

    \begin{table}
        \centering
        \caption{Map dataset.}
        \label{tab:maps}
        \scriptsize
        \setlength{\tabcolsep}{4pt} 
        \begin{threeparttable}
            \begin{tabularx}{\columnwidth}{@{} rXrrrrrrr @{}}
                \toprule
                \# & \textsc{map name}             & $m$   & $h$ & $x$ & $y$ & $b$     & $f$ & $a$     \\
                \midrule
                1  & \textsc{narrow corridor}      & 18    & 0   & 20  & 20  & 4.00e+2 & 55  & 2.20e+2 \\
                2  & \textsc{clasp center}         & 24    & 0   & 20  & 20  & 4.00e+2 & 98  & 3.91e+2 \\
                3  & \textsc{slits easy}           & 28    & 0   & 28  & 10  & 2.74e+2 & 92  & 2.54e+2 \\
                4  & \textsc{square spiral}        & 32    & 0   & 20  & 20  & 4.00e+2 & 89  & 3.54e+2 \\
                5  & \textsc{complex3}             & 36    & 1   & 20  & 20  & 4.00e+2 & 58  & 2.30e+2 \\
                6  & \textsc{clasps}               & 44    & 0   & 20  & 20  & 4.00e+2 & 95  & 3.81e+2 \\
                7  & \textsc{rooms easy}           & 44    & 0   & 20  & 20  & 4.00e+2 & 91  & 3.65e+2 \\
                8  & \textsc{tunnel twisted}       & 48    & 0   & 20  & 20  & 4.00e+2 & 48  & 1.93e+2 \\
                9  & \textsc{complex}              & 63    & 1   & 20  & 20  & 4.00e+2 & 80  & 3.18e+2 \\
                10 & \textsc{brick pattern}        & 78    & 14  & 20  & 20  & 4.00e+2 & 29  & 1.17e+2 \\
                11 & \textsc{staggered brick wall} & 92    & 18  & 20  & 20  & 4.00e+2 & 48  & 1.94e+2 \\
                12 & \textsc{maze}                 & 116   & 9   & 20  & 20  & 4.00e+2 & 58  & 2.32e+2 \\
                13 & \textsc{var density4}         & 125   & 28  & 20  & 20  & 4.00e+2 & 72  & 2.88e+2 \\
                14 & \textsc{var density3}         & 139   & 24  & 20  & 20  & 4.00e+2 & 63  & 2.51e+2 \\
                15 & \textsc{warehouse}            & 142   & 24  & 40  & 40  & 1.60e+3 & 74  & 1.19e+3 \\
                16 & \textsc{potholes}             & 153   & 23  & 20  & 20  & 4.00e+2 & 92  & 3.66e+2 \\
                17 & \textsc{plankpile}            & 172   & 28  & 20  & 20  & 4.00e+2 & 68  & 2.72e+2 \\
                18 & \textsc{var density2}         & 189   & 20  & 20  & 20  & 4.00e+2 & 71  & 2.84e+2 \\
                19 & \textsc{geometry}             & 253   & 36  & 20  & 20  & 4.00e+2 & 68  & 2.72e+2 \\
                20 & \textsc{jari-huge}            & 278   & 9   & 21  & 23  & 4.79e+2 & 96  & 4.59e+2 \\
                21 & \textsc{large}                & 326   & 35  & 40  & 40  & 1.60e+3 & 72  & 1.15e+3 \\
                22 & \textsc{rockpile}             & 379   & 20  & 20  & 20  & 4.00e+2 & 45  & 1.80e+2 \\
                \midrule
                23 & \textsc{pol01}                & 959   & 51  & 323 & 133 & 4.28e+4 & 30  & 1.28e+4 \\
                24 & \textsc{2p04}                 & 998   & 52  & 240 & 310 & 7.44e+4 & 71  & 5.28e+4 \\
                25 & \textsc{rus02}                & 1,337 & 72  & 242 & 307 & 7.45e+4 & 42  & 3.11e+4 \\
                26 & \textsc{cha01}                & 1,357 & 112 & 230 & 280 & 6.44e+4 & 65  & 4.20e+4 \\
                27 & \textsc{2p02}                 & 1,428 & 137 & 270 & 270 & 7.29e+4 & 74  & 5.39e+4 \\
                28 & \textsc{sax01}                & 1,583 & 127 & 380 & 485 & 1.84e+5 & 43  & 7.86e+4 \\
                29 & \textsc{sax05}                & 1,623 & 54  & 445 & 420 & 1.87e+5 & 46  & 8.62e+4 \\
                30 & \textsc{2p01}                 & 1,909 & 140 & 189 & 210 & 3.96e+4 & 79  & 3.15e+4 \\
                31 & \textsc{cha02}                & 2,108 & 101 & 335 & 570 & 1.91e+5 & 92  & 1.76e+5 \\
                32 & \textsc{rus07}                & 2,147 & 137 & 460 & 380 & 1.75e+5 & 49  & 8.52e+4 \\
                33 & \textsc{rus01}                & 2,331 & 134 & 331 & 224 & 7.40e+4 & 45  & 3.32e+4 \\
                34 & \textsc{2p03}                 & 2,347 & 153 & 330 & 310 & 1.02e+5 & 58  & 5.89e+4 \\
                35 & \textsc{6p03}                 & 2,464 & 229 & 500 & 500 & 2.50e+5 & 61  & 1.52e+5 \\
                36 & \textsc{sax06}                & 2,524 & 163 & 405 & 465 & 1.88e+5 & 51  & 9.70e+4 \\
                37 & \textsc{sax07}                & 2,758 & 165 & 310 & 340 & 1.05e+5 & 66  & 6.92e+4 \\
                38 & \textsc{sax03}                & 2,827 & 143 & 416 & 462 & 1.92e+5 & 45  & 8.63e+4 \\
                39 & \textsc{pol05}                & 2,860 & 239 & 515 & 395 & 2.03e+5 & 42  & 8.54e+4 \\
                40 & \textsc{4p01}                 & 2,919 & 274 & 320 & 320 & 1.02e+5 & 74  & 7.53e+4 \\
                41 & \textsc{rus04}                & 3,198 & 265 & 338 & 500 & 1.69e+5 & 62  & 1.04e+5 \\
                42 & \textsc{pol02}                & 3,296 & 239 & 470 & 515 & 2.42e+5 & 40  & 9.65e+4 \\
                43 & \textsc{6p02}                 & 3,419 & 214 & 400 & 440 & 1.76e+5 & 74  & 1.30e+5 \\
                44 & \textsc{rus05}                & 3,459 & 220 & 404 & 419 & 1.69e+5 & 50  & 8.41e+4 \\
                45 & \textsc{cha03}                & 3,462 & 320 & 400 & 430 & 1.72e+5 & 58  & 9.97e+4 \\
                46 & \textsc{rus03}                & 3,463 & 295 & 450 & 430 & 1.93e+5 & 36  & 6.93e+4 \\
                47 & \textsc{6p01}                 & 3,558 & 234 & 368 & 498 & 1.84e+5 & 66  & 1.22e+5 \\
                48 & \textsc{4p02}                 & 3,799 & 315 & 380 & 502 & 1.91e+5 & 58  & 1.10e+5 \\
                49 & \textsc{pol04}                & 3,978 & 268 & 350 & 340 & 1.19e+5 & 61  & 7.25e+4 \\
                50 & \textsc{pol03}                & 4,118 & 394 & 420 & 510 & 2.14e+5 & 59  & 1.27e+5 \\
                51 & \textsc{sax02}                & 4,448 & 255 & 403 & 634 & 2.56e+5 & 46  & 1.18e+5 \\
                52 & \textsc{sax04}                & 4,639 & 286 & 585 & 675 & 3.95e+5 & 35  & 1.40e+5 \\
                53 & \textsc{cha04}                & 4,688 & 407 & 440 & 440 & 1.94e+5 & 62  & 1.21e+5 \\
                54 & \textsc{4p03}                 & 4,838 & 300 & 400 & 410 & 1.64e+5 & 60  & 9.77e+4 \\
                55 & \textsc{endmaps}              & 4,923 & 340 & 565 & 770 & 4.35e+5 & 83  & 3.60e+5 \\
                56 & \textsc{rus06}                & 5,145 & 383 & 545 & 455 & 2.48e+5 & 45  & 1.12e+5 \\
                57 & \textsc{pol06}                & 5,315 & 465 & 470 & 480 & 2.26e+5 & 69  & 1.57e+5 \\
                \bottomrule
            \end{tabularx}
            \begin{tablenotes}
                \footnotesize
                \item \emph{Legend:} $m$ -- no.\ vertices, $h$ -- no.\ holes, $x$ -- width [m], $y$ -- height [m], $b$ -- bounding box area [m$^2$], $f$ -- free space ratio [\%], $a$ -- free space area [m$^2$].
                \item Maps are sorted by increasing $m$.
                \item Maps 23--57 are from~\citep{Harabor2022}.
            \end{tablenotes}
        \end{threeparttable}
    \end{table}

    \begin{table*}
        \centering
        \caption{Dataset of 240 \dsearch{} instances organized into 16 subsets (IDs 0--15) with 15 instances each (IDs 0--14).}
        \label{tab:dataset}
        \scriptsize
        \setlength{\tabcolsep}{4pt} 
        \begin{threeparttable}
            \begin{tabularx}{\textwidth}{@{} cccccX @{}}
                \toprule
                & \multicolumn{2}{c}{No.\ guards $n_G$} & \multicolumn{2}{c}{Overlap. factor $o_G$} & \\
                \cmidrule(lr){2-3} \cmidrule(lr){4-5}
                ID & Range        & m\,\textpm{}\,std    & Range & m\,\textpm{}\,std   & List of \dsearch{} instances, each encoded as \textbf{ID:} \textsc{map name}: $r_{\mathrm{vis}}$ [m] ($n_G$, $o_G$)                                                                                                                                                                                                                                                                                                                                                                                                                                                                                                                                                                                                                                                               \\
                \midrule
                0  & 0--200       & 43\,\textpm{}\,29    & 0--1  & 0.7\,\textpm{}\,0.2 & {\tiny \textbf{0:} \textsc{narrow corridor}: 5.87 (8, 0.5); \textbf{1:} \textsc{clasp center}: 8.13 (8, 0.5); \textbf{2:} \textsc{square spiral}: 3.32 (33, 0.6); \textbf{3:} \textsc{complex3}: 5.16 (11, 0.5); \textbf{4:} \textsc{clasps}: 5.44 (16, 0.8); \textbf{5:} \textsc{tunnel twisted}: 3.39 (29, 0.6); \textbf{6:} \textsc{complex}: 6.50 (14, 0.6); \textbf{7:} \textsc{brick pattern}: 1.77 (68, 0.7); \textbf{8:} \textsc{staggered brick wall}: 1.84 (65, 0.7); \textbf{9:} \textsc{maze}: 3.39 (39, 0.6); \textbf{10:} \textsc{var density4}: 1.77 (94, 1.0); \textbf{11:} \textsc{warehouse}: 10.04 (31, 1.0); \textbf{12:} \textsc{plankpile}: 2.69 (69, 0.5); \textbf{13:} \textsc{geometry}: 1.98 (79, 0.9); \textbf{14:} \textsc{rockpile}: 1.56 (78, 0.6)} \\
                1  & 0--200       & 123\,\textpm{}\,52   & 1--2  & 1.5\,\textpm{}\,0.2 & {\tiny \textbf{0:} \textsc{slits easy}: 1.42 (114, 1.4); \textbf{1:} \textsc{tunnel twisted}: 2.26 (56, 1.4); \textbf{2:} \textsc{staggered brick wall}: 1.20 (156, 1.2); \textbf{3:} \textsc{var density4}: 1.91 (88, 1.2); \textbf{4:} \textsc{potholes}: 2.19 (79, 1.5); \textbf{5:} \textsc{var density2}: 3.18 (53, 1.5); \textbf{6:} \textsc{jari-huge}: 2.25 (110, 1.5); \textbf{7:} \textsc{large}: 5.52 (71, 1.5); \textbf{8:} \textsc{rockpile}: 2.90 (59, 1.4); \textbf{9:} \textsc{pol01}: 11.34 (157, 1.5); \textbf{10:} \textsc{2p04}: 39.20 (129, 1.5); \textbf{11:} \textsc{cha01}: 20.83 (193, 1.8); \textbf{12:} \textsc{2p02}: 22.91 (198, 1.8); \textbf{13:} \textsc{sax01}: 33.89 (193, 1.8); \textbf{14:} \textsc{sax05}: 38.24 (183, 1.6)}                 \\
                2  & 0--200       & 80\,\textpm{}\,55    & 2--3  & 2.5\,\textpm{}\,0.1 & {\tiny \textbf{0:} \textsc{var density4}: 16.19 (20, 2.5); \textbf{1:} \textsc{var density3}: 13.22 (23, 2.4); \textbf{2:} \textsc{potholes}: 8.06 (19, 2.6); \textbf{3:} \textsc{var density2}: 14.56 (29, 2.5); \textbf{4:} \textsc{geometry}: 13.79 (35, 2.5); \textbf{5:} \textsc{jari-huge}: 7.69 (37, 2.5); \textbf{6:} \textsc{large}: 25.31 (42, 2.4); \textbf{7:} \textsc{rockpile}: 11.52 (47, 2.4); \textbf{8:} \textsc{pol01}: 23.56 (99, 2.5); \textbf{9:} \textsc{2p04}: 128.39 (110, 2.3); \textbf{10:} \textsc{rus02}: 47.94 (168, 2.6); \textbf{11:} \textsc{cha01}: 31.70 (142, 2.5); \textbf{12:} \textsc{2p02}: 45.82 (131, 2.7); \textbf{13:} \textsc{sax01}: 64.69 (149, 2.5); \textbf{14:} \textsc{sax05}: 85.66 (151, 2.5)}                               \\
                3  & 200--400     & 323\,\textpm{}\,30   & 1--2  & 1.5\,\textpm{}\,0.3 & {\tiny \textbf{0:} \textsc{slits easy}: 0.82 (297, 1.2); \textbf{1:} \textsc{complex}: 0.92 (322, 1.2); \textbf{2:} \textsc{maze}: 0.85 (305, 1.2); \textbf{3:} \textsc{var density2}: 0.92 (300, 1.2); \textbf{4:} \textsc{large}: 1.84 (316, 1.2); \textbf{5:} \textsc{rockpile}: 0.78 (315, 1.4); \textbf{6:} \textsc{2p04}: 15.68 (292, 1.4); \textbf{7:} \textsc{cha01}: 13.59 (310, 1.6); \textbf{8:} \textsc{2p02}: 15.27 (309, 1.5); \textbf{9:} \textsc{sax05}: 19.89 (316, 1.3); \textbf{10:} \textsc{2p01}: 12.70 (334, 1.7); \textbf{11:} \textsc{rus07}: 26.85 (295, 1.8); \textbf{12:} \textsc{2p03}: 16.98 (357, 1.8); \textbf{13:} \textsc{sax06}: 24.66 (379, 2.0); \textbf{14:} \textsc{4p01}: 20.36 (398, 1.9)}                                                \\
                4  & 200--400     & 312\,\textpm{}\,42   & 2--3  & 2.5\,\textpm{}\,0.1 & {\tiny \textbf{0:} \textsc{2p01}: 18.35 (254, 2.4); \textbf{1:} \textsc{cha02}: 59.50 (255, 2.6); \textbf{2:} \textsc{rus07}: 34.31 (267, 2.4); \textbf{3:} \textsc{2p03}: 24.90 (271, 2.5); \textbf{4:} \textsc{6p03}: 58.33 (258, 2.4); \textbf{5:} \textsc{sax06}: 38.54 (295, 2.5); \textbf{6:} \textsc{sax07}: 34.51 (356, 2.7); \textbf{7:} \textsc{pol05}: 34.06 (321, 2.5); \textbf{8:} \textsc{4p01}: 29.41 (320, 2.6); \textbf{9:} \textsc{rus04}: 37.73 (368, 2.6); \textbf{10:} \textsc{pol02}: 43.57 (362, 2.5); \textbf{11:} \textsc{6p02}: 43.09 (331, 2.6); \textbf{12:} \textsc{rus03}: 93.36 (380, 2.7); \textbf{13:} \textsc{6p01}: 43.38 (314, 2.6); \textbf{14:} \textsc{4p02}: 34.61 (330, 2.5)}                                                            \\
                5  & 200--400     & 292\,\textpm{}\,53   & 3--4  & 3.5\,\textpm{}\,0.1 & {\tiny \textbf{0:} \textsc{2p01}: 24.70 (223, 3.3); \textbf{1:} \textsc{rus07}: 55.19 (230, 3.4); \textbf{2:} \textsc{2p03}: 40.75 (211, 3.4); \textbf{3:} \textsc{6p03}: 93.69 (247, 3.4); \textbf{4:} \textsc{sax06}: 63.20 (249, 3.4); \textbf{5:} \textsc{sax07}: 66.71 (329, 3.5); \textbf{6:} \textsc{pol05}: 53.52 (287, 3.4); \textbf{7:} \textsc{4p01}: 39.60 (293, 3.4); \textbf{8:} \textsc{rus04}: 63.38 (331, 3.5); \textbf{9:} \textsc{pol02}: 104.58 (333, 3.5); \textbf{10:} \textsc{6p02}: 65.38 (298, 3.6); \textbf{11:} \textsc{cha03}: 64.60 (395, 3.6); \textbf{12:} \textsc{6p01}: 71.27 (274, 3.6); \textbf{13:} \textsc{4p02}: 48.77 (298, 3.5); \textbf{14:} \textsc{pol03}: 57.81 (383, 3.7)}                                                           \\
                6  & 200--400     & 279\,\textpm{}\,46   & 4--5  & 4.4\,\textpm{}\,0.1 & {\tiny \textbf{0:} \textsc{cha02}: 137.18 (216, 4.6); \textbf{1:} \textsc{rus07}: 135.73 (215, 4.4); \textbf{2:} \textsc{rus01}: 53.89 (270, 4.5); \textbf{3:} \textsc{6p03}: 385.36 (233, 4.0); \textbf{4:} \textsc{sax06}: 174.19 (237, 4.5); \textbf{5:} \textsc{sax07}: 218.54 (324, 4.4); \textbf{6:} \textsc{sax03}: 107.29 (278, 4.3); \textbf{7:} \textsc{pol05}: 100.56 (270, 4.6); \textbf{8:} \textsc{4p01}: 67.88 (262, 4.5); \textbf{9:} \textsc{rus04}: 191.65 (323, 4.5); \textbf{10:} \textsc{6p02}: 268.93 (280, 4.5); \textbf{11:} \textsc{cha03}: 110.11 (379, 4.5); \textbf{12:} \textsc{6p01}: 156.48 (263, 4.5); \textbf{13:} \textsc{4p02}: 64.50 (284, 4.5); \textbf{14:} \textsc{pol03}: 128.83 (352, 4.5)}                                              \\
                7  & 400--600     & 504\,\textpm{}\,24   & 1--2  & 1.4\,\textpm{}\,0.2 & {\tiny \textbf{0:} \textsc{brick pattern}: 0.49 (453, 1.1); \textbf{1:} \textsc{var density2}: 0.71 (481, 1.2); \textbf{2:} \textsc{jari-huge}: 0.85 (527, 1.2); \textbf{3:} \textsc{rockpile}: 0.57 (520, 1.2); \textbf{4:} \textsc{2p04}: 10.78 (494, 1.3); \textbf{5:} \textsc{2p02}: 10.50 (514, 1.3); \textbf{6:} \textsc{sax05}: 13.77 (517, 1.2); \textbf{7:} \textsc{2p01}: 9.17 (480, 1.5); \textbf{8:} \textsc{rus07}: 14.92 (521, 1.4); \textbf{9:} \textsc{2p03}: 12.45 (493, 1.5); \textbf{10:} \textsc{sax06}: 18.50 (477, 1.6); \textbf{11:} \textsc{4p01}: 14.71 (535, 1.7); \textbf{12:} \textsc{pol02}: 20.92 (502, 1.7); \textbf{13:} \textsc{6p01}: 20.14 (506, 1.6); \textbf{14:} \textsc{4p02}: 18.88 (546, 1.9)}                                           \\
                8  & 400--600     & 471\,\textpm{}\,58   & 2--3  & 2.4\,\textpm{}\,0.2 & {\tiny \textbf{0:} \textsc{sax07}: 26.45 (403, 2.4); \textbf{1:} \textsc{rus04}: 28.67 (414, 2.2); \textbf{2:} \textsc{pol02}: 33.12 (404, 2.3); \textbf{3:} \textsc{6p02}: 28.23 (435, 2.2); \textbf{4:} \textsc{rus05}: 27.65 (458, 2.5); \textbf{5:} \textsc{cha03}: 27.89 (487, 2.3); \textbf{6:} \textsc{rus03}: 38.90 (420, 2.3); \textbf{7:} \textsc{6p01}: 26.34 (418, 2.0); \textbf{8:} \textsc{4p02}: 23.60 (437, 2.2); \textbf{9:} \textsc{pol04}: 21.96 (549, 2.5); \textbf{10:} \textsc{pol03}: 31.38 (479, 2.5); \textbf{11:} \textsc{sax02}: 30.06 (514, 2.7); \textbf{12:} \textsc{sax04}: 49.13 (501, 2.5); \textbf{13:} \textsc{cha04}: 29.56 (584, 2.7); \textbf{14:} \textsc{endmaps}: 78.79 (565, 2.8)}                                                      \\
                9  & 600--800     & 702\,\textpm{}\,43   & 1--2  & 1.5\,\textpm{}\,0.3 & {\tiny \textbf{0:} \textsc{brick pattern}: 0.42 (604, 1.2); \textbf{1:} \textsc{var density2}: 0.57 (729, 1.2); \textbf{2:} \textsc{rockpile}: 0.49 (647, 1.1); \textbf{3:} \textsc{2p04}: 8.82 (685, 1.2); \textbf{4:} \textsc{2p02}: 8.59 (697, 1.2); \textbf{5:} \textsc{sax05}: 10.71 (758, 1.2); \textbf{6:} \textsc{2p01}: 7.06 (688, 1.4); \textbf{7:} \textsc{rus07}: 11.93 (691, 1.3); \textbf{8:} \textsc{sax06}: 13.87 (673, 1.5); \textbf{9:} \textsc{4p01}: 11.31 (719, 1.5); \textbf{10:} \textsc{6p01}: 15.49 (677, 1.5); \textbf{11:} \textsc{pol03}: 18.17 (699, 1.8); \textbf{12:} \textsc{cha04}: 18.67 (749, 1.8); \textbf{13:} \textsc{endmaps}: 38.20 (730, 1.9); \textbf{14:} \textsc{pol06}: 23.51 (782, 2.0)}                                            \\
                10 & 800--1,000   & 888\,\textpm{}\,39   & 1--2  & 1.4\,\textpm{}\,0.2 & {\tiny \textbf{0:} \textsc{var density4}: 0.49 (969, 1.2); \textbf{1:} \textsc{rockpile}: 0.42 (890, 1.2); \textbf{2:} \textsc{2p04}: 7.84 (843, 1.3); \textbf{3:} \textsc{sax01}: 9.24 (893, 1.3); \textbf{4:} \textsc{2p01}: 5.65 (941, 1.2); \textbf{5:} \textsc{rus07}: 10.44 (832, 1.3); \textbf{6:} \textsc{sax06}: 10.79 (926, 1.4); \textbf{7:} \textsc{4p01}: 10.18 (841, 1.4); \textbf{8:} \textsc{pol02}: 12.20 (872, 1.4); \textbf{9:} \textsc{6p01}: 12.39 (890, 1.4); \textbf{10:} \textsc{4p02}: 12.59 (850, 1.5); \textbf{11:} \textsc{pol03}: 14.86 (858, 1.6); \textbf{12:} \textsc{cha04}: 15.56 (902, 1.7); \textbf{13:} \textsc{endmaps}: 28.65 (887, 1.7); \textbf{14:} \textsc{pol06}: 18.47 (933, 1.7)}                                                   \\
                11 & 1,000--1,200 & 1,103\,\textpm{}\,58 & 1--2  & 1.3\,\textpm{}\,0.1 & {\tiny \textbf{0:} \textsc{potholes}: 0.49 (1,154, 1.2); \textbf{1:} \textsc{rockpile}: 0.35 (1,161, 1.1); \textbf{2:} \textsc{2p04}: 6.86 (1,065, 1.3); \textbf{3:} \textsc{2p02}: 6.68 (1,069, 1.2); \textbf{4:} \textsc{sax05}: 9.18 (1,004, 1.2); \textbf{5:} \textsc{2p01}: 4.94 (1,189, 1.3); \textbf{6:} \textsc{rus07}: 8.95 (1,061, 1.3); \textbf{7:} \textsc{sax06}: 9.25 (1,160, 1.3); \textbf{8:} \textsc{4p01}: 9.05 (1,001, 1.4); \textbf{9:} \textsc{pol02}: 10.46 (1,065, 1.3); \textbf{10:} \textsc{6p01}: 10.85 (1,067, 1.3); \textbf{11:} \textsc{pol03}: 11.56 (1,159, 1.4); \textbf{12:} \textsc{cha04}: 12.44 (1,116, 1.5); \textbf{13:} \textsc{endmaps}: 21.49 (1,148, 1.5); \textbf{14:} \textsc{pol06}: 15.11 (1,125, 1.6)}                             \\
                12 & 1,200--1,400 & 1,301\,\textpm{}\,51 & 1--2  & 1.3\,\textpm{}\,0.1 & {\tiny \textbf{0:} \textsc{brick pattern}: 0.28 (1,320, 1.3); \textbf{1:} \textsc{var density2}: 0.42 (1,276, 1.2); \textbf{2:} \textsc{large}: 0.85 (1,286, 1.2); \textbf{3:} \textsc{2p04}: 5.88 (1,371, 1.2); \textbf{4:} \textsc{cha01}: 5.43 (1,223, 1.2); \textbf{5:} \textsc{sax05}: 7.65 (1,343, 1.2); \textbf{6:} \textsc{6p03}: 10.61 (1,306, 1.3); \textbf{7:} \textsc{4p01}: 7.92 (1,215, 1.3); \textbf{8:} \textsc{pol02}: 8.71 (1,379, 1.3); \textbf{9:} \textsc{rus03}: 7.78 (1,337, 1.2); \textbf{10:} \textsc{6p01}: 9.30 (1,356, 1.2); \textbf{11:} \textsc{pol04}: 8.54 (1,236, 1.4); \textbf{12:} \textsc{cha04}: 10.89 (1,325, 1.4); \textbf{13:} \textsc{endmaps}: 19.10 (1,296, 1.4); \textbf{14:} \textsc{pol06}: 13.44 (1,249, 1.4)}                     \\
                13 & 1,400--1,600 & 1,485\,\textpm{}\,56 & 1--2  & 1.3\,\textpm{}\,0.1 & {\tiny \textbf{0:} \textsc{rooms easy}: 0.42 (1,536, 1.2); \textbf{1:} \textsc{potholes}: 0.42 (1,538, 1.2); \textbf{2:} \textsc{2p02}: 5.73 (1,404, 1.2); \textbf{3:} \textsc{2p01}: 4.23 (1,541, 1.2); \textbf{4:} \textsc{rus07}: 7.46 (1,449, 1.2); \textbf{5:} \textsc{sax06}: 7.71 (1,575, 1.2); \textbf{6:} \textsc{sax03}: 7.77 (1,405, 1.3); \textbf{7:} \textsc{4p01}: 6.79 (1,523, 1.2); \textbf{8:} \textsc{cha03}: 8.81 (1,445, 1.3); \textbf{9:} \textsc{pol04}: 7.32 (1,530, 1.3); \textbf{10:} \textsc{pol03}: 9.91 (1,430, 1.3); \textbf{11:} \textsc{sax02}: 9.39 (1,474, 1.3); \textbf{12:} \textsc{endmaps}: 16.71 (1,527, 1.4); \textbf{13:} \textsc{rus06}: 10.65 (1,405, 1.4); \textbf{14:} \textsc{pol06}: 11.76 (1,494, 1.4)}                            \\
                14 & 1,600--1,800 & 1,693\,\textpm{}\,68 & 1--2  & 1.3\,\textpm{}\,0.1 & {\tiny \textbf{0:} \textsc{clasp center}: 0.42 (1,612, 1.2); \textbf{1:} \textsc{clasps}: 0.42 (1,616, 1.2); \textbf{2:} \textsc{var density3}: 0.35 (1,601, 1.2); \textbf{3:} \textsc{geometry}: 0.35 (1,773, 1.2); \textbf{4:} \textsc{jari-huge}: 0.47 (1,655, 1.2); \textbf{5:} \textsc{cha01}: 4.53 (1,673, 1.2); \textbf{6:} \textsc{sax01}: 6.16 (1,783, 1.2); \textbf{7:} \textsc{2p03}: 5.66 (1,649, 1.2); \textbf{8:} \textsc{6p03}: 8.84 (1,790, 1.2); \textbf{9:} \textsc{rus04}: 7.55 (1,789, 1.2); \textbf{10:} \textsc{rus05}: 7.28 (1,676, 1.3); \textbf{11:} \textsc{4p02}: 7.87 (1,685, 1.3); \textbf{12:} \textsc{cha04}: 9.33 (1,620, 1.4); \textbf{13:} \textsc{4p03}: 8.59 (1,701, 1.4); \textbf{14:} \textsc{rus06}: 8.87 (1,765, 1.3)}                    \\
                15 & 1,800--2,000 & 1,892\,\textpm{}\,55 & 1--2  & 1.2\,\textpm{}\,0.1 & {\tiny \textbf{0:} \textsc{staggered brick wall}: 0.28 (1,972, 1.2); \textbf{1:} \textsc{var density2}: 0.35 (1,806, 1.2); \textbf{2:} \textsc{large}: 0.71 (1,824, 1.2); \textbf{3:} \textsc{rockpile}: 0.28 (1,817, 1.2); \textbf{4:} \textsc{2p04}: 4.90 (1,885, 1.2); \textbf{5:} \textsc{2p02}: 4.77 (1,986, 1.2); \textbf{6:} \textsc{rus01}: 3.99 (1,939, 1.2); \textbf{7:} \textsc{pol05}: 6.49 (1,918, 1.3); \textbf{8:} \textsc{pol02}: 6.97 (1,960, 1.3); \textbf{9:} \textsc{rus03}: 6.22 (1,871, 1.2); \textbf{10:} \textsc{6p01}: 7.75 (1,863, 1.2); \textbf{11:} \textsc{pol03}: 8.26 (1,881, 1.3); \textbf{12:} \textsc{sax04}: 8.93 (1,918, 1.3); \textbf{13:} \textsc{endmaps}: 14.33 (1,901, 1.3); \textbf{14:} \textsc{pol06}: 10.08 (1,843, 1.4)}            \\
                \bottomrule
            \end{tabularx}
            \begin{tablenotes}
                \footnotesize
                \item \emph{Legend:} $n_G$ -- no.\ guards, $o_G$ -- overlapping factor, $r_{\mathrm{vis}}$ -- visibility range.
                \item The individual values of $r_{\mathrm{vis}}$ and $o_G$ for each instance have been rounded.
                \item Three instances with IDs 2, 7, and 12 were excluded from each subset for preliminary experiments and parameter tuning, leaving 12 instances (0, 1, 3--6, 8--11, 13, 14) for the main evaluation.
            \end{tablenotes}
        \end{threeparttable}
    \end{table*}

    \bibliographystyle{IEEEtran}
    \bibliography{main}

\end{document}